%% file: arxiv.tex
\documentclass[preprint,
a4paper]{elsarticle}

\usepackage{amssymb}
\journal{%Journal of Logical and Algebraic Methods in Programming
}

\usepackage{xspace,stmaryrd,amsmath,amssymb,mathtools,comment,numprint,adjustbox,subcaption} 
\usepackage{todonotes}

\allowdisplaybreaks

\usepackage{algorithm}
\usepackage[noend]{algpseudocode} 
\usepackage{etoolbox}

\algnewcommand\algorithmicforeach{\textbf{for each}}
\algdef{S}[FOR]{ForEach}[1]{\algorithmicforeach\ #1\ \algorithmicdo}
\usepackage{wrapfig}
\usepackage[all,pdftex]{xy}
\CompileMatrices 
\xyoption{v2}
\xyoption{curve}
\xyoption{2cell}
\SelectTips{cm}{}  % Tips (of arrows) are in accordance with Computer Modern
\UseAllTwocells
\usepackage{tikz}
\usepackage{tikz-qtree}
\usetikzlibrary{automata,positioning,matrix,shapes.callouts,patterns,arrows}
\usepackage{aliascnt}
\usepackage{apxproof}

\usepackage[most]{tcolorbox}

\tcbset{
    frame code={}
    center title,
    left=0pt,
    right=0pt,
    top=0pt,
    bottom=0pt,
    colback=gray!70,
    colframe=white,
    width=\dimexpr\textwidth\relax,
    enlarge left by=0mm,
    boxsep=5pt,
    arc=0pt,outer arc=0pt,
    }

\specialcomment{auxproof}
{\mbox{}\newline\textbf{BEGIN: AUX-PROOF}\dotfill\newline}
{\mbox{}\newline\textbf{END: AUX-PROOF}\dotfill\newline}
\excludecomment{auxproof}

\usepackage{listings}

\usepackage{lstautogobble}
\makeatletter
\let\old@lstKV@SwitchCases\lstKV@SwitchCases
\def\lstKV@SwitchCases#1#2#3{}
\makeatother
\usepackage{lstlinebgrd}
\makeatletter
\let\lstKV@SwitchCases\old@lstKV@SwitchCases

\lst@Key{numbers}{none}{%
    \def\lst@PlaceNumber{\lst@linebgrd}%
    \lstKV@SwitchCases{#1}%
    {none:\\%
     left:\def\lst@PlaceNumber{\llap{\normalfont
                \lst@numberstyle{\thelstnumber}\kern\lst@numbersep}\lst@linebgrd}\\%
     right:\def\lst@PlaceNumber{\rlap{\normalfont
                \kern\linewidth \kern\lst@numbersep
                \lst@numberstyle{\thelstnumber}}\lst@linebgrd}%
    }{\PackageError{Listings}{Numbers #1 unknown}\@ehc}}
\makeatother

\newcommand{\PIMP}{\textsc{pIMP}}
\newcommand{\Dist}{\mathbf{Dist}}
\newcommand{\observeClause}[1]{\mathtt{observe}\;#1}
\newcommand{\weightClause}[1]{\mathtt{weight}\;#1}
\newcommand{\returnClause}[1]{\mathtt{return}\;#1}
\newcommand{\Var}{\mathbf{Var}}
\newcommand{\uop}{\mathop{\mathbf{uop}}}
\newcommand{\bop}{\mathbin{\mathbf{bop}}}

\newcommand{\skipCmd}{\mathtt{skip}}
\newcommand{\ifClause}[3]{\mathtt{if}\;#1\;\mathtt{then}\;#2\;\mathtt{else}\;#3}
\newcommand{\ifpClause}[3]{\mathtt{ifp}\;#1\;\mathtt{then}\;#2\;\mathtt{else}\;#3}
\newcommand{\whileClause}[2]{\mathtt{while}(#1)\{#2\}}
\newcommand{\true}{\mathtt{true}}
\newcommand{\false}{\mathtt{false}}
\newcommand{\R}{{\mathbb{R}}}
\newcommand{\Rnn}{{\mathbb{R}_{\ge 0}}}

\newcommand{\Zpos}{{\mathbb{Z}_{>0}}}

\newcommand{\DDom}{{\mathbb{D}}}
\newcommand{\linit}{l_{\mathrm{init}}}
\newcommand{\siginit}{\sigma_{\mathrm{init}}}
\newcommand{\lfinal}{l_{\mathrm{final}}}
\newcommand{\efinal}{e_{\mathrm{final}}}
\newcommand{\trrel}{\rightarrow}

\newcommand{\dist}{\mathcal{D}}
\newcommand{\subdist}{\mathcal{D}_{\le 1}}
\newcommand{\StrLn}{\mathrm{SLP}}

\newcommand{\CP}{\mathrm{CP}^{\mathrm{DR}}}
\newcommand{\CPpr}{\mathrm{CP}}
\newcommand{\sem}[1]{\llbracket #1 \rrbracket} 
\newcommand{\place}{\underline{\phantom{n}}\,} % place holder
\newcommand{\St}{\mathbf{St}}
\newcommand{\CF}{\mathbf{CF}}
\newcommand{\CCF}{\mathbf{CCF}}

\newcommand{\st}{\mathsf{st}}
\newcommand{\wst}{\mathsf{w\text{-}st}}

\newcommand{\paramDist}{\vec{e}}
\newcommand{\fuzzyPred}{f}
\newcommand{\sharpPred}{\varphi}

\newcommand{\Schism}{\textrm{Schism}}

\newcommand{\UCBV}{\mathsf{UCB\text{-}V}}

\newcommand{\decayFn}{\mathcal{C}}
\newcommand{\flowl}{\raisebox{0ex}[0ex][0ex]{$\vec{l}$}}
\newcommand{\truelik}[1]{p_{#1}}
\newcommand{\loc}{\mathop{\mathrm{loc}}}
\newcommand{\height}{\mathop{\mathrm{height}}}
\newcommand{\cnode}{\mathsf{cnode}}
\DeclareMathOperator*{\argmin}{arg\,min}

\newcommand\numberthis{\addtocounter{equation}{1}\tag{\theequation}}

\newcommand{\myparagraph}[1]{\paragraph{#1}}
\newcommand{\url}[1]{\texttt{#1}}

\usepackage{amsthm}
\theoremstyle{definition}
\newtheorem{mydefinition}{Definition}
\theoremstyle{plain}

\newtheorem{myproposition}[mydefinition]{Proposition}
\newtheorem{mytheorem}[mydefinition]{Theorem}

\theoremstyle{definition}
\newtheorem{myremark}[mydefinition]{Remark}

\newtheorem{myexample}[mydefinition]{Example}

\theoremstyle{remark}
\newtheorem*{myproof}{Proof}
\def\myqed{\qed}

\begin{document}

\begin{frontmatter}



\title{Control-Data Separation and Logical Condition Propagation for Efficient Inference on Probabilistic Programs}


\author[nii,sokendai]{Ichiro Hasuo\fnref{equal}} %{equal,nii,sokendai}
\author[ey,nii,sokendai]{Yuichiro Oyabu\fnref{equal}} %{equal,nii,sokendai}
\author[nii]{Clovis Eberhart\fnref{equal}} %{equal,nii}
\author[kyoto]{Kohei Suenaga} %{kyoto}
\author[nii]{Kenta Cho} %{nii}
\author[nii]{Shin-ya Katsumata} %{nii}

\fntext[equal]{Equal contribution.}

\affiliation[nii]{organization={National Institute of Informatics},%Department and Organization
            addressline={Hitotsubashi 2-1-2}, 
            city={Chiyoda},
            postcode={101-8430}, 
            state={Tokyo},
            country={Japan}}

\affiliation[sokendai]{organization={Department of Informatics, SOKENDAI (The Graduate University for Advanced Studies)},%Department and Organization
            addressline={Shonan Village}, 
            city={Hayama},
            postcode={240-0193}, 
            state={Kanagawa},
            country={Japan}}

\affiliation[ey]{organization={Ernst \& Young ShinNihon LLC},%Department and Organization
            addressline={Yurakucho 1-1-2}, 
            city={Chiyoda},
            postcode={100-0006}, 
            state={Tokyo},
            country={Japan}}

\affiliation[kyoto]{organization={Graduate School of Informatics, Kyoto University},%Department and Organization
            addressline={Yoshida-Honmachi 36-1}, 
            city={Kyoto},
            postcode={606-8501}, 
            state={Kyoto},
            country={Japan}}

\begin{abstract}
We present a novel sampling framework for probabilistic programs. The framework combines two recent ideas---\emph{control-data separation} and \emph{logical condition propagation}---in a nontrivial manner so that the two ideas boost the benefits of each other. We implemented our algorithm on top of Anglican. The experimental results demonstrate our algorithm's efficiency, especially for programs with while loops and rare observations.

\end{abstract}

% hogehoge

% \begin{highlights}
% \end{highlights}


\end{frontmatter}



\emph{To Luis Barbosa on the occasion of his sixtieth birthday}. Luis's works have always been inspirations and encouragements for us, showing the remarkable power of logic in various applications, such as reactive systems, cyber-physical systems, quantum systems, and data governance. The current work follows this spirit, demonstrating the power of (program) logic in statistics.

\section{Introduction}\label{sec:intro}

\myparagraph{Probabilistic Programs}
In the recent rise of 
%data science and
 statistical  machine learning,
% and artificial intelligence, 
\emph{probabilistic programming languages} are
attracting a lot attention as a programming infrastructure for data
processing tasks. Probabilistic programming frameworks allow  users to
express statistical models as programs, and offer a variety of methods for
analyzing the models.

Probabilistic
programs feature \emph{randomization} and \emph{conditioning}. 
% See
% Program~\ref{listing:ProbProgramCoinEmulation} \&~\ref{listing:obsInLoop}. 
Randomization can take
different forms, such as probabilistic branching ($\mathtt{ifp}$ in Program~\ref{listing:ProbProgramCoinEmulation}) and
random assignment from a
probability distribution (denoted by $\sim$, see Program~\ref{listing:obsInLoop}). 

 Conditioning---also
called \emph{observation} and \emph{evidence}---makes the
operational meaning of probabilistic programs unique and distinct from
 usual programs. Here, the meaning of ``execution'' is blurry since
some execution traces get discarded due to observation
violation. For example, in Line~8 of Program~\ref{listing:ProbProgramCoinEmulation}, we
have $(c1,c2)=(\true,\true)$ with probability $0.36^{2}$; such an execution
violates the observation \texttt{!(c1 = c2)} in Line~8 and is thus discarded.

It is  suitable to think of a probabilistic programming
language as a \emph{modeling} language:
%  (rather than a programming
% language): 
a probabilistic program is not executed but is
inferred on.  Specifically, the randomization constructs in a program define the
\emph{prior} distribution; it is transformed by  other commands
such as deterministic assignment; and the semantics of the program is the \emph{posterior}
distribution, conditioned by the observation commands therein. This
\emph{Bayesian} view\footnote{Our interpretation of the word ``Bayesian'' in this paper is a permissive one, the one described in the preceding discussion. More restricted interpretations are common, too, in which 1) examples such as Program~\ref{listing:poisCdS} and~\ref{listing:nestLp} may be called Bayesian, but 2) other examples such as Program~\ref{listing:ProbProgramCoinEmulation} may not. In any case, our examples (and other papers in the field) demonstrate that probabilistic programming languages embrace even the former permissive interpretation of the word Bayesian.}  makes probabilistic programming a useful
infrastructure for
various statistical inference tasks.
It poses a challenge to the programming language
community, too, namely to come up with statistical inference techniques that
are efficient, generic, and language-based.

\begin{figure}[tbp]
 \footnotesize 
\centering
    \begin{minipage}[t]{.4\textwidth}
      \footnotesize

 \begin{lstlisting}[%numbers=right,
 basicstyle={\scriptsize\ttfamily},escapechar=|,caption={\texttt{coin}},label={listing:ProbProgramCoinEmulation}]
 bool c1, c2 := true;
 ifp (0.36)
  then c1 := true;
  else c1 := false;
 ifp (0.36)
  then c2 := true;
  else c2 := false;
 observe(!(c1 = c2));|\label{line:FairCoinEmuOb}|
 return(c1);
 \end{lstlisting}
 \end{minipage}
 \quad
    \begin{minipage}[t]{.4\textwidth}
      \footnotesize 

 \begin{lstlisting}[numbers=right,basicstyle={\scriptsize\ttfamily},escapechar=|,caption={\texttt{obsLoop}},label={listing:obsInLoop}]
 double x := 0;
 double y := 0;
 int n := 0;
 while (x < 3) {
  n := n + 1;
  y $\sim$ normal(1,1);
  observe(0 <= y <= 2);
  x := x + y;}
 observe(n >= 5);
 return(n);
 \end{lstlisting}
 \end{minipage}

\end{figure}

The community's efforts have produced a number of languages and  inference frameworks. Many of them are sampling-based (as opposed to symbolic and exact): Anglican~\citep{TolpinMYW16},
Venture~\citep{MansinghkaSP14},
Stan~\citep{CarpenterGHLGBBGLR17}, and Pyro~\citep{DBLP:journals/jmlr/BinghamCJOPKSSH19}.
% ; others feature  symbolic and exact inference,
% such as Hakaru~\citep{ShanR17}, PSI~\citep{GehrMV16} and EfProb~\citep{ChoJ17}. 
Recent topics include easing description of inference algorithms 
%(Gen from~\cite{DBLP:conf/pldi/Cusumano-Towner19}), 
\citep{DBLP:conf/pldi/Cusumano-Towner19},
assistance by DNNs~\citep{pmlr-v54-le17a}, and proximity to general-purpose languages~\citep{DBLP:conf/oopsla/Tolpin19}.
% and use of variational inference~\citep{zhang2018advances}.

\begin{auxproof}
 The community's efforts have produced a number of languages and  inference frameworks. Most practical ones are sampling-based, such as Anglican~\citep{TolpinMYW16},
 Venture~\citep{MansinghkaSP14},
 Stan~\citep{CarpenterGHLGBBGLR17}, and Pyro~\citep{DBLP:journals/jmlr/BinghamCJOPKSSH19}; others feature  symbolic and exact inference,
 such as Hakaru~\citep{ShanR17}, PSI~\citep{GehrMV16} and EfProb~\citep{ChoJ17}. Recent topics include flexibility of a framework~\citep{DBLP:conf/pldi/Cusumano-Towner19}, assistance by DNNs~\citep{pmlr-v54-le17a}, and use of variational inference~\citep{DBLP:journals/jmlr/BinghamCJOPKSSH19}.
\end{auxproof}

\myparagraph{Challenges in Inference on Probabilistic Programs} 
% Exact  inference methods apply mainly to simple models with symbolically expressible  posteriors. 
% Therefore, 
In this paper, we pursue sampling-based approximate inference of probabilistic programs. In doing so, we
 encounter the following two major challenges. 

The first  is  \emph{weight collapse}, a challenge  widely known in the community (see e.g.~\cite{Tsay18}). 
% It happens especially when observations are rare events.
In sampling a statistical model, one typically sweeps it  with a number of particles (i.e.\ potential samples). However, in case the model imposes rare observations, the particles' weights quickly decay to zero, making the  effective sample number tiny. 
Study of weight collapse  has resulted in a number of advanced sampling methods. They include
%. Two major classes of such methods are
\emph{sequential Monte Carlo (SMC)} featuring \emph{resampling}, and
 \emph{Markov chain Monte Carlo (MCMC)}.
% SMC features \emph{resampling} in the course of a sweep; MCMC collects samples through a random walk suitably defined with proposals and acceptance probabilities. 

The second challenge, specific to probabilistic programs, is the compatibility
between  advanced sampling methods (such as SMC and MCMC) and \emph{control structures}
of probabilistic programs. There are a number of sampling anomalies resulting from control structures (see e.g.~\cite{HurNRS15}): MCMC walks often find difficulties in traversing different control flows; for gradient-based sampling methods (such as Hamiltonian MC and variational inference), discontinuities due to different control flows emerge as a major burden.
See~\cite{Kiselyov16,ZhouYTR20} and~\cite[Section~3.4.2]{DBLP:journals/corr/abs-1809-10756} for further discussions.
% For example, \cite{HurNRS15} observed that the implementations
% of Stan and R2---which are both based on the MCMC
% algorithm in~\cite{WingateSG11}---can yield incorrect sampling results.
% The  issue here is the difficulty for MCMC walks to \emph{traverse
% different control flows}.
% See~\cite{Kiselyov16,ZhouYTR20} for further discussions. 

% This last challenge occurs also in the state-of-the-art  framework Anglican~\citep{TolpinMYW16}.
% Anglican offers a number of sampling methods (MCMC, SMC, etc.), but some
% advanced ones  disallow ``non-global'' observations. This roughly means the number of observations must not depend on control flows---Program~\ref{listing:obsInLoop} is disallowed since there is an observation in the while loop. Discontinuities arising from different control flows make it hard to employ gradient-based sampling methods, too (such as Hamiltonian MC and variational inference). See e.g.~\cite[Section~3.4.2]{DBLP:journals/corr/abs-1809-10756}. 

% \begin{wrapfigure}[9]{r}{0pt}
\begin{figure}[tbp]
 %---------------------------------------------
\centering
% \parbox{.45\textwidth}{
\scalebox{.85}{\begin{math}
 \entrymodifiers={++[F]}
 \vcenter{\xymatrix@R+.5em@C=1em{
   {\txt{Multi-armed \textbf{sampling of control flows}}}
            \ar@/^/[d]^-{\txt{a control flow $\vec{l}$}}
  \\
   {\txt{\textbf{Sampling of data} along
   the\\ straight-line program induced by \flowl\\ (simplified by \emph{condition propagation})}}
            \ar@/^/[u]^-{\txt{estimated\\ likelihood of $\flowl$}\;}
            \ar[r]
   &
   *{\txt{accumulate\\ data\\ samples}}
 }}
 \end{math}
} \caption{our hierarchical sampling architecture. It features control-data separation and logical condition propagation}
 \label{fig:hierarchicalSamplerArch}
% }

%\vspace{-2em}
\end{figure}
% \end{wrapfigure}

\myparagraph{Proposed Sampling Framework: Combining Two Ideas}
In order to alleviate the above two challenges
 (weight collapse and compatibility between sampling and control), 
 this paper proposes a hierarchical sampling framework shown in Figure~\ref{fig:hierarchicalSamplerArch}. It combines two recent ideas for efficient inference on probabilistic programs, namely \emph{control-data separation}~\cite{ZhouYTR20} 
and \emph{logical condition propagation}~\cite{NoriHRS14}. We  combine  the two ideas in a nontrivial manner so that
 they boost each other's benefits.

\myparagraph{Idea 1: Hierarchical Sampler via Control-Data Separation}
One feature of the proposed framework
% (Figure~\ref{fig:hierarchicalSamplerArch}) 
is the separation of control flow sampling and data sampling. In Figure~\ref{fig:hierarchicalSamplerArch}, the top level sampler chooses a specific control flow $\flowl$, which is passed to the bottom level data sampler. The data sampler then focuses on the \emph{straight-line program} that arises from the  control flow $\flowl$; it is thus freed from the duty of dealing with control flows
 such as if branches and while loops. 

%  and the bottom one is for data values. The bottom level focuses on a specific control flow \flowl\ that is chosen in the top level. This way we free the
%  data value sampling from the duty of dealing with different control
% flows. 

In our framework, 
the top-level sampling  is identified as what we call the \emph{infinite-armed sampling} problem: from a (potentially infinite) set of control flows, we aim to draw samples  according to their likelihoods, but the likelihoods are unknown and we learn them as the sampling goes on. This problem is  a variation of the classic problem of multi-armed bandit (MAB), with the differences that 1) we sample rather than optimize, and 2) we have potentially infinitely many arms. We show how the well-known $\varepsilon$-greedy algorithm for MAB can be adapted for our purpose. 

The bottom-level data sampling can be  by various algorithms. We use SMC in this paper.
This is because SMC---much like other importance sampling and particle filter methods, but unlike MCMC---can estimate the likelihood of a control flow \raisebox{0ex}[0ex][0ex]{$\vec{l}$} via the average weight of particles. This estimated likelihood is used in the top-level  flow sampling (Figure~\ref{fig:hierarchicalSamplerArch}). See Section~\ref{subsubsec:whySMC} for discussion.
% , for its ability of 
% estimating the likelihood of a control flow $\vec{l}$ at the same time; but the use of other algorithms shall be investigated in future work.
% via the average weight of particles
%, ,  and rely on  Anglican~\citep{TolpinMYW16} for its implementation. 
% This is because SMC, unlike MCMC, can estimate the likelihood of a control flow \raisebox{0ex}[0ex][0ex]{$\vec{l}$} via the average weight of particles. This estimated likelihood is used in the top-level  flow sampling (Figure~\ref{fig:hierarchicalSamplerArch}).
Use of other  algorithms for the bottom-level data sampling is future work. We expect that the control-data separation will ease the use of gradient-based algorithms (such as Hamiltonian MC and variational inference), since often each control flow denotes a differentiable function.

The idea of control-data separation appears in the recent work~\cite{ZhouYTR20}, where it is pursued under the slogan ``divide, conquer, and combine.'' Our hierarchical framework (Figure~\ref{fig:hierarchicalSamplerArch}) has a lot in common with the one in the work. At the same time, our framework---we started its development before the publication  of~\cite{ZhouYTR20}---is designed as a \emph{language-based} and \emph{static analysis-oriented} realization of the idea of control-data separation. This is in contrast with \cite{ZhouYTR20} that takes an alternative, more statistics-oriented approach (see~\cite[Section~6.2]{ZhouYTR20}).

As shown in Figure~\ref{fig:hierarchicalSamplerArch}, the two levels of sampling are interleaved in our framework, accumulating data samples in each iteration of the bottom-level sampling. Those data samples are suitably weighted so that the resulting data samples are unbiased. We note that our hierarchical framework does not aim to make the sampled distribution smooth---usually a probabilistic program \emph{does} denote a distribution that is not smooth. Instead, it tries to make the most of partial smoothness that is inherent in the target probabilistic program, by letting a data sampling algorithm focus on a single control flow which is more likely to be smooth within.

 Specifically, major differences between~\cite{ZhouYTR20} and ours are in 1) how we find new control flows (Remark~\ref{rem:ZhouEtAlFlowDiscovery}); 2) how control flows are  sampled (Remark~\ref{rem:ZhouEtAlFlowSampling}); and, most notably, 3) the use of logical condition propagation (below).

\myparagraph{Idea 2: Logical Condition Propagation, Domain Restriction, and Blacklisting}
In this paper, we find an advantage of control-data separation in easing the application of   \emph{logical reasoning} for sampling efficiency. Specifically, we combine \emph{condition propagation}---an idea originally introduced in R2~\citep{NoriHRS14}---in the hierarchical framework in Figure~\ref{fig:hierarchicalSamplerArch}.
% \footnote{Gen~\cite{DBLP:conf/pldi/Cusumano-Towner19} has what is called \emph{Static Modeling Language}. Their ``static analysis'' is different from logical condition propagation in this paper and in R2~\citep{NoriHRS14}: ...}
 It logically propagates observations upwards in a program, so that sample rejection    happens earlier and unnecessary samples are spared.

Condition propagation is much like  a weakest
precondition calculus, a well-known technique in program verification (see e.g.~\cite{Winskel93}). The original condition propagation in R2 is targeted at arbitrary programs,  and this resulted in limited applicability. For example, it requires explicit loop invariants for condition propagation over while loops, but loop  invariants are hard to find. 
In contrast, in our framework (Figure~\ref{fig:hierarchicalSamplerArch}), condition propagation is only applied to straight-line programs (without if branchings or while loops), making the application of condition propagation easier and more robust. 

We also introduce two important derivatives of condition propagation. One is \emph{domain restriction} (Section~\ref{subsec:conditionPropag}): we restrict the domain of distributions to sample from, using the conditions obtained by condition propagation. The resulting distribution prohibits samples that will anyway be rejected. 

The other is \emph{logical blacklisting of control flows} (Section~\ref{subsec:mainAlgorithm}). Condition propagation often reveals that a control flow $\flowl$ is logically infeasible and  thus has  zero likelihood.  This information is passed upwards in Figure~\ref{fig:hierarchicalSamplerArch}, and we blacklist the flow $\flowl$ in the  henceforth sampling. This use of condition propagation is unique to its combination with control-data separation (Figure~\ref{fig:hierarchicalSamplerArch}); its 
 effect is experimentally verified in Section~\ref{sec:implemenationAndExperiments}.
% ---makes its combination with control-data separation even more appealing.
% ; its effect is experimentally verified (Section~\ref{sec:implemenationAndExperiments}).
%Sampling from domain-restricted distributions is easy, too, via the inverse transform sampling. 

% \myparagraph{Comparison with~\cite{ZhouYTR20}} 

There are other probabilistic programming frameworks that use static analysis.
For example, Gen~\citep{DBLP:conf/pldi/Cusumano-Towner19} applies static analysis  to its fragment called the \emph{static modeling language}; Birch~\citep{MURRAY201829} uses static reasoning to delay sampling and enable analytical
optimizations. The difference is that their use of static analysis is essentially limited to straight-line programs, which is in contrast with the current work where condition propagation  applies to \emph{every} program via control-data separation. 

In~\cite{OlmedoGJKKM18} a program transformation called \emph{hoisting} is introduced. Inspired by R2~\citep{NoriHRS14}, the transformation eliminates all observation commands in a program by propagating them backwards. Such elimination of observations is possible in~\cite{OlmedoGJKKM18} since their target language has randomization only in the form of probabilistic branching (such as $\{x\coloneqq 0\}[1/2]\{x\coloneqq 1\}$ for a fair coin). While probabilistic assignments from \emph{discrete} distributions may be emulated by probabilistic branching, the language in~\cite{OlmedoGJKKM18} does not allow probabilistic assignments from \emph{continuous} distributions, such as {\ttfamily y $\sim$ normal(1,1)} in Program~\ref{listing:obsInLoop}. Due to the presence of the latter, our condition propagation does not totally eliminate observations in general.

\myparagraph{Contributions and Organization} Our main contribution is a hierarchical sampling algorithm (Figure~\ref{fig:hierarchicalSamplerArch}) that combines the ideas of control-data separation (one used in~\cite{ZhouYTR20}) and condition propagation~\citep{NoriHRS14}. It is mathematically derived in Section~\ref{sec:algorithm}. This is preceded by Section~\ref{sec:theLang}, where we  introduce the syntax and the semantics of our target language $\PIMP$, as well as  condition propagation.

 Our other theoretical contributions  include the formulation of the infinite-armed sampling (IAS) problem, its $\varepsilon$-greedy algorithm, and 
 % the mathematical derivation of 
 % the sampler (Section~\ref{sec:algorithm}) and 
a convergence proof for the algorithm (restricting to finite arms, Section~\ref{subsec:multiArmedSamplingForControlFlows}). Our hierarchical sampler is introduced as a refinement of this IAS algorithm (Section~\ref{subsec:mainAlgorithm}).

Our implementation  is built on top of Anglican~\citep{TolpinMYW16} and is  called $\Schism$.
 % (for SCalable HIerarchical
 % SaMpling). 
In our experimental comparison with Anglican in Section~\ref{sec:implemenationAndExperiments}, we witness $\Schism$'s performance advantages, 
% successfully addresses the challenge of compatibility between sampling and control.
%   Its performance advantage is
 especially with programs  with while loops and rare observations. We discuss potential application in the domain of testing automotive systems.
% For those programs, condition propagation successfully rules out many control flows, allowing the sampler to focus on those flows whose  prior likelihood is small but posterior likelihood is large.

% We observe that the former performs better for programs that have control structures and observations intertwined. 
% We present experiment results, in which we compare the performance of $\Schism$ and Anglican~\citep{TolpinMYW16}.  
% \todo[inline]{update the following sentence. Summarize the experiment results. }
% A few benchmarks  demonstrate our comparative performance advantage; it is mainly due to the compatibility problem between MCMC and control flows in Anglican, and the benefit of condition propagation in $\Schism$. 

Many details are deferred to the appendix.
%, provided in a supplementary material.

% \myparagraph{Organization}
% In Section~\ref{sec:theLang} we introduce the syntax and semantics of our target language $\PIMP$. 
% % In particular, we introduce \emph{probabilistic control flow graphs} (pCFGs), a graph presentation of $\PIMP$ programs.
%  Condition propagation
% % for straight-line programs
%  is introduced there, too. 
% % In Section~\ref{subsec:derivation} we mathematically justify our control-data separation, and
% We introduce the infinite-armed sampling problem  in Section~\ref{subsec:multiArmedSamplingForControlFlows}, and present  our main algorithm as its refinement in Section~\ref{subsec:mainAlgorithm}. In Section~\ref{sec:implemenationAndExperiments} we discuss our implementation and experiments. 
% Many details are deferred to the appendix, provided in a supplementary material.

\myparagraph{Notations}
The sets of real numbers and nonnegative reals are $\R$ and $\Rnn$, respectively.
% We regard them as measurable spaces equipped with the standard Borel $\sigma$-algebras.
% . The set of  nonnegative reals, augmehted with  $\infty$, is denoted by $[0,\infty]$.
% This is not used
We 
%follow the statistics convention to
 write $x_{1:N}$ for a sequence $x_{1},x_{2},\dotsc,x_{N}$. 
% It extends to higher dimensions: $x_{1:N}^{1:M}$ is for $x_{1}^{1},\dotsc,x_{N}^{1},\dotsc,x_{1}^{M},\dotsc,x_{N}^{M}$  (length $MN$). 
% For a measurable space $X$,  the set of probability distributions over $X$ is denoted by $\dist(X)$. The set of  \emph{subprobability distributions} $\mu$ over $X$---for which $\mu(X)$ is required to be $\le 1$ rather than $=1$---is denoted by $\subdist(X)$. 
% We equip both $\dist(X)$ and $\subdist(X)$
% with suitable $\sigma$-algebras given by  \cite{Giry82}.
% For a  distribution $\mu$, its support is denoted by $\mathsf{supp}(\mu)$. 
% The Dirac distribution at $x\in X$ is denoted by $\delta_{x}$. 

%Given a Boolean formula $\varphi$, its characteristic function $\mathbf{1}_{\varphi}$ is defined as follows:
% \begin{displaymath}
%  \mathbf{1}_{\varphi}(x)
% \;=\;
%  \begin{cases}
%   1 &\text{if $x$ satisfies $\varphi$}
%  \\
%   0 &\text{otherwise}
%  \end{cases}
% \end{displaymath}
% $  \mathbf{1}_{\varphi}(x)
%  = 1$
% if $x$ satisfies $\varphi$; and 
% $0$ otherwise.

% The function $\last$ takes a nonempty list and returns its last element, that is, $\last(l_{1}l_{2}\dotsc l_{n})=l_{n}$. 

We use the following notation for Lebesgue integrals. Let $(X,\mu)$ be a measure space, 
% $U\subseteq X$ be a measurable subset, $\mu$ be a measure over $X$, 
and $f\colon X\to \R$ be a measurable function.  We write
\begin{equation}\label{eq:intNota}
\textstyle \int_{x\in X} f(x) \, \mu(\mathrm{d}x)
\end{equation}
for the integration of $f$ over $(X,\mu)$. This is almost the standard notation 
\begin{math}
 \int_{X} f(x) \, \mu(\mathrm{d}x)
\end{math}, where $\mathrm{d}x$ is understood as a measurable subset around $x$ (thus $x$ and $\mathrm{d}x$ are tied with each other). In our notation~(\ref{eq:intNota}), we write $x\in X$ to make it explicit that $x$ is a bound variable that ranges over $X$. 

This notation is extended as follows, which might be found a bit  more unconventional. When the measure space in question is of the form $(Y\times Z,\mu)$, 
%i.e.\ when the underlying measurable space $X=Y\times Z$ is the product of two, then
 we write
\begin{displaymath}
\textstyle \int_{y\in Y, z\in Z} f(y, z) \, \mu(\mathrm{d}y\times\mathrm{d}z)
\end{displaymath}
for the integration. Note here that the rectangles $\mathrm{d}y\times\mathrm{d}z$ generate the $\sigma$-algebra of the product measurable space $Y\times Z$.

\section{The Language $\PIMP$ and  {pCFGs}}\label{sec:theLang}

\subsection{Probabilistic Programming Language $\PIMP$}\label{subsec:PIMP}
We use an imperative  probabilistic programming language  whose syntax (Figure~\ref{fig:ProbSyntax}) closely follows~\cite{GordonHNR14}. The language is called $\PIMP$;
%We therefore use the same name $\PIMP$.
 Program~\ref{listing:ProbProgramCoinEmulation} \&~\ref{listing:obsInLoop} are examples.
% Definition~\ref{def:ProbLanguage}. 
 $\PIMP$ features the following probabilistic constructs.

 The \emph{probabilistic assignment} command $x\sim \Dist(\paramDist)$ samples a value from the designated distribution and assigns it to the variable $x$. 
 %where $x$ is a variable, 
 Here $\Dist$ specifies a family of probability distributions (such as normal, Bernoulli,
 % binomial,
 %exponential, Poisson, 
 etc.; they can be discrete or continuous), and $\paramDist$ is a vector of parameters for the family $\Dist$, given by expressions for real numbers. An example is $x\sim \mathtt{normal}(0,1)$.
 %where the value of  $x$ is drawn from the normal distribution with mean $0$ and variance $1$. 

 The \emph{probabilistic branching} $\ifpClause{p}{c_{1}}{c_{2}}$
% has a real number   $p\in [0,1]$, and 
chooses one from  $c_{1}$ and $c_{2}$ with the probabilities $p\in [0,1]$ and $1-p$, respectively. We introduce this as a shorthand for $b\sim\mathtt{Bernoulli}(p); \ifClause{b}{c_{1}}{c_{2}}$, where $b$ is a fresh Boolean variable.

The \emph{soft conditioning} command  $\weightClause\fuzzyPred$ is a primitive in $\PIMP$, as is common in probabilistic programming languages.  Here, $\fuzzyPred$ is a \emph{fuzzy predicate}---a function that returns a nonnegative real number---that tells how much weight  the current execution should acquire. 

We also use \emph{sharp conditioning} $\observeClause\sharpPred$---where $\sharpPred$ is a Boolean formula rather than a fuzzy predicate---in our examples (see e.g.\ Program~\ref{listing:ProbProgramCoinEmulation} \&~\ref{listing:obsInLoop}). This is a shorthand for $\weightClause\mathbf{1}_{\varphi}$, where $\mathbf{1}_{\varphi}$ is the characteristic function of the formula $\varphi$ ($\mathbf{1}_{\varphi}$ returns $1$ if $\varphi$ is true, and $0$ if $\varphi$ is false). 
% We keep $\observeClause\sharpPred$ as a primitive, however, for the convenience in condition propagation (Section~\ref{subsec:conditionPropag}). 

% Given a Boolean formula $\sharpPred$, the \emph{observation} command $\observeClause\sharpPred$
% %---where $\fuzzyPred$ is a predicate---
% conditions the current distribution over memory states to those which satisfy $\sharpPred$. Such \emph{conditioning}, also called \emph{evidence} in the literature, is central to the Bayesian aspect of probabilistic programming languages. 

As usual, in Figure~\ref{fig:ProbSyntax},
 only (sharp) Boolean formulas $\varphi$ are allowed as guards of if branchings and while loops.

% Thirdly, we have \emph{probabilistic branching} $\ifpClause{p}{c_{1}}{c_{2}}$, where $p\in [0,1]$ is a real number and $c_{1},c_{2}$ are commands. It chooses one $c_{1}$ and $c_{2}$ with the probabilities $p$ and $1-p$, respectively. 

% In a command $\observeClause\fuzzyPred$, we allow a \emph{fuzzy predicate} as $f$, which takes nonnegative real numbers as truth values (scores, weights). The use of such \emph{soft conditioning}  in statistical modeling is widely recognized. Moreover,  soft conditioning is crucial in our \emph{domain restriction} technique (Section~\ref{subsec:conditionPropag}). Note, however, that
% we allow only (sharp) Boolean formulas $\varphi$ as guards of if branchings and while loops.

\begin{auxproof}
 
 \begin{myremark}\label{rem:softConstraint}
  In addition to Boolean formulas,  we allow \emph{fuzzy predicates}---i.e.\ functions  to nonnegative real values---as a formula $\fuzzyPred$ in $\observeClause{\fuzzyPred}$. This realizes so-called \emph{soft constraints}, whose advantage in modeling and efficient sampling is widely acknowledged.\footnote{For example, differentiable fuzzy predicates are amenable to 
 %opens up a possibility of using 
 numeric optimization methods such as gradient descent. 
 %This is not possible with a sharp, Boolean predicate that takes values in $\{\true,\false\}$.
 } While soft constraints are absent in the language in~\cite{GordonHNR14,NoriHRS14}, they are featured by other recent probabilistic programming languages such as Anglican and Venture. 

 We note that we only allow (sharp) Boolean formulas $\varphi$ as guards of if branchings and while loops, unlike in $\observeClause{\fuzzyPred}$.
 \end{myremark}
\end{auxproof}

\begin{auxproof}
 
 \begin{myremark}\label{rem:ifp}
 The construct $(\ifpClause{p}{c_{1}}{c_{2}})$  can also be thought of as a syntactic shorthand for $\bigl( \mathtt{coin}\sim\mathtt{Bernoulli}(p);\;
  \ifClause{\mathtt{coin}}{c_{1}}{c_{2}} \bigr)$, where $\mathtt{coin}$ is a fresh Boolean variable. 
 %We include the $\mathtt{ifp}$ construct for the ease in presenting examples.
 % It is obvious that
 % $ \ifpClause{p}{c_{1}}{c_{2}}$ 
 % can be encoded 
 % by $\bigl( \mathtt{coin}\sim\mathtt{Bernoulli}(p);\;
 %   \ifClause{\mathtt{coin}}{c_{1}}{c_{2}} \bigr)$,
 %  \begin{align*}\label{eq:ifpByBernoulli}
 %  % \ifpClause{p}{c_{1}}{c_{2}}
 %  % \quad\text{by}\quad
 % % \bigl(\;
 %  \mathtt{coin}\sim\mathtt{Bernoulli}(p);\quad
 %  \ifClause{\mathtt{coin}}{c_{1}}{c_{2}} \enspace,
 % %\;\bigr)\enspace,
 %  \end{align*}
 % where $\mathtt{coin}$ is a fresh Boolean variable. 
 % Thie is because the two programs are different in view of our current sampling paradigm of control-data separation
 % We expect, however, that these two programs behave differently in our sampling framework that features the separation of control flow from data.
 %  We  therefore take the $\mathtt{ifp}$ construct as a primitive.

 (Ichiro) I was about to write that the two programs differ in view of our current sampling paradigm of control-data separation. However, now that we don't use MCMC any more, they are much the same.
 \end{myremark}
\end{auxproof}

% \begin{myremark}
%  Our current implementation builds on Anglican~\cite{TolpinMYW16}, and hence on Clojure; therefore we take the Clojure operators as operator primitives. 
% \end{myremark}

% \begin{mydefinition}[$\PIMP$, syntax]\label{def:ProbLanguage}
% The syntax of our  language $\PIMP$ is in Figure~\ref{fig:ProbSyntax}. We fix a countable set $\Var$ of variables. 

% \end{mydefinition}

\begin{figure*}[tbp]
% \!\!\!\!\!\!\!\!\!\!\!\!\!\!\!
\centering
\scalebox{.7}{ 
 \begin{math}
\begin{array}{lcllll}
  &x\;&\in & \Var 
 \\
 &\mathbf{c}\;&::=& \multicolumn{3}{l}{%\cdots \qquad
  \text{constants, e.g.\ $0$, $-1.2$ and $\true$}}
 \\
 &\uop\;&::=& \multicolumn{3}{l}{%\cdots \qquad
 \parbox[t]{.4\linewidth}{
 unary operators, e.g.\ ``$-$'' in $-3$
 %, negation $!$, etc.
 }}
 \\
 &\bop\;&::=& \multicolumn{3}{l}{%\cdots \qquad
    \parbox[t]{.4\linewidth}{binary operations, e.g.\ $+$, $\&\&$}}
 \\
 &\varphi\;&::=& \multicolumn{3}{l}{
 %\cdots \qquad 
 % \parbox[t]{.6\linewidth}{
 % fuzzy predicates, i.e.\ expressions that take nonnegative real values.
 % Fuzzy predicates encompass Boolean formulas by identifying $1$ with $\true$ and $0$ with  $\false$.
 % }
 \text{Boolean formulas}
 }
 \\
 &\fuzzyPred\;&::=& \multicolumn{3}{l}{%\cdots \qquad 
 \parbox[t]{.4\linewidth}{
 fuzzy predicates, i.e.\\\  expressions that return nonnegative real values
 % They encompass Boolean formulas: $1$ is $\true$ and $0$ is  $\false$.
 }}
 %\text{fuzzy predicates, i.e.\ expressions that take nonnegative real values. }}
 %\\
 % &\;&& \multicolumn{3}{l}{\phantom{\cdots} \qquad \text{Fuzzy predicates encompass Boolean formulas by identifying $1$ with $\true$ and $0$ with  $\false$.  }}
 % \\
 % &\;&& \multicolumn{3}{l}{\phantom{\cdots} \qquad \text{Fuzzy predicates encompass Boolean formulas by identifying $1$ with $\true$ and $0$ with  $\false$.  }}
 %\\
 \\[+.5em]
 % & x\;&\in && \Var
 % \\
 % &\uop\;&::=&& \cdots \qquad \text{Clojure unary operations, such as ``$-$'' in $-3$}
 % \\
 % &\bop\;&::=&& \cdots \qquad \text{Clojure binary operations, such as $+$ and $\cdot$}
 % \\
% \Exp\ni
 & e\; & ::= 
% &\text{expressions}
%   \\
  &x\mid \mathbf{c}\mid
  e_{1}\mathbin{\bop} e_{2} \mid
  \uop e &
 % \parbox[t]{.2\linewidth}{expressions ($\mathtt{int}$, $\mathtt{bool}$, $\mathtt{double}$, etc.)}
 \\
 &&& \multicolumn{3}{l}{\parbox[t]{.3\linewidth}{value expressions\\ ($\mathtt{int}$, $\mathtt{bool}$, $\mathtt{double}$, etc.)}}
 % \\
 % &&
 %  &&\text{expressions}
 %  \\[+.1em]
 %  &\multicolumn{4}{l}{
 %  \text{where $x\in\Var$, $r$ is a constant for each  $r\in\R$,
 % % and $\aop\in \{+,-,\cdot,/,{}^{\wedge}\}$
 % }}
 % \\
 % &\multicolumn{4}{l}{
 %  \text{$\aop\in \{+,-,\times, {}^{\wedge}}$, and $\dt$ is a constant (for ``infinitesimal'')}}
 % \\[+.3em]
 % \Fml\ni& \varphi & ::= &
 %  \multicolumn{2}{l}{\true \mid \false \mid   \varphi_{1}\mathbin{\mathtt{\&\&}} \varphi_{2}
 %   \mid \mathop{\mathtt{!}} \varphi \mid   a_{1}< a_{2}}
 % \\&&&&
 %  \text{logical formulas}
 %  \\[+.1em]
 %  &&&
 % \multicolumn{2}{l}{\text{(Note that $a_{1}\le a_{2}$ and $a_{1}= a_{2}$ can be encoded)}}
 % \text{where
 % $\bop\in \{\land,\lor\}$ and $\rop\in\{<,>\}$}}
 % \\[+.3em]
% \\[+.7em]
\end{array}
% \hspace*{-30pt}
%---------------------------------------------------------------
\begin{array}{lcllll}
% \Cmd\ni
 & c & ::= &
 &\text{commands}
 \\
 &&&
   \skipCmd 
   &\text{skip}
 \\
 &&&
 \mid  x := e
   &\text{deterministic assignment}
 \\
 &&&
 \mid  x \sim \Dist(\paramDist)
   &\text{probabilistic assignment}
 % \\
 % &&&
 % \mid  \observeClause{\sharpPred}
 %   &\parbox[t]{.4\textwidth}{sharp conditioning by a Boolean formula}
 \\
 &&&
 \mid  \weightClause{\fuzzyPred}
   &\parbox[t]{.3\textwidth}{soft conditioning by a fuzzy predicate}
 \\%[+.3em]
 &&&
 \mid  c_{1};\,c_{2}
   &\text{sequential composition}
 \\ 
 &&&
 \mid  \ifClause{\varphi}{c_{1}}{c_{2}}
   &\text{conditional branching}
 % \\ 
 % &&&
 % \mid  \ifpClause{p}{c_{1}}{c_{2}}
 %   &\parbox[t]{.35\textwidth}{probabilistic branching ($p\in [0,1]$)}
 \\ 
 &&&
 \mid  \whileClause{\varphi}{c}
   &\text{while loop}
 \\[+.7em]
% \mathbf{Program}\ni
 & \pi & ::= &
 c\,;\; \returnClause{e}
 &\text{program}
 \end{array}
\end{math}}
\caption{the probabilistic programming language $\PIMP$, syntax. The set $\Var$ of variables is fixed. The $\mathtt{ifp}$ construct in Program~\ref{listing:ProbProgramCoinEmulation} is a shorthand; see Section~\ref{subsec:PIMP}.
% Our current implementation builds on Anglican~\citep{TolpinMYW16}, which is implemented in Clojure; therefore we take the Clojure operators as operator primitives. 
}
\label{fig:ProbSyntax}
\end{figure*}

%  \begin{figure}[tbp]
% %---------------------------------------------------------
% \hfill
% \begin{minipage}{.45\textwidth}
%  \footnotesize \centering
%  \begin{lstlisting}[numbers=left,basicstyle={\scriptsize\ttfamily}]
% double x := 0;
% double z := 0;
% int count := 0;
% while (x < 3) {
%   count := count + 1;
%   z $\sim$ normal(1,1);
%   x := x + z;
%   observe(0 <= z <= 2) }
% observe(count >= 5);
% return(count)
% \end{lstlisting}
% \caption{A $\PIMP$ program with a loop and observations}
% \label{listing:obsInLoop}
% \end{minipage}
% \end{figure}

\subsection{Probabilistic Control Flow Graphs (pCFGs)}\label{subsec:pcfg}
We use the notion of \emph{probabilistic control flow graph (pCFG)}, adapted from~\citep{AgrawalCN18},  for presenting $\PIMP$ programs as graphs. It is a natural probabilistic variation of \emph{control flow graphs} for imperative programs.

\begin{wrapfigure}[8]{r}{0pt}
% \begin{minipage}{.6\textwidth}
\scalebox{.6}{
  \begin{math}
   \scriptstyle
   \entrymodifiers={++[F]}
   \vcenter{\xymatrix@C+1.8em@R+1em{
   *{}
   &
   *{}
      \ar[r]^-{\small\begin{array}{l}
              x:=0, y:=0\\
	      n:=0
	      \end{array}}
   &
   {\linit}
      \ar[r]^-{x<3}
      \ar[d]_-{\mathop{!}(x<3)}
   &
   {l^{(2)}}
      \ar[r]^-{\small\begin{array}{l}
              n:=\\
	      n+1
	      \end{array}}
   &
   {l^{(3)}}
      \ar[d]|{\small\begin{array}{l}
              y\sim\\
	      \mathtt{normal}(1,1)
	      \end{array}}
   \\
   *{}
   &
   {\lfinal}
      \ar[l]^-{n}
   &
   {l^{(6)}}
      \ar[l]^-{\small\begin{array}{l}
              \mathtt{observe}
              \\
	      n \ge 5
	      \end{array}}
   &
   {l^{(5)}}
      \ar[lu]|{\small\begin{array}{l}
              x:=
              \\
	      x+y
	      \end{array}}
   &
   {l^{(4)}}
      \ar[l]^-{\small\begin{array}{l}
              \mathtt{observe}
              \\
	      0\le y \le 2
	      \end{array}}
   }}
  \end{math}
}
\hspace{-1.5em}
\\

\vspace*{-1.5em}

\caption{pCFG for \texttt{obsLoop} (Program~\ref{listing:obsInLoop}).
  % here $\linit\in L_{\mathsf{D}}$, $l^{(3)}\in L_{\mathsf{PA}}$,
  % $l^{(2)},l^{(4)}\in L_{\mathsf{DA}}$, and
  % $l^{(5)}, l^{(6)}\in L_{\mathsf{O}}$. 
%  The label for the edge into the initial location $\linit$ denotes
% % $\siginit$ (
% the initial values of the variables; and the label for the edge out of $\lfinal$ specifies
% the return expression.
% % $\efinal$ (the value of which expression to be returned).
}
\label{fig:pCFGexample1}
% \end{minipage}
\end{wrapfigure}
% For space reasons, here we only show an example, deferring the  definition  to \ref{appendix:pCFG}. 
A pCFG is a finite graph,  its nodes roughly correspond to lines of a $\PIMP$ program, and its edges have transition labels that correspond to atomic commands of $\PIMP$. 
An example is in  Figure~\ref{fig:pCFGexample1}:
$\linit$ is the \emph{initial location}; the label into $\linit$ specifies the \emph{initial memory state} $\siginit$;  $\lfinal$ is the \emph{final location}; and the label out of $\lfinal$ ($n$ here) specifies the \emph{return expression} $\efinal$. 
% The depiction of the  pCFG location types (Figure~\ref{fig:typesOfLocations}), as well as an example in Figure~\ref{fig:pCFGexample1}, should convey the ideas. 
%Translation of imperative programs to control flow graphs is straightforward; this is also the case in our setting of $\PIMP$ programs and pCFGs. 

Our formal  definition (Definition~\ref{def:pCFG}) differs from~\cite{AgrawalCN18} mainly in the following: 1)  presence of  weight commands; and 2)  absence of nondeterminism. 

\begin{mydefinition}[pCFG, adapted from~\cite{AgrawalCN18}]\label{def:pCFG}
A \emph{probabilistic control flow graph (pCFG)} is a tuple 
\begin{math}
 \Gamma=(L,V,\linit,\siginit,\lfinal,\efinal,\trrel,\lambda)
\end{math}
that consists of the following components.
\begin{itemize}

\item A finite set $L$ of \emph{locations},  equipped with a partition $L=
% L_{\mathsf{P}}+
L_{\mathsf{D}}+L_{\mathsf{PA}}+L_{\mathsf{DA}}
%+L_{\mathsf{O}}
+
L_{\mathsf{W}}+
\{\lfinal\}$ 
into
% where 
% $L_N$, $L_P$, $L_D$ and $L_A$ are the sets of
% \emph{probabilistic}, 
\emph{deterministic}, \emph{probabilistic assignment}, \emph{deterministic assignment}, 
%\emph{observation}, 
\emph{weight} and \emph{final} locations.

%  \marginpar{Currently $V=\{x_1,\ldots,x_{|V|}\}$ and $V=\{1,\ldots,{|V|}\}$ are mixed. Ichiro thinks we should use $V=\{v_1,\ldots,v_{|V|}\}$ ($x_{j}$ overlaps with vectors). $\Rightarrow$ using $V=\{v_1,\ldots,v_{|V|}\}$ is OK \\
%  Toru: For now we would keep using the previous notation, as then the description like $\x = a_1v_1+\ldots+a_{|V|}v_{|V|}$ looks a bit unusual to me. If any, we would rather change $\real^V$ to $\real^{|V|}$. At the moment we can avoid explicitly describing a sequence $\x_1,\x_2,\ldots$ of vectors, which looks indeed confusing. }

\item A finite set $V=\{x_1,\ldots,x_{|V|}\}$ %\{v_1,\ldots,v_{|V|}\}$ 
of \emph{program variables}. It is a subset of the set $\Var$ of variables.
% (Definition~\ref{def:ProbLanguage}). 

\item An \emph{initial location} $\linit\in L$, and 
 an \emph{initial memory state} $\siginit\colon V\to \R$.

 \item A \emph{final location} $\lfinal\in L$, and 
% a \emph{return expression vector} $\efinal$ of expressions.
 a \emph{return expression} $\efinal$.

\item A \emph{transition relation} ${\trrel}\subseteq L\times L$.
%  which is total (each location has a successor).
% %We write $\trrel_N$ for ${\trrel}\cap
% %(L_N \times L)$.
% %We define $\trrel_P$, $\trrel_D$ and $\trrel_A$ similarly.
% %Moreover, 
% %for $l\in L$, we write $\trrel_{l}$ for $\trrel \cap (\{l\} \times L)$.
% for $l\in L\setminus L_A$, we write $\mysucc(l)$ to denote the set of all successors of $l$, i.e. $\mysucc(l) = \{l'\in L \mid l\trrel l'\}$ .
% We require that each assignment location $l \in L_A$ has a unique successor; in this case, $\mysucc(l)$ denotes this unique location.%\footnote{(Takisaka) Looking for less dirty expression.} 

\item A \emph{labeling function} $\lambda$.
\end{itemize}
A  labeling function $\lambda$ assigns to each transition $l\to l'$ a command, a formula or a real number. 
%The labeling function $\lambda$, together with the transition relation $\to$, 
It is subject to the following conditions. 
\begin{itemize}
 % \item  Each probabilistic location $l\in L_{\mathsf{P}}$, modeling the $\mathtt{ifp}$ branching, has two outgoing transitions. One is labeled with a real number $p\in [0,1]$; the other is  with $1-p$. 
 \item  Each deterministic location $l\in L_{\mathsf{D}}$ has two outgoing transitions. One is labeled with a (sharp) Boolean formula $\varphi$; the other is labeled with its negation $\mathop{!}\varphi$. 
 \item  Each probabilistic assignment location $l\in L_{\mathsf{PA}}$ has one outgoing transition. It is labeled with a probabilistic assignment command $x \sim \Dist(\paramDist)$. 
 \item  Each deterministic assignment location $l\in L_{\mathsf{DA}}$ has one outgoing transition. It is labeled with a deterministic assignment command $x \coloneqq e$. 
% \item  Each \emph{observation} location $l\in L_{\mathsf{O}}$ has one outgoing transition, labeled with a command $\observeClause{\sharpPred}$. 
\item  Each \emph{weight} location $l\in L_{\mathsf{W}}$ has one outgoing transition, labeled with a command $\weightClause{\fuzzyPred}$. 
We also use $\observeClause{\sharpPred}$ as a label from time to time. Recall that $\observeClause{\sharpPred}$ is a shorthand for $\weightClause{\mathbf{1}_{\sharpPred}}$, where $\mathbf{1}_{\sharpPred}$ is the characteristic function for the Boolean formula $\sharpPred$. 
 \item The final location $\lfinal$ has no successor with respect to $\to$. 
\end{itemize}
%We assume the following conditions.
%\begin{itemize}
%\item The transition relation is \emph{total}, i.e.\ $\trrel_l\neq\emptyset$ for each $l\in L$.
%\item The guards are total and mutually exclusive, i.e.\ for each $l\in L_D$, $\bigvee_{\tau\in \trrel_{D,l}}G(\tau)=\top$
%and $G(l,l') \land G(l,l'') = \bot$ for each $l' \neq l''$. 
%\item Each transition from $l \in L_A$ is uiquely determined, i.e. $\trrel_{l,A}$ is a singleton. We write $l_s (l)$ to denote this unique location.\footnote{(Takisaka) Looking for less dirty expression.} 
%\end{itemize}
%The pCFG $\Gamma$ is called \emph{linear} (resp.\ \emph{polynomial}) if 
%for each $l\in L_A$ such that $\myUp(l)=(v,u)$,
%the following conditions are satisfied:
%\begin{itemize}
%\item if $u$ is a measurable function $f:\mathbb{R}^V\to\mathbb{R}$, then there exists an affine (resp.\ polynomial) expression $\mathfrak{a}$ over $V$ s.t.\ 
%$f=\sem{\mathfrak{a}}$.
%
%\item if $u$ is a measurable set $R\subseteq\mathbb{R}$, then $R=\mathbb{R}$, or 
%there exist real numbers $r_{1,1},\ldots,r_{m,1},r_{1,2},\ldots,r_{m,2}\in\mathbb{R}$ such that 
%$R=[r_{1,1},r_{1,2}]\cup\cdots\cup[r_{m,1},r_{m,2}]$.
%%for 
%%(x_v\rhd r^l_{m,1})\wedge (-x_v\rhd -r^l_{m,2})
%%a linear predicate $\mathfrak{P}$ over $\{x_v\}$ 
%%such that $R=\{r\mid r\in\sem{\mathfrak{P}}\}$
%\end{itemize}

% Similarly, if $G$ is represented by linear predicates  (i.e. $G(l,l') = \sem{\mathfrak{P}(l,l')}$ for some linear predicate $\mathfrak{P}(l,l')$ for each $l'$) then we just write $\mathfrak{P}(l,l')$ to denote $G(l,l')$.
% for each $i \in \mathbb N$.
\end{mydefinition}
\noindent
In the last definition, we assume that the values of all the basic types ($\mathtt{int}$, $\mathtt{bool}$, $\mathtt{double}$, etc.) are embedded in $\R$. Using different value domains 
%for different basic types
 is straightforward but cumbersome. 

%---------------------------------------------------------
 \begin{figure*}[tbp]
\begin{minipage}{\textwidth}
 \centering
 % %---------------------------------------------
 % \subcaptionbox{\footnotesize
 %  A probabilistic location
 %  $l\in L_{\mathsf{P}}$. 
 % ($p\in [0,1]$)
 % %\label{cat}
 % }[.18\textwidth]
 % {
 % \scalebox{.7}{\begin{math}
 % \entrymodifiers={++[F]}
 % \vcenter{\xymatrix@R=-0em@C+1.5em{
 %  *{}
 %  &
 %   *{\;\cdots}
 %  \\
 %   {l}
 %          \ar[rd]_{1-p}
 %          \ar[ru]^{p}
 %  &*{}
 %  \\
 %  *{}
 %  &%{l''}
 %   *{\;\cdots}
 % }}
 % \end{math}
 % } }
 %---------------------------------------------
 \;
 \subcaptionbox{A deterministic location
 $l\in L_{\mathsf{D}}$. 
 %Here $\varphi$ is a logical formula
 %\label{cat}
 }[.18\textwidth]
 {
 \scalebox{.7}{\begin{math}
 \entrymodifiers={++[F]}
 \vcenter{\xymatrix@R=0em@C+1.5em{
  *{}
  &
 %   {l'}
   *{\;\cdots}
  \\
   {l}
          \ar[rd]_{\mathop{!}\varphi}
          \ar[ru]^{\varphi}
  &*{}
  \\
  *{}
  &
   %{l''}
   *{\;\cdots}
 }}
 \end{math}
 } }
 %---------------------------------------------
 \;
 \subcaptionbox{A probabilistic assignment location
 $l\in L_{\mathsf{PA}}$. 
 %Here $\varphi$ is a logical formula
 %\label{cat}
 }[.18\textwidth]
 {
 \scalebox{.7}{\begin{math}
 \entrymodifiers={++[F]}
 \vcenter{\xymatrix@R=0em@C+4em{
   {l}
          \ar[r]^{x \sim \Dist(\paramDist)}
  &   *{\;\cdots}
 }}
 \end{math}
 } }
 %---------------------------------------------
 \;
 \subcaptionbox{A deterministic assignment location
 $l\in L_{\mathsf{DA}}$. 
 %Here $\varphi$ is a logical formula
 %\label{cat}
 }[.18\textwidth]
 {
 \scalebox{.7}{\begin{math}
 \entrymodifiers={++[F]}
 \vcenter{\xymatrix@R=0em@C+2em{
   {l}
          \ar[r]^{x \coloneqq e}
  &   *{\;\cdots}
 }}
 \end{math}
 } }
 % %---------------------------------------------
 % \;
 % \subcaptionbox{An observation location
 % $l\in L_{\mathsf{O}}$. 
 % %Here $\varphi$ is a logical formula
 % %\label{cat}
 % }[.18\textwidth]
 % {
 % \scalebox{.7}{\begin{math}
 % \entrymodifiers={++[F]}
 % \vcenter{\xymatrix@R=0em@C+3em{
 %   {l}
 %          \ar[r]^{\observeClause{\sharpPred}}
 %  &   *{\;\cdots}
 % }}
 % \end{math}
 % } }
 %---------------------------------------------
 \;
 \subcaptionbox{A weight location
 $l\in L_{\mathsf{W}}$. 
 %Here $\varphi$ is a logical formula
 %\label{cat}
 }[.18\textwidth]
 {
 \scalebox{.7}{\begin{math}
 \entrymodifiers={++[F]}
 \vcenter{\xymatrix@R=0em@C+3em{
   {l}
          \ar[r]^{\weightClause{\fuzzyPred}}
  &   *{\;\cdots}
 }}
 \end{math}
 } }
 \caption{ pCFG locations}\label{fig:typesOfLocations}
\end{minipage} 
\end{figure*}
Figure~\ref{fig:typesOfLocations} illustrates the five types of pCFG locations.

The  translation from  $\PIMP$ programs to pCFGs is straightforward, following~\cite{AgrawalCN18}; its details are thus omitted.
\begin{auxproof}
 Translation of imperative programs to control flow graphs is straightforward; this is also the case in our setting of $\PIMP$ programs and pCFGs. Our  translation follows~\cite{AgrawalCN18}; its details are thus omitted.
\end{auxproof}

\subsection{Semantics of pCFGs}\label{subsec:pCFGSemantics}
We introduce formal semantics of pCFGs in a denotational style; see~\cite{Winskel93} for basics. We need semantics to formulate  soundness of condition propagation (Section~\ref{subsec:conditionPropag});  we  also use the semantics in the description of our sampling algorithm. 
We follow~\cite{StatonYWHK16}---which is inspired ultimately by~\cite{Kozen81}---and introduce the following two semantics: 
\begin{itemize}
 \item The \emph{weighted state transformer} semantics $\sem{\place}^{\wst}$  (Definition~\ref{def:pCFGwstSemantics}), where  
%truth values (for sharp observations) 
  weights from soft conditioning  are recorded by the weights of samples. 
 \item The \emph{(unweighted, normalized) state transformer} semantics  $\sem{\place}^{\st}$ (Definition~\ref{def:pCFGstSemantics}),  obtained by normalizing the weighted semantics  $\sem{\place}^{\wst}$. In particular, the normalization process removes samples of weight $0$, i.e.\ those which violate observations. This is the semantics  we wish to sample from.
\end{itemize}
These semantics are  defined in terms of \emph{memory states}  (assignments of values to variables). The actual definition is by induction on the construction of programs, exploiting the $\omega$-cpo structure of $\subdist X$, the set of subprobability distributions. An example is given later (Example~\ref{ex:pCFGSemExample}).

%  and define the \emph{weighted state transformer} semantics.\footnote{In~\cite{StatonYWHK16}, the conditioning construct under a soft constraint is denoted by $\mathtt{score}$. } In particular, the weights are needed in order to deal with soft constraints.

% The semantics that gives the posterior distribution in the Bayesian sense is defined later in Definition~\ref{def:pCFGstSemantics} by normalizing the weighted state transformer semantics. This is the distribution that we wish to sample from.

% In this paper we take the  backward predicate-transformer style, studied e.g.\ in~\cite{MorganMS96,Kozen81,Kozen85,OlmedoGJKKM18}. Our presentation here closely follows that in~\cite{GordonHNR14}, but is adapted from $\PIMP$ to pCFGs.

\begin{mydefinition}\label{def:subdist}
Let $X$ be a measurable space. The set of probability distributions over $X$ is denoted by $\dist(X)$. The set of  \emph{subprobability distributions} $\mu$ over $X$---for which $\mu(X)$ is required to be $\le 1$ rather than $=1$---is denoted by $\subdist(X)$. 
We equip both $\dist(X)$ and $\subdist(X)$
with suitable $\sigma$-algebras given by~\cite{Giry82}. 

The set $\subdist(X)$ comes with a natural order $\le$ defined as follows: $\mu\le \nu$ if $\mu(U)\le \nu(U)$ (in $\R$) for each measurable subset $U$. The set $\subdist(X)$  is an $\omega$-cpo with respect to $\le$; its bottom element is
 given by the \emph{zero distribution} (assigning $0$ to every measurable subset). 
The $\omega$-cpo structure is inherited by function spaces of the form $(\subdist(X))^{A}$, where $A$ is an arbitrary set, where the order is defined pointwise.
Note that this order becomes trivial in $\dist(X)$ due to the normalization condition.

For a  distribution $\mu$, its support is denoted by $\mathsf{supp}(\mu)$. 
The Dirac distribution at $x\in X$ is denoted by $\delta_{x}$. 
\end{mydefinition}

\begin{mydefinition}[memory state, interpretation of expressions]\label{def:memStates}
 Let $ \Gamma=(L,V,
%\linit,\siginit,\lfinal,\efinal,\trrel,\lambda
\dotsc
)$ be a pCFG. 
 A \emph{memory state} $\sigma$ for $\Gamma$ is a function $\sigma\colon V\to \R$ that maps each variable $x\in V$ to its value $\sigma(x)$. (Recall from Section~\ref{sec:theLang} that, for simplicity, we embed the values of all basic types in $\R$.) The \emph{interpretation} $\sem{e}_{\sigma}$ of an expression $e$ under $\sigma$ is defined inductively; so is the interpretation $\sem{\varphi}_{\sigma}$ of a Boolean formula.

 We define $\St$ to be the measurable space over the set
 of functions of type  $V\to \R$, equipped with the coarsest $\sigma$-algebra making the evaluation function
 $(\place)(x):(V\to \R)\to\R$ measurable for each $x\in V$. This makes $\St$
 isomorphic to the product of $|V|$-many copies of $\R$.
\end{mydefinition}

\begin{mydefinition}[weighted state transformer semantics $
 \sem{\place}^{\wst}$]\label{def:pCFGwstSemantics}

 Let $ \Gamma=(L,V,\linit,\siginit,\lfinal,\efinal,\trrel,\lambda)$ be a pCFG. 
For each location $l\in L$ of the pCFG $\Gamma$, we define its \emph{interpretation}
\begin{displaymath}
 \sem{\Gamma,l}^{\wst}\colon \St\longrightarrow \subdist\bigl(
%\Rnn
\Rnn 
\times\St\bigr),
%\longrightarrow(\St\to[0,1])
%\enspace,
\end{displaymath}
which is a measurable function,
by the \emph{least} solution of the system of recursive equations shown
 in Figure~\ref{fig:wstSemOfpCFG}. Intuitively,  $\sem{\Gamma,l}^{\wst}(\sigma)$ is the subprobability
 distribution of weighted samples of memory states,
 obtained at the final state
 $\lfinal$ after an execution of $\Gamma$ that starts at the location $l$
 with a memory state $\sigma$. It is a \emph{sub}-probability distribution
 since an execution might be nonterminating. 

For the whole 
 pCFG $\Gamma$, its  \emph{weighted state transformer semantics} 
\begin{equation}\label{eq:weightedSemTypeWhole}
  \sem{\Gamma}^{\wst}\in \subdist\bigl(
%\Rnn
\Rnn 
\times\R\bigr)
\end{equation}
is defined as follows, using an intermediate construct
\begin{math}
 \sem{\Gamma}^{\overline{\wst}}\in \subdist\bigl(
%\Rnn
\Rnn 
\times\St\bigr)
\end{math}.
\begin{align}
&\sem{\Gamma}^{\overline{\wst}} \;\coloneqq\;  \sem{\Gamma,\linit}^{\wst}(\siginit), 
\nonumber
% \\
% & \sem{\Gamma}^{\wst}(w,v)
% \;\coloneqq\;
%  \int \mathbf{1}_{\sem{\efinal}_{\sigma}=v} \,
%  % \xi(w, \sigma)
% \sem{\Gamma}^{\overline{\wst}}(w, \sigma)
% \;\mathrm{d}\sigma\enspace.
% \label{eq:defOfSemOfWholeProgram}
\\
& \sem{\Gamma}^{\wst}(\mathrm{d}w\times\mathrm{d}v)
\;\coloneqq\;
 \int_{\sigma\in\St}
% \mathbf{1}_{\sem{\efinal}_{\sigma}\in\mathrm{d}v}
\delta_{\sem{\efinal}_{\sigma}}(\mathrm{d}v)
 \,\cdot\,
 % \xi(w, \sigma)
\sem{\Gamma}^{\overline{\wst}}(\mathrm{d}w\times \mathrm{d}\sigma)
%\;\mathrm{d}v
\enspace.
\label{eq:defOfSemOfWholeProgram}
%  \\
%  &\text{where}\quad
%  \xi
% \coloneqq
%  \sem{\Gamma,\linit}^{\wst}(\siginit)
% \quad\in\;
% \subdist\bigl(
% %\Rnn
% \Rnn 
% \times\St\bigr)\enspace. \nonumber
\end{align}
Recall that $\delta_{\sem{\efinal}_{\sigma}}$ is a Dirac distribution:
$\delta_{\sem{\efinal}_{\sigma}}(\mathrm{d}v)=1$ if and only if $\sem{\efinal}_{\sigma}\in \mathrm{d}v$; its value is $0$ otherwise.
%  $\mathbf{1}_{\sem{\efinal}_{\sigma}=
% v}$ is the characteristic function for the designated equality.
\end{mydefinition}

\begin{figure*}[tbp]
\scalebox{.8}{\begin{minipage}{1.25\textwidth}
 \begin{align*}
 % \sem{\Gamma,l}^{\wst}(\sigma)
 % &=
 % \sem{\Gamma,l'}^{\wst}(\sigma) \mathbin{{}_{p}{+}} \sem{\Gamma,l''}^{\wst}(\sigma)
 % &&
 % \begin{minipage}[t]{.45\textwidth}
 %  if $l^{}\in L_{\mathsf{P}}$, with 
 %       transitions $l^{}\xrightarrow{p}l'$ and
 %       $l^{}\xrightarrow{1-p}l''$.
 % Here the operator ${}_{p}{+}$ denotes a weighted sum of distributions.
 % \end{minipage}
 \\
 \sem{\Gamma,l}^{\wst}(\sigma)
 &=
 \begin{cases}
  \sem{\Gamma,l'}^{\wst}(\sigma)
  &\text{if $\sem{\varphi}_{\sigma}=\true$}
 \\
  \sem{\Gamma,l''}^{\wst}(\sigma)
  &\text{if $\sem{\varphi}_{\sigma}=\false$}
 \end{cases}
 &
 &
 \begin{minipage}[t]{.45\textwidth}
  if $l^{}\in L_{\mathsf{D}}$, with 
       transitions $l^{}\xrightarrow{\varphi}l'$ and
       $l^{}\xrightarrow{\mathop{!}\varphi}l''$.
 \end{minipage} 
 \\
 \sem{\Gamma,l}^{\wst}(\sigma)
 &=
 \int_{v\in\R} \sem{\Gamma,l'}^{\wst}\bigl(\sigma[x\mapsto v]\bigr) \cdot \Dist\bigl(\sem{\paramDist}_{\sigma}\bigr)(v)\; \mathrm{d}v
 &
 &
 \begin{minipage}[c]{.45\textwidth}
 if $l^{}\in L_{\mathsf{PA}}$, with a
       transition $l^{}\xrightarrow{x \sim \Dist(\paramDist)}l'$.
 Here $\Dist(\sem{\paramDist}_{\sigma})$ is the distribution $\Dist$ with its parameters instantiated with $\sem{\paramDist}_{\sigma}$.  A state update, assigning $v$ to $x$, is denoted by $\sigma[x\mapsto v]$.
 \end{minipage} 
 % \\
 %  &\parbox[t]{.8\textwidth}{Here $\Dist(\sem{\paramDist}_{\sigma})$ is the distribution $\Dist$ instantiated with parameters $\sem{\paramDist}_{\sigma}$.  A state update is denoed by $\sigma[x\mapsto v]$.}
 \\
 \sem{\Gamma,l}^{\wst}(\sigma)
 &=
 \sem{\Gamma,l'}^{\wst}\bigl(\,\sigma\bigl[x\mapsto \sem{e}_{\sigma}\bigr]\,\bigr) 
 &&\begin{minipage}[c]{.45\textwidth}
   if $l^{}\in L_{\mathsf{DA}}$, with a
  transition $l^{}\xrightarrow{x \coloneqq e}l'$.\end{minipage}
 % \\
 % \sem{\Gamma,l}^{\wst}(\sigma)
 % &=
 % \Bigl[\; \bigl(\,\sem{\mathbf{1}_{\sharpPred}}_{\sigma}\cdot\mathrm{d}r,\;\mathrm{d}\sigma'\,\bigr)
 % \longmapsto
 % \sem{\Gamma,l'}^{\wst}(\sigma)(\mathrm{d}r,\mathrm{d}\sigma')
 % \;\Bigr]
 % &&\begin{minipage}[t]{.35\textwidth}
 %    if $l^{}\in L_{\mathsf{O}}$, with a
 %       transition $l^{}\xrightarrow{\observeClause{\sharpPred}}l'$.
 %   \end{minipage}
 % \addtocounter{equation}{1}\tag{\theequation}\label{eq:wstSemConditioning}
 \\
 \sem{\Gamma,l}^{\wst}(\sigma)
 &=
\begin{cases}
 \pmb{\lambda} (\mathrm{d}r\times\mathrm{d}\sigma').\;
\sem{f}_{\sigma}\cdot
  \sem{\Gamma,l'}^{\wst}(\sigma)\bigl(
%\frac{1}{\sem{f}_{\sigma}}
(\mathrm{d}r/\sem{f}_{\sigma})\times\mathrm{d}\sigma'\bigr)
\\
  \qquad\qquad\qquad\qquad\text{if $\sem{f}_{\sigma}\neq 0$;}
\\
  \pmb{\lambda} (\mathrm{d}r\times\mathrm{d}\sigma').\;
\\
  \qquad\begin{cases}
	 \int_{r'\in \Rnn}   \sem{\Gamma,l'}^{\wst}(\sigma)(\mathrm{d}r'\times\mathrm{d}\sigma')
	 \\
	 \phantom{0}\qquad\text{if $0\in \mathrm{d}r$;}
         \\
         0
	 \qquad\text{otherwise;}
	\end{cases}
\\
  \qquad\qquad\qquad\qquad\text{if $\sem{f}_{\sigma}= 0$}
	\end{cases}
 % \Bigl[\; \bigl(\,\sem{f}_{\sigma}\cdot\mathrm{d}r,\;\mathrm{d}\sigma'\,\bigr)
 % \longmapsto
 % \sem{\Gamma,l'}^{\wst}(\sigma)(\mathrm{d}r,\mathrm{d}\sigma')
 % \;\Bigr]
 &&\begin{minipage}[c]{.35\textwidth}
    if $l^{}\in L_{\mathsf{W}}$, with a
       transition $l^{}\xrightarrow{\weightClause{f}}l'$. More explanations, such as the definition of $\mathrm{d}r/\sem{f}_{\sigma}$,
 are given after Definition~\ref{def:pCFGwstSemantics}.
   \end{minipage}
 % \\
 % &\text{if $l^{}\in L_{\mathsf{O}}$, with a
 %       transition $l^{}\xrightarrow{\observeClause{f}}l'$.}
 \addtocounter{equation}{1}\tag{\theequation}\label{eq:wstSemWeight}
 \\
 \sem{\Gamma,\lfinal}^{\wst}(\sigma)
 &=
 \delta_{(1,\sigma)}
  &&\begin{minipage}[t]{.45\textwidth}
     where $\delta_{(1,\sigma)}$ is the Dirac distribution at $(1,\sigma)$.
 \end{minipage}
 % \\
 % &\text{if $l^{}\in L_{\mathsf{O}}$, with a
 %       transition $l^{}\xrightarrow{\observeClause{f}}l'$.}
\end{align*}\end{minipage}
}
\caption{Recursive definition of $ \sem{\Gamma,l}^{\wst}$ used in the weighted state transformer semantics  of pCFGs}\label{fig:wstSemOfpCFG}
\end{figure*}

Intuitively, the definition~(\ref{eq:defOfSemOfWholeProgram}) is the continuous variation of the following one (that only makes sense if all the value domains are discrete). 
\begin{displaymath}
  \sem{\Gamma}^{\wst}(w, v)
\;\coloneqq\;
 \sum_{\sigma \text{ s.t.\ } \sem{\efinal}_{\sigma}=v }
 \sem{\Gamma,\linit}^{\wst}(\siginit)(w,\sigma)\enspace.
\end{displaymath}

Similarly, the case~(\ref{eq:wstSemWeight}) for $\mathtt{weight}$ locations in Figure~\ref{fig:wstSemOfpCFG} can be described more simply if the distributions involved are discrete. Let the distribution $\sem{\Gamma,l'}^{\wst}(\sigma)$ be $\bigl[\,(r_{i},\sigma'_{i})\mapsto p_{i}\,\bigr]_{i}$, where each pair $(r_{i},\sigma'_{i})$ of a weight and a state gets a probability  $p_{i}$ assigned. Then the distribution $\sem{\Gamma,l}^{\wst}(\sigma)$ should be given by $\bigl[\,\bigl(\,\sem{f}_{\sigma}\cdot r_{i},\,\sigma'_{i}\,\bigr)\mapsto p_{i}\,\bigr]_{i}$, where we multiply all the  weights $r_{i}$ by the nonnegative real number $\sem{f}_{\sigma}$. In particular, if $\sem{f}_{\sigma}=0$, then we should have
\begin{equation}\label{eq:weightSemDisdc}
\begin{aligned}
  \bigl(\,\sem{\Gamma,l}^{\wst}(\sigma)\,\bigr)(0,\sigma') &=
 \textstyle\sum_{r'\in \Rnn} \bigl(\,\sem{\Gamma,l'}^{\wst}(\sigma)\,\bigr)(r',\sigma')\enspace,
\\
  \bigl(\,\sem{\Gamma,l}^{\wst}(\sigma)\,\bigr)(r,\sigma') &= 0
 \qquad\text{if $r>0$.}
\end{aligned}
\end{equation}

Based on the last intuition, the full definition (\ref{eq:wstSemWeight}) in Figure~\ref{fig:wstSemOfpCFG} is explained as follows. In the first line, $\mathrm{d}r/\sem{f}_{\sigma}$ denotes the measurable set $\mathrm{d}r$ scaled by $1/\sem{f}_{\sigma}$; for example, if $\mathrm{d}r=[s,t]$, then $\mathrm{d}r/\sem{f}_{\sigma}=\bigl[s/\sem{f}_{\sigma}, t/\sem{f}_{\sigma}\bigr]$. This achieves the effect of ``multiplying weights $r_{i}$ by $\sem{f}_{\sigma}$,'' with the side-effect, however, of changing the measure of the set $\mathrm{d}r$. The first occurrence of $\sem{f}_{\sigma}$ in the first line of (\ref{eq:wstSemWeight}) cancels this side-effect. 
 The second case of (\ref{eq:wstSemWeight}) in Figure~\ref{fig:wstSemOfpCFG} (Lines~3--7) is a continuous adaptation of~(\ref{eq:weightSemDisdc}). 
% The case~(\ref{eq:wstSemConditioning}) is the special case of~(\ref{eq:wstSemWeight}) where $f=\mathbf{1}_{\varphi}$ is the characteristic function $\varphi$; recall that we are keeping $\observeClause\sharpPred$ as a primitive only for the convenience in condition propagation (Section~\ref{subsec:conditionPropag}).

The  least solution of the recursive equation for $\bigl(\sem{\Gamma,l}^{\wst}\bigr)_{l\in L}$ in Figure~\ref{fig:wstSemOfpCFG} exists. It can be  constructed as the supremum of a suitable increasing $\omega$-chain, exploiting the $\omega$-cpo structure of the set of
measurable functions of type $\St\times L\to \subdist\bigl(\Rnn \times\St\bigr)$. See e.g.~\cite{GordonHNR14}.

This construction via a supremum also matches the operational intuition of collecting the return values of all the execution paths of length $0, 1, 2, \dotsc$.

The weighted semantics $\sem{\Gamma}^{\wst}\in \subdist\bigl(
%\Rnn
\Rnn 
\times\R\bigr)$ in~(\ref{eq:weightedSemTypeWhole}) is randomized,
since an execution of $\Gamma$ has  uncertainties that come from the randomizing construct
%probabilistic assignments
 $x \sim \Dist(\paramDist)$.
%and
% probabilistic branching
% $\mathtt{ifp}$. 
It is a \emph{subprobability} distribution---meaning that the probabilities need not sum up to $1$, see Definition~\ref{def:subdist}---since an  execution of $\Gamma$ may not terminate. See e.g.~\citep{MorganMS96}. Note that violation of
observations $\observeClause{\sharpPred}$ is recorded as  the zero weight
(in the $\Rnn$ part in~(\ref{eq:weightedSemTypeWhole})), rather than making the  sample disappear.

By applying normalization to  the weighted semantics in Definition~\ref{def:pCFGwstSemantics}, we obtain the following (unweighted) state transformer semantics. This semantics is the posterior distribution in the sense of Bayesian inference. It is therefore the distribution that we would like to sample from.

\begin{mydefinition}[(unweighted) state transformer semantics $ \sem{\place}^{\st}$]
 \label{def:pCFGstSemantics}
 Let 
%$ \Gamma=(L,V,\linit,\siginit,\lfinal,\efinal,\trrel,\lambda)$ 
$\Gamma$ be a pCFG. Its \emph{state transformer semantics} is the probability distribution
\begin{equation}\label{eq:unweightedSemTypeWhole}
 \sem{\Gamma}^{\st}\in \dist(\R)
\end{equation}
defined as follows, using an intermediate construct 
$\sem{\Gamma}^{\overline{\st}}\in \dist(\St)$.
\begin{align}
&
% \sem{\Gamma}^{\overline{\st}}\in \dist(\St);
% \quad
 \sem{\Gamma}^{\overline{\st}}(
         \mathrm{d}
         \sigma)\coloneqq
\frac{
\int_{r\in\R}\,r\cdot  \sem{\Gamma}^{\overline{\wst}}(\mathrm{d}r\times
\mathrm{d}
\sigma) 
%\;\mathrm{d}r
\,
}{
\int_{r\in\R, \sigma'\in \St}\,r\cdot  \sem{\Gamma}^{\overline{\wst}}(\mathrm{d}r\times\mathrm{d}\sigma') 
%\;\mathrm{d}r\,\mathrm{d}\sigma'
}
 \enspace;
 \label{eq:unweightedSemanticsIntermediate}
\\
% &\sem{\Gamma}^{\st}\in \dist(\R);
% \quad
%\quad\text{and}\quad
& \sem{\Gamma}^{\st}(
\mathrm{d}
v)\coloneqq
 \textstyle\int_{\sigma\in\St}\,
% \mathbf{1}_{\sem{\efinal}_{\sigma}=
% %\mathrm{d}
% v}
  \delta_{\sem{\efinal}_{\sigma}}(\mathrm{d}v)
 \cdot
 \sem{\Gamma}^{\overline{\st}}(\mathrm{d}\sigma)
% \;\mathrm{d}\sigma
\enspace.
 \nonumber
\end{align}
Here $\delta_{\sem{\efinal}_{\sigma}}$ is the Dirac distribution at $\sem{\efinal}_{\sigma}$. 
%  $\mathbf{1}_{\sem{\efinal}_{\sigma}=
% v}$ is the characteristic function for the designated equality.
% See Section~\ref{subsec:notation}.
The semantics is undefined in case the denominator $\int_{r\in\R,\sigma'\in\St}\,r\cdot  \sem{\Gamma}^{\overline{\wst}}(\mathrm{d}r\times\mathrm{d}\sigma') 
%\;\mathrm{d}r\,\mathrm{d}\sigma'
$ is $0$. 
\end{mydefinition}

\begin{myexample}\label{ex:pCFGSemExample}
 The program \texttt{coin} in Program~\ref{listing:ProbProgramCoinEmulation}  emulates a fair coin using a biased one, crucially relying on
the $\mathtt{observe}$
command (Line~\ref{line:FairCoinEmuOb}).
Let $\Gamma$ be the pCFG induced by it. The return value domain is $\DDom=\mathbf{Bool}$. 

 The weighted semantics
%  $\sem{\Gamma}^{\wst}
% \in \subdist\bigl(
% %\Rnn
% \Rnn 
% \times\mathbf{Bool}\bigr)
% $ 
is 
\begin{math}
\sem{\Gamma}^{\wst}=
\bigl[
  (0,\true) \mapsto 0.36^{2}, %\times 0.36,
  (1,\true) \mapsto 0.36\times 0.64,
  (1,\false) \mapsto 0.64\times 0.36,
  (0,\false) \mapsto 0.64^{2}%\times 0.64
\bigr]
\in \subdist\bigl(
\Rnn 
\times\mathbf{Bool}\bigr)
\end{math}. Here the first sample $(0,\true)$---it comes from the memory state $[c1\mapsto\true, c2\mapsto \true]$---has a  weight $0$ due to its violation of the observation $\mathtt{!(c1 = c2)}$. 
%
% \vskip-\parskip  % new paragraph, but no additional space
% {\scriptsize\begin{displaymath}
% \bigl[ 
% \begin{array}{l}
%  (0,\true) \mapsto 0.36\times 0.36,\quad
%   (1,\true) \mapsto 0.36\times 0.64,\quad
%   (1,\false) \mapsto 0.64\times 0.36,\quad
%   (0,\false) \mapsto 0.64\times 0.64
% \end{array}
% \bigr].
% \end{displaymath}}
% Note, for example, that the memory state $[c1=c2=\true]$ occurs with the probability $0.36\times 0.36$, returns $\true$ as the value of $c1$, and its violation of the observation (Line~9) is recorded as the zero weight.
%
After normalization (which wipes out the contribution of the weighted sample $(0,\true)$), we obtain the unweighted semantics $\sem{\Gamma}^{\st}=[\true\mapsto 1/2, \false\mapsto 1/2]$, as expected.

\begin{auxproof}
 It is not hard to see that the weighted semantics $\sem{\Gamma,\linit}^{\wst}(\sigma)\in 
 \subdist(
 \Rnn 
 \times\St)
 $, for any $\sigma$, is given by
 \begin{displaymath}
 \left[ \small
 \begin{array}{l}
 (0,(c1,c2)) \mapsto 0.36\times 0.36,\quad
  (1,(c1,!c2)) \mapsto 0.36\times 0.64,
 \\
  (1,(!c1,c2)) \mapsto 0.64\times 0.36,\quad
  (0,(!c1,!c2)) \mapsto 0.64\times 0.64
 \end{array}
 \right],
 \end{displaymath}
 where $(c1,!c2)$ stands for the valuation $[c1\mapsto\true,c2\mapsto\false]$, and so on. 
 Note that the violation of the observation $c1\neq c2$ is reflected in the weights of the first and last entries. 
  The weighted semantics $\sem{\Gamma}^{\wst}$ of the whole pCFG $\Gamma$ is then derived as $\bigl[\,(0,\true) \mapsto 0.36\times 0.36,\dotsc\,\bigr]$, focusing on the value of the return expression $c1$. After normalization, we obtain the (unweighted) semantics $[\true\mapsto 1/2, \false\mapsto 1/2]$, as expected. 
\end{auxproof}
\end{myexample}

\subsection{Control Flows and Straight-Line Programs}\label{subsec:controlFlow}
In our hierarchical architecture (Figure~\ref{fig:hierarchicalSamplerArch}), the top-level chooses a \emph{complete control flow} $\flowl$ of a pCFG; $\flowl$ induces a  \emph{straight-line program} $\StrLn(\flowl)$; and the bottom-level data sampler works on $\StrLn(\flowl)$.  These notions are formally defined below; roughly, a \emph{control flow} is a path in a pCFG that starts at the initial state $\linit$; and a \emph{complete} control flow is one that ends at the final location $\lfinal$. 
In the definition of the  straight-line program $\StrLn(\flowl)$,
% induced by $\flowl$, 
the main point is to turn guards (for if branchings and while loops)
%(in (a,b,e) of Figure~\ref{fig:typesOfLocations}) 
into suitable observations.

\begin{mydefinition}[control flow]\label{def:controlFlow}
 Let $ \Gamma=(L,V,\linit,\siginit,\lfinal,\efinal,\trrel,\lambda)$ be a pCFG. A \emph{control flow} of $\Gamma$ is a finite sequence $\flowl=l_{1} l_{2}\dotsc l_{n}$ of locations, where we require $l_{1}=\linit$ and $(l_{i},l_{i+1})\in{\to}$ for each $i\in [1,n-1]$. A control flow is often denoted together with  transition labels, that is, 
\begin{displaymath}
 \linit 
  \;=\;
 l_{1}
  \xrightarrow{\lambda(l_{1},l_{2})}
 l_{2}
  \xrightarrow{\lambda(l_{2},l_{3})}
  \cdots
  \xrightarrow{\lambda(l_{n-1},l_{n})}
 l_{n}\enspace.
\end{displaymath}
A control flow $\flowl=l_{1} l_{2}\dotsc l_{n}$ is said to be \emph{complete} if $l_{n}=\lfinal$. 

The set of  control flows of a pCFG $\Gamma$ is denoted by $\CF(\Gamma)$; 
the set of complete ones is denoted by $\CCF(\Gamma)$. 
\end{mydefinition}

\begin{mydefinition}[straight-line program]\label{def:straightLineProgram}
A \emph{straight-line program} is a pCFG that
has no deterministic locations.

% such that
% \begin{itemize}
%  \item 
%   it has no deterministic locations, and 
%  \item 
%    each of its probabilistic locations has only one outgoing transition, labeled by $1$.
% \end{itemize}
Therefore, a straight-line program is identified with a triple
 \begin{displaymath}
  \bigl(\,\linit\xrightarrow{\lambda_{1}} l_{2}\xrightarrow{\lambda_{2}} \cdots\xrightarrow{\lambda_{n-1}} \lfinal,\;\,\siginit,\;\,\efinal\,\bigr)\enspace,
% \\
% \text{also denoted by}\quad\xrightarrow{\siginit} \linit\xrightarrow{\lambda_{1}} l_{2}\xrightarrow{\lambda_{2}} \cdots\xrightarrow{\lambda_{n-1}} \lfinal\xrightarrow{\efinal}\enspace.
 \end{displaymath}
that shall be also denoted by 
\begin{displaymath}
 \xrightarrow{\siginit} \linit\xrightarrow{\lambda_{1}} l_{2}\xrightarrow{\lambda_{2}} \cdots\xrightarrow{\lambda_{n-1}} \lfinal\xrightarrow{\efinal}\enspace.
\end{displaymath}
The first component of the above triple is 
 a chain that   consists of the following types of locations: 
%1) probabilistic locations $l\xrightarrow{1}\cdot\,$; 
1) assignment locations $l\xrightarrow{x \sim \Dist(\paramDist)}\cdot\,$ and $l\xrightarrow{x \coloneqq e}\cdot\,$; 
% 2) observation locations $l\xrightarrow{\observeClause{\sharpPred}}\cdot$; 
2) weight locations $l\xrightarrow{\weightClause{\fuzzyPred}}\cdot$ (where we might use $l\xrightarrow{\observeClause{\sharpPred}}\cdot$ as a shorthand); 
and 3) a final location. The remaining components are an initial memory state $\siginit$ and a return expression  $\efinal$. 
%Another possible notation for the above straight-line program is 
\end{mydefinition}

% For a complete control flow $\flowl$ of a pCFG, we define 
% %two straight-line programs $\StrLnd(\flowl)$, $\StrLnf(\flowl)$ 
% a straight-line program $\StrLn(\flowl)$
% induced by $\flowl$. 
% They are used on different levels in our sampling algorithm: $\StrLnd(\flowl)$ is for data sampling; and  $\StrLnf(\flowl)$ is for control flow sampling.

\begin{mydefinition}[the straight-line program $\StrLn(\flowl)$%
%the straight-line programs $\StrLnd(\flowl)$, $\StrLnf(\flowl)$
]
\label{def:fromControlFlowToStrLn}
 Let $\Gamma$ be a pCFG, and $\flowl=l_{1}\dotsc l_{n}$ be a complete control flow of $\Gamma$. 
The straight-line  program $\StrLn(\flowl)$ is obtained by applying the following 
%operations
operation
 to $\flowl=l_{1}\dotsc l_{n}$. 

\begin{quote}
 Each deterministic location $l_{i}\in L_{\mathsf{D}}$ occurring in $\flowl$ is changed into an observation location. Accordingly, the label of the transition $l_{i}\xrightarrow{\varphi}l_{i+1}$ is made into an observation command, resulting in 
$l_{i}\xrightarrow{\observeClause{\varphi}}l_{i+1}$. 
\end{quote}
\end{mydefinition}

\begin{myexample}[$\StrLn(\flowl)$]\label{ex:strLn1}
 For the pCFG in Figure~\ref{fig:pCFGexample1}, the set of complete control flows is
 \begin{math}
  \{\flowl_{n}\mid n\ge 0\}  
 \end{math} where
\begin{math}
 \flowl_{n}\,:=\,\linit\, \bigl(l^{(2)}l^{(3)}l^{(4)}l^{(5)}\linit\bigr)^{n}\,l^{(6)}\,\lfinal
\end{math}.
 The flow $\flowl_{1}$
% \begin{math}
% \flowl=
% \bigl(\,
%  \linit\, l^{(2)}\, l^{(3)}\,l^{(4)}\,l^{(5)}\,\linit\,l^{(6)}\,\lfinal
%  \,\bigr)
% \end{math},
% \begin{displaymath}
% \flowl\;:=\;
% %\bigl(\,
%  \linit\, l^{(2)}\, l^{(3)}\,l^{(4)}\,l^{(5)}\,\linit\,l^{(6)}\,\lfinal
% % \,\bigr)
% \end{displaymath}
%The two straight-line programs $\StrLnd(\flowl)$, $\StrLnf(\flowl)$ are defined as follows. Here $\siginit=\bigl[\,x\mapsto 0,\, z\mapsto 0,\, n\mapsto 0\bigr]$. 
induces the following  straight-line program.
\begin{align*}\small
\begin{array}{l}
  \StrLn(\flowl_{1})
 =
 \left(\,
 \begin{array}{l}
  \xrightarrow{\siginit}
 \linit
 \xrightarrow{\observeClause{x<3}}
  l^{(2)}
 \xrightarrow{  n := n + 1}
  l^{(3)}
 \\
 \xrightarrow{y\sim
	      \mathtt{normal}(1,1)}
  % \phantom{\xrightarrow{\siginit}}
 l^{(4)}
 \xrightarrow{\observeClause{0\le y \le 2} }
 l^{(5)}
   \xrightarrow{{\color{black} x := x + y}}
 %\xrightarrow{{\color{red}\observeClause{\true }}}
 \linit
 \\
 \xrightarrow{\observeClause{\mathop{!}(x<3)}}
 % \\
 %  \phantom{\xrightarrow{\siginit}}
 l^{(6)}
  \xrightarrow{{\color{black}\observeClause{n \ge 5}}}
 %\xrightarrow{{\color{red}\observeClause{\true }}}
 \,\lfinal
 \xrightarrow{n}
 \end{array} 
 \,\right)
\end{array}
% \\
%  \StrLnf(\flowl)
%  &\coloneqq
%  \left(\,
% \begin{array}{r}
%   \xrightarrow{\siginit}
%  \linit
%  \xrightarrow[{\color{red}x<3}]{{\color{red}\observeClause{\true }}}
%   l^{(2)}
%  \xrightarrow{  n := n + 1}
%   l^{(3)}
%  \xrightarrow{z\sim
% 	      \mathtt{normal}(1,1)}
%  l^{(4)}
%  \xrightarrow{  x := x + z}
%  l^{(5)}
%  \\
%  % \xrightarrow{\observeClause{0\le z \le 2}}
%  \xrightarrow[{\color{red}0\le z \le 2}]{{\color{red}\observeClause{\true }}}
%  \linit
%  \xrightarrow[{\color{red}\mathop{!}(x<3)}]{{\color{red}\observeClause{\true }}}
%  l^{(6)}
%  % \xrightarrow{\observeClause{n \ge 5}}
%  \xrightarrow[{\color{red}n \ge 5}]{{\color{red}\observeClause{\true }}}
%  \,\lfinal
%  \xrightarrow{n}
% \end{array} 
% \,\right)
\end{align*}
% It is denoted by $\StrLn(\flowl)$. 
Here $\siginit=[\,x\mapsto 0,\, y\mapsto 0,\, n\mapsto 0]$. Note that the  label $x<3$ going out of $\linit$ in Figure~\ref{fig:pCFGexample1} has been turned into the  observation command $\observeClause{x<3}$ in $\StrLn(\flowl_{1})$.
% the latter imposes satisfaction of the guard $x<3$ so that the  control flow $\flowl_{1}$ is realized.
% The red color signifies the difference between the two straight-line programs. We note that the formulas below $\rightarrow$ are annotations to the programs: they do not affect the semantics of the programs;  they are used later to estimate certain likelihood.
\end{myexample}

\begin{auxproof}
 \begin{myexample}\label{ex:strLn2}
 Consider the pCFG in Figure~\ref{fig:pCFGexample2}. The sequence 
 \begin{math}
 \flowl\;:=\;
 %\bigl(\,
 \linit\, l^{(2)}\, l^{(4)}\,l^{(6)}\,l^{(7)}\,\lfinal
 % \,\bigr)
 \end{math}
 is a complete control flow. 
 %The two straight-line programs $\StrLnd(\flowl)$, $\StrLnf(\flowl)$ are defined as follows. Here $\siginit=\bigl[\,c1\mapsto \true,\, c2\mapsto \true]$. 
 The straight-line program $\StrLn(\flowl)$ is as follows. 
 %Here $\siginit=\bigl[\,c1\mapsto \true,\, c2\mapsto \true]$. 
 \begin{align*}
 % \StrLn(\flowl) 
 % &\coloneqq
 % \left(\,
 \begin{array}{r}
  \xrightarrow{\siginit}
 \linit
 \xrightarrow{1}
  l^{(2)}
 \xrightarrow{ c1 := \true}
  l^{(4)}
 \xrightarrow{1}
 l^{(6)}
 \xrightarrow{ c2 := \false}
 \\
  \phantom{\xrightarrow{\siginit}}
 l^{(7)}
  \xrightarrow{\observeClause{\mathop{!}(c1 = c2)}}
 %\xrightarrow{{\color{red}\observeClause{\true }}}
 \lfinal
 \xrightarrow{c1}
 \end{array} 
 %\,\right)
 % \\
 %  \StrLnf(\flowl) 
 %  &\coloneqq
 %  \left(\,
 % \begin{array}{r}
 %   \xrightarrow{\siginit}
 %  \linit
 %  \xrightarrow{1}
 %   l^{(2)}
 %  \xrightarrow{ c1 := \true}
 %   l^{(4)}
 %  \xrightarrow{1}
 %  l^{(6)}
 %  \xrightarrow{ c2 := \false}
 %  l^{(7)}
 %  % \xrightarrow{\observeClause{\mathop{!}(c1 = c2)}}
 %  \xrightarrow[\color{red}\mathop{!}(c1 = c2)]{{\color{red}\observeClause{\true }}}
 %  \lfinal
 %  \xrightarrow{c1}
 % \end{array} 
 % \,\right)
 \end{align*}
 % The red color signifies the difference between the two straight-line programs. 
 \end{myexample}
\end{auxproof}

We use a breadth-first search algorithm for the static discovery of control flows. 
A high-level description of the algorithm is as follows. A \emph{search tree} $\mathcal{T}$ is obtained by (partially) unrolling the pCFG $\Gamma$. Additionally, each node $n$ of $\mathcal{T}$ records its distance from its shallowest descendant that is an open leaf (called the \emph{height} of $n$). The search for a new control flow goes down from the root, choosing a child with the smallest height. When there are multiple such children, we pick one of them randomly. 

While the algorithm is quite straightforward, we formally describe it
in Algorithm~\ref{algo:ctrlFlowDiscovery} for the record.
 The algorithm uses the following notions. A \emph{search tree} $\mathcal{T}$ for a pCFG \begin{math}
 \Gamma=(L,V,\linit,\siginit,\lfinal,\efinal,\trrel,\lambda)
\end{math} (as in Definition~\ref{def:pCFG}) is a finite tree with a branching degree $\le 2$, identified as a subset $\mathcal{T}\subseteq \{0,1\}^{*}$ as usual. Each node $n$ of a search tree $\mathcal{T}$ is labeled with a location $\loc(n)\in L$ of $\Gamma$, and each edge of $\mathcal{T}$ is labeled in the same manner as with the labeling function $\lambda$ of $\Gamma$ (precisely, the edges from a node $n$ have the same labels as the edges from $\loc(n)$ in $\Gamma$). A node $n'$ of $\mathcal{T}$ is an \emph{open leaf} if it is a leaf in $\mathcal{T}$ and $\loc(n')$ is not final ($\loc(n')\neq\lfinal$), in which case the control flow represented by $n'$ is still incomplete. The \emph{height} $\height(n)$ of a node $n$ is the distance from its shallowest open leaf. Initially, a search tree $\mathcal{T}$ is set to be the one-node tree $\{\varepsilon\}$---where $\varepsilon$ represents the empty sequence, identified as the root---with $\loc(\varepsilon)=\linit$ and $\height(\varepsilon)=0$. 

We do not use the usual queue-based implementation of breadth-first search. The reason is that the search tree $\mathcal{T}$ we explicitly construct in Algorithm~\ref{algo:ctrlFlowDiscovery} is a convenient data structure for \emph{sampling} control flows (Algorithms~\ref{algo:infinitelyArmedSampling} \& \ref{algo:ourHierarchicalSampler}).

 %%%%%%%%%%%%%%%%%%%%%%%%%%%%%%%%%%%%%%%%%%%%%%%%%%%%%%%%%%%%%%%%%%%
\begin{figure*}[tbp]
\begin{adjustbox}{scale=.9}
 \begin{minipage}[t]{1.1\textwidth}
% \null
 \footnotesize
  \begin{algorithm}[H]
  \caption{Our static algorithm for control flow discovery by breadth-first search. It grows a search tree $\mathcal{T}$, potentially yielding a control flow 
  }
  \label{algo:ctrlFlowDiscovery}
  \begin{algorithmic}[1]
  \Require
  pCFG \begin{math}
 \Gamma=(L,V,\linit,\siginit,\lfinal,\efinal,\trrel,\lambda)
\end{math}, search tree $\mathcal{T}$ with at least one open leaf (i.e.\ $\height(\varepsilon)<\infty$)
  % an infinite set $\mathcal{K}$ of arms. Each $k\in \mathcal{K}$ has its (true, not necessarily normalized) \emph{likelihood} $p(k)\in \Rnn$. The \emph{round count} $t$ is initialized to $0$
  \Ensure search tree, control flow of $\Gamma$

  \State $\cnode\gets\epsilon$
         \Comment{set the current node to the root}

  \While{$\mathsf{newCtrlFlow}$ is not defined}
  \While{$\cnode$ is not a leaf of $\mathcal{T}$}
  \Comment{search for a shallowest open leaf}
  \If{$\loc(\cnode)\in L_{\mathsf{PA}}\cup L_{\mathsf{DA}}\cup L_{\mathsf{W}}$}
  \Comment{$\cnode$ has only one child} 
    \State $\cnode\gets \cnode\cdot  0$
  \Else 
  \Comment{$\loc(\cnode)\in L_{\mathsf{D}}$ and thus $\cnode$ has two children}
    \State $\cnode\gets \cnode\cdot \bigl(\,\argmin_{i\in\{0,1\}} \height(\cnode\cdot i)\,\bigr)$
  \EndIf
  \EndWhile
  \Statex 
  \State 
   \begin{minipage}[t]{.8\textwidth}
    add the children of $\cnode$ ($\cnode\cdot 0$ and $\cnode\cdot 1$ if $\loc(\cnode)\in L_{\mathsf{D}}$, or only $\cnode\cdot 0$ otherwise) to $\mathcal{T}$
   \end{minipage}  
\Statex \Comment now $\cnode$ is an open leaf by construction
  \State 
      \begin{minipage}[t]{.8\textwidth}
   label the new children to define $\loc(\cnode\cdot 0)$ and  $\loc(\cnode\cdot 1)$, using the successors of $\loc(\cnode)$ in $\Gamma$
       \end{minipage}
  \State label the edges from $\cnode$ to its children accordingly
  \ForEach{child $\cnode\cdot i$ of $\cnode$}
   \If{$\loc(\cnode\cdot i)=\lfinal$}
     \State $\height(\cnode\cdot i)\gets \infty$ 
     \State $\mathsf{newCtrlFlow}\gets \loc(\cnode_{\le 0}) \loc(\cnode_{\le 1}) 
      % \loc(\cnode_{\le 2})
   \dotsc
       \loc(\cnode)$
     \Statex \qquad\qquad\qquad where $\cnode_{\le i}=m_{1}m_{2}\dotsc m_{i}$ for $\cnode = m_{1}m_{2}\dotsc m_{|\cnode|}$
   \Else
     \State $\height(\cnode\cdot i)\gets 0$
     \Comment open leaf
   \EndIf
  \EndFor
  \While{$\cnode\neq \varepsilon$}
    \Comment we backpropagate and update $\height$
    \State $\height(\cnode)\gets 1 +
    (\text{min height of $\cnode$'s children})
 % \min \{\height(\cnode\cdot i)\mid \text{$\cnode\cdot i$ is a child of $\cnode$}\}
$
    \State $\cnode\gets\mathop{\text{dropLast}}(\cnode)$
    \Comment go to the parent
  \EndWhile
  \EndWhile

  \State \Return $\mathcal{T}, \mathsf{newCtrlFlow}$
  
  \end{algorithmic}
 \end{algorithm}
\end{minipage}
\end{adjustbox}
\end{figure*}

\begin{myremark}\label{rem:ZhouEtAlFlowDiscovery}
Here we discuss comparison with \cite{ZhouYTR20} in terms of flow discovery strategies. 
In our framework (Figure~\ref{fig:hierarchicalSamplerArch}), 
 we discover new control flows \emph{statically} by breadth-first search in a pCFG, also using condition propagation (Section~\ref{subsec:conditionPropag}) and blacklisting (Section~\ref{subsec:mainAlgorithm}) as assistance. 
% The breadth-first search works more specifically as follows: each known node records the depth of its shallowest unknown descendants, and the search goes down from the root choosing a child with the smallest such depth. When there are multiple such children, we pick one of them randomly. 
In contrast, the work~\cite{ZhouYTR20} 
%takes an alternative approach, identifying
identifies \emph{sub-programs} (that correspond to flows) in a \emph{dynamic} manner~\citep[Section~6.2]{ZhouYTR20}. 
 They also use a top-level MCMC sampling for quickly moving from low-likelihood sub-programs to  ones with dominant likelihoods.

The last feature of~\cite{ZhouYTR20} is suited to programs where many sub-programs have non-zero likelihoods. In contrast, our static control flow discovery---combined with condition propagation---is advantageous when many flows have zero likelihoods. It can exhaustively explore the space of (mostly likelihood-zero) control flows, blacklisting those which are found logically infeasible. See Section~\ref{subsec:mainAlgorithm} later.
\end{myremark}

\subsection{Condition Propagation and Domain Restriction}\label{subsec:conditionPropag}
Condition propagation pushes observations upwards in a program, so that those samples which eventually violate observations get
filtered away  earlier. Its contribution to sampling efficiency is demonstrated in R2~\citep{NoriHRS14}; it is also used to completely eliminate observations in programs without probabilistic assignments, in the so-called hoisting transformation in~\cite{OlmedoGJKKM18}.   
Following R2, our  propagation rules are essentially the \emph{weakest precondition
calculus} (see~\cite{Winskel93}). Condition propagation is  more universally
applicable here than in R2, since in our framework (Figure~\ref{fig:hierarchicalSamplerArch}), 
%data sampling only concerns 
condition propagation is applied only to
 \emph{straight-line} programs.  Manual loop invariant annotations are not needed, for example, unlike in R2. 

In this paper,  we also introduce a technique called \emph{domain restriction}. It restricts distributions to certain domains, so that we do not generate  samples that are logically deemed unnecessary. Sampling from domain-restricted distributions is easy to implement, via the inverse transform sampling. 

The  operation that combines the above two is denoted by $\CP$: it transforms a
straight-line program into another semantically equivalent straight-line program, applying domain restriction if possible. 

\myparagraph{Condition Propagation}
For pedagogical reasons, we first introduce the operation $\CPpr$ that conducts condition propagation (but not domain restriction). 

The following definition closely follows the one  in~\cite{NoriHRS14}. It is  much simpler though, since we deal only with straight-line programs.

\begin{mydefinition}[condition propagation $\CPpr$]\label{def:conditionPropagationPlain}
 We define the \emph{condition propagation} operation on straight-line programs, denoted by $\CPpr$, as follows. 

The definition is via an extended operation $\overline{\CPpr}$ 
that takes  a straight-line program\footnote{See Example~\ref{ex:strLn1} for an example of a straight-line program. While we use the notation $\flowl$ to denote a straight-line program, it specifies not only a location sequence but also transition labels between them.} $\flowl$ and returns a pair $(\vec{l'},\fuzzyPred)$ of a straight-line program $\vec{l'}$ and a ``continuation'' fuzzy predicate $\fuzzyPred$. The operation $\overline{\CPpr}$ is defined in the following  backward inductive manner.

For  the base case, we define $\overline{\CPpr}(\lfinal)\coloneqq (\lfinal, 1)$, where $1$ is the constant fuzzy predicate that returns the real number $1$. 

For  the step cases,  we shall define
	  $\overline{\CPpr}(l\xrightarrow{\lambda_{1}} \vec{l'})$---where $\lambda_{1}$ is the label for the first  transition in the straight-line program (it goes out of $l$)---assuming that $\overline{\CPpr}(\vec{l'})$ is already defined by induction. 
	\begin{itemize}
  % 	 \item If $l$ is a probabilistic location, the label $\lambda_{1}$ must be $1$ by the definition of straight-line programs (Definition~\ref{def:straightLineProgram}). We define
  % $\overline{\CPpr}(l\xrightarrow{1} \vec{l'})\coloneqq \overline{\CPpr}(\vec{l'})$. 
	 \item 
Let $l$ be a probabilistic assignment location, and let $(\vec{l^{\bullet}}, \fuzzyPred^{\bullet})=\overline{\CPpr}(\vec{l'})$. We define
	       \begin{equation}\label{eq:condPropagationNextToProbAssignment}
		\begin{aligned}
		 &\overline{\CPpr}\bigl(\,l\xrightarrow{x \sim \Dist(\paramDist)} \vec{l'}\,\bigr)
		 \;\coloneqq
		 % \bigl(\,l\xrightarrow{x \sim \bigl(\Dist(\paramDist)|\fuzzyPred^{\bullet}\bigr)} \vec{l^{\bullet}} \,,\,\psi\,\bigr)\enspace,
		 \\&\quad
		 \begin{cases}
		  \bigl(\,l\xrightarrow{x \sim \Dist(\paramDist)}l^{*}\xrightarrow{\weightClause{\fuzzyPred^{\bullet}}} \vec{l^{\bullet}} \,,\;\mathbf{1}_{\psi}\,\bigr) 
		  % \\\qquad\qquad
                  \quad
		  \text{if $x$  occurs in the fuzzy predicate $\fuzzyPred^{\bullet}$,}
		  \\
		  \bigl(\,l\xrightarrow{x \sim \Dist(\paramDist)}
		  %l^{*}\xrightarrow{\observeClause{\fuzzyPred^{\bullet}}} 
		  \vec{l^{\bullet}} \,,\;\fuzzyPred^{\bullet}\,\bigr) 
		  \quad \text{otherwise.}
		 \end{cases}
		\end{aligned}	       
	       \end{equation}
	       In the first case of the above, $l^{*}$  is a fresh location, and  $\psi$ is a choice of a (sharp) Boolean formula  that makes the following valid.
	       \begin{equation}\label{eq:choiceOfFmlPsi}
		\bigl(\,\exists x\in \mathsf{supp}\bigl(\Dist(\paramDist)\bigr).\; \fuzzyPred^{\bullet}>0\,\bigr) \;\Longrightarrow\; \psi
	       \end{equation}

	       The intuition behind the above definition---it  follows~\cite{NoriHRS14}---is described later in Remark~\ref{rem:intuitionOfDefOfCondPropTheProbAssgnmentCase}. 
The choice of $\psi$ in~(\ref{eq:choiceOfFmlPsi}) will be discussed in Remark~\ref{rem:condPropagationNextToProbAssignment3}.

%Remark~\ref{rem:condPropagationNextToProbAssignment1}--\ref{rem:condPropagationNextToProbAssignment2}, 
%Remark~\ref{rem:condPropagationNextToProbAssignment2}
 Later in this section, we introduce the \emph{domain restriction} technique that can be used in~(\ref{eq:condPropagationNextToProbAssignment}). The technique utilizes (part of) $\fuzzyPred^{\bullet}$ to restrict the distribution $\Dist(\paramDist)$, preventing unnecessary samples from being generated at all.

% Note that there are two extremes in the choices of $\psi$: $\true$ and $\exists x\in \mathsf{supp}\bigl(\Dist(\paramDist)\bigr).\; \fuzzyPred^{\bullet}>0$.
% The former ($\true$) is simple but it means no condition is propaged further, limiting the effectivity of condition propagation. The latter is the most informative and restrictive, but its complexity---especially the quantifier therein---makes the subsequent reasoning harder. 

	 \item Let $l$ be a deterministic assignment location, and let  $(\vec{l^{\bullet}}, \fuzzyPred^{\bullet})=\overline{\CPpr}(\vec{l'})$. We define
	       \begin{align*}
		\overline{\CPpr}\bigl(\,l\xrightarrow{x \coloneqq e} \vec{l'}\,\bigr)
		\;\coloneqq\;
          % \bigl(\,l\xrightarrow{x \sim \bigl(\Dist(\paramDist)|\fuzzyPred^{\bullet}\bigr)} \vec{l^{\bullet}} \,,\,\psi\,\bigr)\enspace,
          \bigl(\,
		l\xrightarrow{x \coloneqq e} \vec{l^{\bullet}}
 \,,\;\fuzzyPred^{\bullet}[e/x]\,\bigr),
	       \end{align*}
	       where the predicate $\fuzzyPred^{\bullet}[e/x]$ is obtained by replacing every free occurrences of $x$
% in $\fuzzyPred^{\bullet}$ 
with the expression $e$.

	 \item Let $l$ be a weight location,  and let  $(\vec{l^{\bullet}}, \fuzzyPred^{\bullet})=\overline{\CPpr}(\vec{l'})$. We define
	       \begin{align*}
		&\overline{\CPpr}\bigl(\,l\xrightarrow{\weightClause{\fuzzyPred}} \vec{l'}\,\bigr)
		\;\coloneqq\;
          % \bigl(\,l\xrightarrow{x \sim \bigl(\Dist(\paramDist)|\fuzzyPred^{\bullet}\bigr)} \vec{l^{\bullet}} \,,\,\psi\,\bigr)\enspace,
          % \\&\qquad\qquad
		\bigl(\,
		% l\xrightarrow{\observeClause{\true }} 
		\vec{l^{\bullet}}
 \,,\;\fuzzyPred
% \mathbin{\mathtt{\&\&}}
 \times
\fuzzyPred^{\bullet}\,\bigr)\enspace,
	       \end{align*}
where the ``conjunction'' fuzzy predicate 
$\fuzzyPred
 \times
\fuzzyPred^{\bullet}$ is defined by 
\begin{math}
 \sem{\fuzzyPred
 \times
\fuzzyPred^{\bullet}}_{\sigma}
\coloneqq
 \sem{\fuzzyPred}_{\sigma}
 \times
\sem{\fuzzyPred^{\bullet}}_{\sigma}
\end{math}, where $\times$ on the right-hand side is the usual multiplication of reals, for each memory state  $\sigma$. 
\end{itemize}
Finally, the condition propagation operation $\CPpr$ on a straight-line program 
\;
\begin{math}
% (\flowl,\siginit,\efinal)
% \;=\;
%\bigl(\, 
\xrightarrow{\siginit} \linit\xrightarrow{\lambda_{1}}  \cdots\xrightarrow{\lambda_{n-1}} \lfinal\xrightarrow{\efinal}
%\,\bigr)
\end{math}
\;
%$(\flowl,\siginit,\efinal)$ 
is defined as follows. Let
% $(\vec{l^{\bullet}}, \fuzzyPred^{\bullet})=\overline{\CPpr}(\flowl)$. 
\begin{displaymath}
 \bigl(\,
l_{1}^{\bullet}\xrightarrow{\lambda_{1}^{\bullet}}  \cdots\xrightarrow{\lambda_{n^{\bullet}-1}^{\bullet}} \lfinal^{\bullet}\,,\;
 \fuzzyPred^{\bullet}
\,\bigr)
= 
\overline{\CPpr}\bigl(\,\linit\xrightarrow{\lambda_{1}}  \cdots\xrightarrow{\lambda_{n-1}} \lfinal\,\bigr)\enspace.
\end{displaymath}
Then we define, using a fresh initial location $\linit^{\bullet}$, 
\begin{align*}
& \CPpr\bigl(\,
\xrightarrow{\siginit} \linit\xrightarrow{\lambda_{1}}  \cdots\xrightarrow{\lambda_{n-1}} \lfinal\xrightarrow{\efinal}
%\flowl,\siginit,\efinal
\,\bigr)
\\ &\qquad\qquad\coloneqq\;
%\begin{cases}
\bigl(
\;\xrightarrow{\siginit}\linit^{\bullet}\xrightarrow{\weightClause{\fuzzyPred^{\bullet}}}
 l_{1}^{\bullet}\xrightarrow{\lambda_{1}^{\bullet}}  \cdots\xrightarrow{\lambda_{n^{\bullet}-1}^{\bullet}} \lfinal^{\bullet}
\xrightarrow{\efinal}\;
\bigr)\enspace.
% &\text{if $\siginit$ satisfies $\varphi^{\bullet}$,}
% \\
% \;\xrightarrow{\siginit}
%  \linit\xrightarrow{\observeClause{\false}}
% \lfinal^{\bullet}
% \xrightarrow{\efinal}
% &\text{otherwise.}
% \end{cases}
\end{align*}
\end{mydefinition}

\begin{myremark}\label{rem:intuitionOfDefOfCondPropTheProbAssgnmentCase}
 	       The definition~(\ref{eq:condPropagationNextToProbAssignment}) follows~\cite{NoriHRS14}; its intuition is as follows. 
 The continuation predicate $\fuzzyPred^{\bullet}$ is the observation that is passed over from the remaining part $\vec{l'}$ of the program.  
We would like to pass as much of the content of $\fuzzyPred^{\bullet}$  as possible to the further left (i.e.\ as the continuation predicate of $\overline{\CPpr}(l\xrightarrow{x \sim \Dist(\paramDist)} \vec{l'})$), since early conditioning should aid efficient sampling.
%\begin{itemize}
% \item 

 In case $x$ does not occur in $\fuzzyPred^{\bullet}$, the probabilistic assignment $x \sim \Dist(\paramDist)$ does not affect the value of  $\fuzzyPred^{\bullet}$, hence we can pass  the predicate $\fuzzyPred^{\bullet}$ itself to the left. This is the second case of~(\ref{eq:condPropagationNextToProbAssignment}). 

% \item
 If $x$  does occur in $\fuzzyPred^{\bullet}$, the conditioning is \emph{blocked} by the probabilistic assignment $x \sim \Dist(\paramDist)$, and the conditioning by $\fuzzyPred^{\bullet}$ at this stage is mandated. This results in the straight-line program in the first case of~(\ref{eq:condPropagationNextToProbAssignment}). 

However, it still makes sense to try to reject those earlier samples which would eventually ``violate'' $\fuzzyPred^{\bullet}$ (i.e.\ yield $0$ as the value of $\fuzzyPred^{\bullet}$, to be precise, since $\fuzzyPred^{\bullet}$ is fuzzy). The Boolean formula $\psi$ in~(\ref{eq:condPropagationNextToProbAssignment}) serves this purpose. 

The strongest choice for $\psi$ is the Boolean formula $\exists x\in \mathsf{supp}\bigl(\Dist(\paramDist)\bigr).\; \fuzzyPred^{\bullet}>0$ itself. However, the quantifier therein makes it  hard to deal with in implementation. The weakest choice of $\psi$ is $\true$, which means we do not pass any content of $\fuzzyPred^{\bullet}$ further to the left. 
% \end{itemize}
\end{myremark}

% The following remark, on~(\ref{eq:condPropagationNextToProbAssignment}--\ref{eq:choiceOfFmlPsi}), are important for efficient implementation of the proposed algorithm.
% \begin{myremark}\label{rem:condPropagationNextToProbAssignment1}
%  	       In general, there are multiple formulas $\psi$ that satisfy the condition~(\ref{eq:choiceOfFmlPsi}). We can restrict $\psi$ so that its free variables are those of $\fuzzyPred^{\bullet}$ but without $x$. It is not hard to see that, the stronger $\psi$ is, the more advantageous it is for efficient sampling: this is because a stronger $\psi$ will help to reject more samples earlier.  The following are some canonical choices of $\psi$.
% \begin{itemize}
%  \item 
%   The weakest choice is $\psi=\true $.
%  \item 
%  The strongest choice is $\psi = \bigl(\,\exists x\in \mathsf{supp}\bigl(\Dist(\paramDist)\bigr).\; \fuzzyPred^{\bullet}\,\bigr)$ itself, although for implemantation purposes one would probably like to avoid quantifiers. 
%  \item If $x$ does not appear in $\fuzzyPred^{\bullet}$, the stronger choice of $\psi$ boils down to $\fuzzyPred^{\bullet}$, which is quantifier-free (assuming $\fuzzyPred^{\bullet}$ is). 
% \end{itemize}
% \end{myremark}

\begin{myremark}\label{rem:condPropagationNextToProbAssignment3}
On the choice of a predicate $\psi$ in~(\ref{eq:choiceOfFmlPsi}), besides the two extremes discussed in Remark~\ref{rem:intuitionOfDefOfCondPropTheProbAssgnmentCase} ($\true$ and $\exists x\in \mathsf{supp}\bigl(\Dist(\paramDist)\bigr).\; \fuzzyPred^{\bullet}>0$), we find the following choice useful. 

Assume that the truth of $\fuzzyPred^{\bullet}>0$ is monotone in $x$, that is, $\fuzzyPred^{\bullet}(x_{1})>0$ and $x_{1}\le x_{2}$ imply $\fuzzyPred^{\bullet}(x_{2})>0$. Assume further that the support $\mathsf{supp}\bigl(\Dist(\paramDist)\bigr)$ has a supremum $x_{\mathrm{sup}}$. Then the implication
\begin{equation}\label{eq:psiViaSupport}
 		\bigl(\,\exists x\in \mathsf{supp}\bigl(\Dist(\paramDist)\bigr).\; \fuzzyPred^{\bullet}>0\,\bigr) \;\Longrightarrow\; \fuzzyPred^{\bullet}(x_{\mathrm{sup}})>0
\end{equation}
is obviously valid, making $\fuzzyPred^{\bullet}(x_{\mathrm{sup}})>0$ a viable candidate of $\psi$. 

Here is an example. Assume $\Dist(\paramDist)=\mathtt{Beta}(1,1)$, for which we have $\mathsf{supp}(\mathtt{Beta}(1,1))=[0,1)$. Let $\fuzzyPred^{\bullet} = \mathbf{1}_{x + z \ge 0}$, the characteristic function for the formula $x + z \ge 0$, where $z$ is another variable. Then we can take $\psi=(1+z\ge 0)$, that is, $\psi=(z \ge -1)$. 
\end{myremark}

\myparagraph{Domain Restriction}
The following is an improvement of $\CPpr$ (Definition~\ref{def:conditionPropagationPlain}). 
\begin{mydefinition}[$\CP$] \label{def:conditionPropagation}
 The operation $\CP$ is defined similarly to $\CPpr$ (Definition~\ref{def:conditionPropagationPlain})---using an extended  inductively-defined operation denoted by $\overline{\CP}$---except for the following difference.

The first case in~(\ref{eq:condPropagationNextToProbAssignment}) is now given by
\begin{equation}\label{eq:condPropagationNextToProbAssignmentOnlyX}
\begin{aligned}
&  		\overline{\CP}(l\xrightarrow{x \sim \Dist(\paramDist)} \vec{l'})
		\;\coloneqq\;
\\
&\qquad
           \bigl(\,l
              \xrightarrow{x \sim \bigl(\Dist(\paramDist)\,\big|\,\xi\bigr)}
	      l^{\sharp}
            \xrightarrow{\weightClause{{p(\xi\mid x \sim \Dist(\paramDist))}}}
              l^{*}
            \xrightarrow{\weightClause{\fuzzyPred^{\bullet}}}
               \vec{l^{\bullet}} \,,\,\mathbf{1}_{\psi}\,\bigr)\enspace,
\end{aligned}
\end{equation}
where
\begin{itemize}
 \item $\vec{l^{\bullet}}$ and $\fuzzyPred^{\bullet}$ are such that $(\vec{l^{\bullet}}, \fuzzyPred^{\bullet})=\overline{\CP}(\vec{l'})$, 
 \item  $l^{\sharp}$ and $l^{*}$ are fresh locations,
 \item $\xi$ is  a choice of a (sharp) Boolean formula such that the following is valid:
 \begin{equation}\label{eq:soundDomainRestriction}
  \forall x\in \mathsf{supp}\bigl(\Dist(\paramDist)\bigr).\;
   \bigl(\,\fuzzyPred^{\bullet}>0\,\Longrightarrow\, \xi\,\bigr),
 \end{equation}

 \item ${p(\xi\mid x \sim \Dist(\paramDist))}$ is the  fuzzy predicate that returns the probability % $p(\xi\mid x \sim \Dist(\paramDist))$ 
of a sample $x$ drawn from $\Dist(\paramDist)$ satisfying $\xi$, and
 \item  $\Dist(\paramDist)\mid\xi$ denotes the distribution $\Dist(\paramDist)$ conditioned by  $\xi$. Precisely, we obtain  $\Dist(\paramDist)\mid\xi$  by 1) first multiplying the density function $\mathbf{1}_{\xi(x)}$ and 2) normalizing it to a (proper, not sub-) distribution. See an example below.

% , whose (unnormalized) probability density function is given as follows:
% 		      \begin{displaymath}
% 		       \bigl(\Dist(\paramDist)\,\big|\,\xi\bigr)(x)
% 		       \;=\;
% 		       \begin{cases}
% 			% \frac{
% 			\Dist(\paramDist)(x)
% % }{%
% %                               \int \mathbf{1}_{\fuzzyPred^{\bullet}}\, \Dist(\paramDist)(x)\, \mathrm{d}x}  
% & \text{if $x$ satisfies $\xi$,}
% 			\\
% 			0 &\text{otherwise,}
% 		       \end{cases}
% 		      \end{displaymath}
 \item and $\psi$ is, much like in~(\ref{eq:choiceOfFmlPsi}), a (sharp) Boolean predicate such that the following is valid.
	       \begin{align*}
		\bigl(\,\exists x\in \mathsf{supp}\bigl(\Dist(\paramDist)\bigr).\; \fuzzyPred^{\bullet}>0\,\bigr) \;\Longrightarrow\; \psi
	       \end{align*}

\end{itemize}

\end{mydefinition}

% In the above definition of the operation $\CPpr$, 
% the first case  in~(\ref{eq:condPropagationNextToProbAssignment}) can be further transformed as follows: if $(\vec{l^{\bullet}}, \fuzzyPred^{\bullet})=\overline{\CP}(\vec{l'})$, 
% where

The transformation~(\ref{eq:condPropagationNextToProbAssignmentOnlyX}) restricts the domain of distribution $\Dist(\paramDist)$ by $\xi$; therefore we call this transformation \emph{domain restriction}. Its essence is really that of \emph{importance sampling}: in~(\ref{eq:condPropagationNextToProbAssignmentOnlyX}), the restriction to $\Dist(\paramDist)\mid\xi$ must be compensated by discounting the weights of the obtained samples, which is done by $\weightClause{{p(\xi\mid x \sim \Dist(\paramDist))}}$. 

An example is as follows.
% The distribution $\Dist(\paramDist)\mid\fuzzyPred^{\bullet}$ can have a simple representation that is easy to sample from, for some $\Dist(\paramDist)$ and $\fuzzyPred^{\bullet}$. 
 If $\Dist(\paramDist)=\mathtt{unif}(1,5)$ and $\fuzzyPred^{\bullet}=\mathbf{1}_{2\le x \le 4}(x)$, then we can choose $\xi$ to be $\xi=(2\le x \le 4)$, in which case the conditioned distribution $\Dist(\paramDist)\mid\xi$ is $\mathtt{unif}(2,4)$. In this way,  the transformation~(\ref{eq:condPropagationNextToProbAssignmentOnlyX}) allows direct sampling without conditioning. The subsequent ``discounting'' conditioning $\weightClause{{p(\xi\mid x \sim \Dist(\paramDist))}}$ (in the middle of the second line of~(\ref{eq:condPropagationNextToProbAssignmentOnlyX})) uses the discounting factor $p(\xi\mid x \sim \Dist(\paramDist))=p(2\le x\le 4\mid x\sim \mathtt{unif}(1,5))=1/2$. 

We note that sampling from a restricted distribution $\Dist(\paramDist)\mid\xi$ is often not hard. For example, assume that $\xi$ is given in the form of \emph{excluded intervals}, that is, 
\begin{equation}\label{eq:excludedIntervals}
 \xi = \bigl\{\,x\;\big|\; x\not\in (a_{1},b_{1}]\cup  (a_{2},b_{2}]\cup\cdots\cup  (a_{m},b_{m}]\,\bigr\},
\end{equation}
where $b_{1}\le a_{2},\dotsc, b_{m-1}\le a_{m}$. Assuming that we know the inverse $F$ of the cumulative density function  of $\Dist(\paramDist)$ (which is the case with most common distributions), we can
\begin{itemize}
 \item transform $F$ into the inverse $F_{\xi}$ of the CDF of the restricted distribution  $\Dist(\paramDist)\mid\xi$, via case distinction and rescaling, 
 \item and use this $F_{\xi}$ in the inverse transform sampling for the distribution  $\Dist(\paramDist)\mid\xi$.
\end{itemize} 
This method of inverse transform sampling from   $\Dist(\paramDist)\mid\xi$, when $\xi$ is given in the form of~(\ref{eq:excludedIntervals}), is implemented in our prototype $\Schism$.

\begin{figure*}[tbp]
  \centering\scriptsize
    \begin{minipage}{.22\textwidth}
      \footnotesize
     \begin{lstlisting}[%numbers=right,
      basicstyle={\scriptsize\ttfamily},escapechar=|,caption={\scriptsize\texttt{cond-prop-demo}},label={listing:condPropDemo}]
        x $\sim$ unif(0,20) ;
        while (x < 10) {
          y $\sim$ Beta(1,1);
          x := x + y;
        }
      \end{lstlisting}
      % \caption{Condition propagation in a loop}
      % \label{fig:ex:conditionPropagation:program}
    \end{minipage}
     \qquad\qquad
%    \begin{minipage}{.78\textwidth}
      \begin{minipage}{.17\textwidth}
      \begin{lstlisting}[numbers=left,basicstyle={\scriptsize\ttfamily}]
        x $\sim$ unif(0,20);

        obs(x < 10);
        y $\sim$ Beta(1,1);
        
        x := x + y;
        obs(x < 10);
        y $\sim$ Beta(1,1);
        
        x := x + y;
        obs(x < 10);
        y $\sim$ Beta(1,1);
        
        x := x + y;
        obs(10 $\leq$ x);
      \end{lstlisting}
      \end{minipage}
      \quad
      \begin{minipage}{.23\textwidth}
      % original
      % \begin{lstlisting}[basicstyle={\scriptsize\rmfamily}]
      %   // $\textnormal{no knowledge}$
      %   7 < x < 10, x $\sim$ unif(0,20)
      %   7 < x < 10
      %   8 < x + y < 10, 0 $\leq$ y < 1
      %   8 < x + y < 10
      %   8 < x < 10
      %   8 < x < 10
      %   9 < x + y < 10, 0 $\leq$ y < 1
      %   9 < x + y < 10
      %   9 < x < 10
      %   9 < x
      %   10 $\leq$ x + y, 0 $\leq$ y < 1
      %   10 $\leq$ x + y
      %   10 $\leq$ x
      % \end{lstlisting}
      \begin{lstlisting}[basicstyle={\scriptsize\rmfamily}]
        $7 < x < 10$
        $7 < x < 10, 0 \le x \le 20$
        $7 < x < 10$
        $7 < x < 10$
        $8 < x + y < 10, 0 \leq y < 1$
        $8 < x + y < 10$
        $8 < x < 10$
        $8 < x < 10$
        $9 < x + y < 10, 0 \leq y < 1$
        $9 < x + y < 10$
        $9 < x < 10$
        $9 < x$
        $10 \leq x + y, 0 \leq y < 1$
        $10 \leq x + y$
        $10 \leq x$
      \end{lstlisting}
      \end{minipage}
      \;
      \begin{minipage}{.21\textwidth}
        \begin{lstlisting}[basicstyle={\scriptsize\ttfamily},
             linebackgroundcolor={%
               \ifnum\value{lstnumber}=1
               \color{green!35}
               \fi
               \ifnum\value{lstnumber}=2
               \color{yellow!35}
               \fi
               \ifnum\value{lstnumber}=5
               \color{green!35}
               \fi
               \ifnum\value{lstnumber}=9
               \color{green!35}
               \fi
               \ifnum\value{lstnumber}=13
               \color{green!35}
               \fi}
          ]
          x $\sim$ unif(7,10);
          weight(3/20)
          // $\textnormal{no observation}$
          y $\sim$ Beta(1,1);
          obs(8 < x+y < 10);
          x := x + y;
          // $\textnormal{no observation}$
          y $\sim$ Beta(1,1);
          obs(9 < x+y < 10);
          x := x + y;
          // $\textnormal{no observation}$
          y $\sim$ Beta(1,1);
          obs(10 $\leq$ x+y);
          x := x + y;
          // $\textnormal{no observation}$
      \end{lstlisting}
      \end{minipage}
      \caption{condition propagation. In Program~\ref{listing:condPropDemo},  let
 $\flowl$ be the  control flow that runs the loop exactly three times. 
 The corresponding straight-line program $\StrLn(\flowl)$ is  the second column. 
The third column illustrates condition propagation, which should be read from bottom to top.   The last column is
 the resulting program
 $\CP(\StrLn(\flowl))$, where  changes are highlighted.
}
      \label{fig:ex:conditionPropagation:linear}
%    \end{minipage}
\end{figure*}
\begin{myexample}[application of $\CP$]\label{ex:conditionPropagation}
 See Figure~\ref{fig:ex:conditionPropagation:linear}, where a straight-line program (Column~2) gets optimized into the one in   Column~4 by the operation $\CP$.   Column~3 illustrates  condition propagation:  logical conditions get propagated  upwards, collecting observations. 

Many  propagation steps follow the weakest precondition
 calculus~\citep{Winskel93}.  The deterministic assignment commands (Lines~6, 10, 14) cause the   occurrences of $x$ in the conditions (Column~3) replaced by  $x+y$.  Observation commands add conditions to the propagated one; see Line~11.

When we encounter a probabilistic assignment (say Line~12),  we do the following: 1) we make an observation of the propagated condition ($10\le x+y$  in Line~13); 
2) we collect the support of the distribution as a new condition ($0\le y<1$ in Line~13); 
and 3) we compute a logical consequence $\psi$ of the last two conditions ($\psi=(9<x)$ in Line~12).  

For the probabilistic assignment in Line~1, we can moreover apply the domain restriction operation.  The propagated condition  $7<x<10$ allows us to restrict the original distribution
 \texttt{unif(0,20)}
 to \texttt{unif(7,10)}, as is done in Column~4. This way we spare generation of samples of $x$ that are eventually conditioned out. The idea is much like that of \emph{importance sampling}; similarly, we need to discount the weights of the obtained samples. This is done by the new $\mathtt{weight}$ command \texttt{weight(3/20)} in Line~2, Column~4, where \texttt{3/20} is the constant fuzzy predicate that returns the weight $3/20$. 
\end{myexample}

\begin{auxproof}
  For some control flows, condition propagation can even derive an impossible
 observation ($\observeClause{\false}$) at
 the very beginning,
 %which means that the algorithm can
 allowing us to
  simply skip the considered control flow. We use this for
 \emph{blacklisting} control flows;
 see Section~\ref{sec:implemenationAndExperiments}. 
\end{auxproof}

% One of the key ingredients of our main algorithm (Algorithm~\ref{algo:pCFGsampling}) is \emph{condition propagation}, that is, to back-propagate conditioning observations so that rejection of samples can be done earlier and more efficiently, or even better, not at all. The idea of condition propagation has been pursued in the probabilistic programming system R2~\cite{NoriHRS14,HurNRS15}.

% In our hierarchical sampling framework, we separate sampling of control
% flow from that of data values, thereby conductin data sampling from only
% straight-line programs. This allows us to apply condition propagation
% \emph{universally}, to an arbitrary probabilistic program. In contrast,
% the R2 framework~\cite{NoriHRS14} requires an invariant provided for
% each while loop.

\begin{myproposition}[soundness of $\CP$]\label{prop:correctnessConditionPropagation}
The operation  $\CP$ preserves the semantics of straight-line programs. That is, referring to  $\sem{\place}^{\wst}, \sem{\place}^{\st}$ from Section~\ref{subsec:pCFGSemantics},
\begin{align*}
 &\bigl\llbracket \,
(\flowl,\siginit,\efinal)
 \,\bigr\rrbracket^{\wst}
 \;=\;
 \bigl\llbracket \,
 \CP(\flowl,\siginit,\efinal)
 \,\bigr\rrbracket^{\wst}, \quad\text{and}
\\ 
 &\bigl\llbracket \,
(\flowl,\siginit,\efinal)
 \,\bigr\rrbracket^{\st}
 \;=\;
 \bigl\llbracket \,
 \CP(\flowl,\siginit,\efinal)
 \,\bigr\rrbracket^{\st}\enspace.
\end{align*}
\end{myproposition}
\begin{myproof}
 The first statement (coincidence of the weighted semantics, Definition~\ref{def:pCFGwstSemantics}) is shown easily by induction on the definition of $\overline{\CP}$. Restriction of $\psi$ (in~(\ref{eq:choiceOfFmlPsi})) to a sharp (i.e.\ $\{0,1\}$-valued) predicate  is crucial here: we rely on the nilpotency and idempotency of $0$ and $1$, respectively.  The second statement (on $\sem{\place}^{\st}$) follows from the first. \qed
\end{myproof}

\section{A Hierarchical Sampling Algorithm}\label{sec:algorithm}
The architecture in Figure~\ref{fig:hierarchicalSamplerArch}
%features separation of control flow sampling and data sampling. It
arises from the following equality.
%equality~(\ref{eq:equationForControlDataSeparation}). 
\begin{equation}\label{eq:equationForControlDataSeparation}
% \footnotesize
\begin{array}[t]{l}
    p(\sem{\efinal}_{\sigma_{N}}\in\mathrm{d}v\mid\Gamma)
\\
=
\int_{l_{1:N}}
 \Bigl(
 \textstyle\int_{\sigma_{1:N}}
 p\bigl(\sem{\efinal}_{\sigma_{N}}\in\mathrm{d}v\mid \sigma_{N}\bigr)
 \cdot
 p(
 \mathrm{d} \sigma_{1:N}
 \mid
  l_{1:N},\Gamma
 )\,
 % \Bigr.
 % \mathrm{d}\sigma_{1:N}
 \Bigr)\,
 % \\
 % \qquad\qquad\qquad\qquad\qquad\qquad
 p(
 \mathrm{d}l_{1:N}\mid\Gamma
 )
 % \;
 % \mathrm{d}l_{1:N}\,
\end{array}
% \begin{array}[t]{l}
% \\
%     p(v=\sem{\efinal}_{\sigma_{N}}\mid\Gamma)
% \\
% =
% \int
%  \Bigl(
%  \textstyle\int
%  p(v=\sem{\efinal}_{\sigma_{N}}\mid \sigma_{N})\,
%  p(
%  \sigma_{1:N}
%  \mid
%   l_{1:N},\Gamma
%  )\,
%  % \Bigr.
%  \mathrm{d}\sigma_{1:N}
%  \Bigr)\,
%  % \\
%  % \qquad\qquad\qquad\qquad\qquad\qquad
%  p(
%  l_{1:N}\mid\Gamma
%  )
%  \;
%  \mathrm{d}l_{1:N}\,
% \end{array}
% \!\!\!\!\!\!\!
\end{equation}
The left-hand side is the probability we want---the probability of the return expression $\efinal$ belonging to a certain measurable set $\mathrm{d}v$,
% $\sem{\efinal}_{\sigma_{N}}$, 
under the final memory state $\sigma_{N}$ that is sampled from the execution of the pCFG $\Gamma$. It is expressed using two nested integrals: the inner integral is over data samples $\sigma_{1:N}$ under a fixed control flow $l_{1:N}$; and the outer integral is over control flows $l_{1:N}$ of $\Gamma$. 
The proof of~(\ref{eq:equationForControlDataSeparation}) is by marginalization and conditional probabilities; see  \ref{appendix:derivation}. Note also that the set of complete control flows $l_{1:N}$ is countable; therefore the outer integral can also be expressed simply as an infinite sum. 

We estimate  the two integrals in~(\ref{eq:equationForControlDataSeparation}) by sampling.
We use SMC for the bottom-level data sampling, i.e.\ for the inner integral in~(\ref{eq:equationForControlDataSeparation}). See Section~\ref{subsec:mainAlgorithm}.   In Section~\ref{subsec:multiArmedSamplingForControlFlows}, we discuss the top level.

\subsection{The Infinite-Armed Sampling Problem}\label{subsec:multiArmedSamplingForControlFlows}
The top-level control flow sampling  (Figure~\ref{fig:hierarchicalSamplerArch}) is formulated as   \emph{infinite-armed sampling (IAS)}, a problem we shall now describe. Formal definitions and proofs are in \ref{appendix:multiArmedSampling}. The problem is  a variation of the classic problem of multi-armed bandit (MAB) (see e.g.~\cite{Slivkins19} for an introduction). In the instance of the problem that we use, an arm will be a complete control flow. See Section~\ref{subsec:mainAlgorithm} for details.

In the infinite-armed sampling problem, infinite arms $\{1,2,\dotsc\}$ are given, and each arm $k$ is given a real number  $\truelik{k}$ called its \emph{likelihood}. Our goal is to sample arms, pulling one arm  each time, so that the resulting histograms converge to (the normalization of) the distribution $(\truelik{1}, \truelik{2}, \dotsc)$. 
The challenge, however, is that the likelihoods $\truelik{1}, \truelik{2}, \dotsc$ are not know a priori. We assume that, when we pull an arm $k$ at time $t$, we sample a random variable $X_{k,t}$ whose mean is the (unknown) true likelihood $\truelik{k}$. The random variables $X_{k,1}, X_{k,2}, \dotsc$ are i.i.d.

 The IAS problem is an \emph{infinite} and \emph{sampling} variant of 
%the
 multi-armed bandit
% problem
  (MAB). The goal in MAB is to optimize, while our goal is to sample. 
The IAS problem indeed describes the top-level sampling in Figure~\ref{fig:hierarchicalSamplerArch}: a control flow is an arm; there are countably many of them in general; and the likelihood of each control flow is only estimated by sampling the corresponding straight-line program and measuring the weights of the samples. 
Further discussions are found in Section~\ref{subsec:mainAlgorithm}.

Algorithm~\ref{algo:infinitelyArmedSampling} is our algorithm for
 the IAS problem. It is an adaptation of the well-known  \emph{$\varepsilon$-greedy algorithm} for MAB. In each iteration, it conducts one of the  following: (\textbf{proportional sampling}, Line~\ref{line:proportional}) sampling a known arm according to the empirical likelihoods; (\textbf{random sampling}, Line~\ref{line:random})  sampling a known arm uniformly randomly; and (\textbf{expansion}, Line~\ref{line:discovernew1}) sampling an unknown arm and making it known. In Line~\ref{line:updateEmpLkl}, the empirical likelihood $\hat{p}_{k}$ is updated using the newly observed likelihood $p$, so that the result is the mean of all the likelihoods of $k$ observed so far. 

Comparing to the original $\varepsilon$-greedy algorithm for MAB,  proportional sampling corresponds to the \emph{exploitation} action, while random sampling corresponds to the \emph{exploration} action. 
% More specifically, at time $t$, we do expansion if the number of known arms are smaller than $t^{2/3}$. Otherwise, we do random sampling with  probability $\varepsilon_{t}=(\frac{K\log t}{t})^{\frac{1}{3}}$, and do proportional sampling with  probability $1-\varepsilon_{t}$. 
The \emph{exploration rate} $\varepsilon_{t}=(\frac
{K\log t}{t})^{\frac{1}{3}}$ in Algorithm~\ref{algo:infinitelyArmedSampling} is the one commonly used for MAB. See e.g.~\citep{Slivkins19}. 

We give a theoretical guarantee, restricting to the \emph{finite}-armed setting. Its proof is in \ref{appendix:multiArmedSampling}. 
\begin{mytheorem}[convergence, finite-armed]\label{thm:conv}
 In Algorithm~\ref{algo:infinitelyArmedSampling}, assume that $\mathcal{K}=\{1,\dotsc, K\}$, $\mathcal{K}_{\text{known}}$ is initialized to $\mathcal{K}$, and no expansion is conducted. 
Then Algorithm~\ref{algo:infinitelyArmedSampling} satisfies, for each arm
$k\in \mathcal{K}$,
\begin{math}
\textstyle
 \bigl|\,\frac{\mathbb{E}(T_{k}(T)
)}{T} - \frac{p_{k}}{\sum_{k}p_{k}}
  \,\bigr| =O\bigl(K^{\frac{7}{3}}\bigl(\frac{\log T}{T}\bigr)^{\frac{1}{4}}\bigr)
% \quad\text{for each $k\in \mathcal{K}$.}
\end{math}, where
$T_{k}(T)
=|\{t\mid k^{(t)}=k\}|
$ is how often the arm $k$ is pulled in time $T$. \qed
\end{mytheorem}
We note that the above convergence is slower than in the MAB case (optimization, see~\citep{Slivkins19}).   

Extension of the above convergence theorem to infinite arms is  future work. Infinite-arm variations of multi-armed bandit have been studied in many works, see e.g.~\cite{WangAM08,CarpentierV15,KimVY22}. We believe that their proof techniques can be ported to our sampling (as opposed to optimization) problem. Bandits with a continuum of arms have been studied, too~\cite{Agrawal95}. 
% It would require some assumption on the distribution  $(p_{1}, p_{2},\dotsc)$, which is hard to check for probabilistic programs.

 %%%%%%%%%%%%%%%%%%%%%%%%%%%%%%%%%%%%%%%%%%%%%%%%%%%%%%%%%%%%%%%%%%%
\begin{figure*}[tbp]
 % \begin{adjustbox}{scale=.9}
 % \begin{minipage}[t]{7.2cm}
% \null
 \footnotesize
  \begin{algorithm}[H]
  \caption{Our $\varepsilon$-greedy algorithm for infinite-armed sampling. Here
  $\varepsilon_{t}=(\frac{|\mathcal{K}_{\text{known}}|\log t}{t})^{\frac{1}{3}}$ 
  }
  \label{algo:infinitelyArmedSampling}
  \begin{algorithmic}[1]
  \Require
  arms $\mathcal{K}=\{1, 2, \dotsc\}$, budget $T$
  % an infinite set $\mathcal{K}$ of arms. Each $k\in \mathcal{K}$ has its (true, not necessarily normalized) \emph{likelihood} $p(k)\in \Rnn$. The \emph{round count} $t$ is initialized to $0$
  \Ensure sequence  $k^{(1)},k^{(2)},\dotsc, k^{(T)}$ of arms
  \State $t\gets 1$;  $\mathcal{K}_{\text{known}}\gets\emptyset$;
  \State      $\hat{p}_{k}\gets 0$ \text{ for $\forall k\in\mathcal{K}$}
     \Comment{empirical likelihood}
  \While{$t\le T$}
  \If{$|\mathcal{K}_{\text{known}}|<t^{2/3}$} 
    \Comment{\emph{expand}} \label{line:decideWhetherToExpand}
    \State pick  $k\not\in \mathcal{K}_{\text{known}}$, add $k$ to $\mathcal{K}_{\text{known}}$
  \label{line:discovernew1}
    %        pick any $k\not\in \mathcal{K}_{\text{known}}(t)$  
    % \State \label{line:discovernew2} $\mathcal{K}_{\text{known}}(t+1)\gets \mathcal{K}_{\text{known}}(t)\cup\{k\}$
  \ElsIf{(with prob.\ $\varepsilon_{t}$)}   \Comment{\emph{random}}
    \State\label{line:random}
    pick $k\in\mathcal{K}_{\text{known}}$ unif.\ randomly 
  \Else  \Comment{\emph{proportional}}
    \State\label{line:proportional}
    pick $k\in\mathcal{K}_{\text{known}}$ proportionally to $(\hat{p}_{k})_{k}$

  %    \State \label{line:pullKnown1} pick
  % $k %\in \mathcal{K}_{\text{known}}(t)
  %  $
  %     from 
  % \begin{math}
  %  \left[\,
  %  k\mapsto \UCBV(k,t)
  %  \,\right]_{k\in \mathcal{K}_{\text{known}}}
  % \end{math}
  %   \State  \label{line:pullKnown2}
  %   \parbox[t]{.83\textwidth}{pull the arm $k$, obtain an observed
  % likelihood $X_{k,t}$, and update $E(k,t),V(k,t), N(k,t)$ accordingly}
  \EndIf
  \State pull arm $k$ and observe a likelihood $p$ \label{line:pullArm}
  \setcounter{ALG@line}{12}
  \State 
    $\hat{p}_{k}\gets \displaystyle\frac{p + \hat{p}_{k}\cdot T_{k}(t-1)}{1+T_{k}(t-1)}$
          \label{line:updateEmpLkl}
  \Statex
  \Comment{
\begin{tabular}[t]{l}
    update the empirical likelihood $\hat{p}_{k}$ using $p$,
   \\
    where $T_{k}(t-1)=|\{t'\in [1,t-1]\mid k^{(t')}=k\}|$ 
    \\
    is how often the arm $k$ is pulled so far
\end{tabular}
}
  \State  $k^{(t)}\gets k$; \; $t\gets t+1$
  \EndWhile
  \end{algorithmic}
 \end{algorithm}
 % \end{minipage} 
% \end{adjustbox}			
 %%%%%%%%%%%%%%%%%%%%%%%%%%%%%%%%%%%%%%%%%%%%%%%%%%%%%%%%%%%%%%%%%%%

\vspace{1em}

 % \begin{adjustbox}{
 % %max width=.6\textwidth
 %  scale=.9}
 % \begin{minipage}[t]{.67\textwidth}
 \begin{algorithm}[H]
 \caption{Our hierarchical sampler
 % for $\PIMP$ programs
 }
 \label{algo:ourHierarchicalSampler}
 \begin{algorithmic}[1]
 \Require
 a pCFG $\Gamma$, constant $J$ 
  \Ensure
 a sequence $\overrightarrow{(w',v)}$ of weighted samples

 \vspace{.5em}

 % \Statex
 % \parbox[t]{\linewidth}{The same as Algorithm~\ref{algo:infinitelyArmedSampling}, where $\mathcal{K}$ is the set of complete control flows of $\Gamma$, and Line~\ref{line:pullArm} is refined into  the following two steps.
 % }

\Statex Take Algorithm~\ref{algo:infinitelyArmedSampling}, and apply the following adaptation: 1) $\mathcal{K}$ is the set of complete control flows of $\Gamma$; 2) we maintain a pool of weighted samples $\overrightarrow{(k,w,v)}$, initialized to empty;   3) we refine Line~\ref{line:pullArm}  of Algorithm~\ref{algo:infinitelyArmedSampling} into   Lines~\ref{line:drawSamples}--\ref{line:appendSamples} shown below; and 4) we add Line~\ref{line:weightAdjustment}, shown below, for adjusting weights

 \vspace{.25em}
\hrule

 \vspace{.25em}
\setcounter{ALG@line}{9}
\State \label{line:drawSamples}%
draw $J$ weighted samples $(w_{t,1:J},v_{t,1:J})$ 
      from the  weighted semantics 
 $\sem{\StrLn(k)}^{\wst}$
$
\in \subdist(
\Rnn 
\times\DDom)
$ (cf.~(\ref{eq:weightedSemTypeWhole})). 
Concretely, this is  by running SMC for the program 
$\CP(\StrLn(k))$
% .
% We assume that at least one of $w_{t,1:J}$ is nonzero
\State $p\gets \sum w_{t,1:J}/J$; 
\Comment{used in Line~\ref{line:updateEmpLkl}}
% \Statex\Comment  \parbox[t]{.9\linewidth}{resampling forces $w_{t,1}=\dotsc = w_{t,J}$ }
\State 
% append $x_{t,1:J}$ to the sample pool $\vec{x}$
append $(k,w_{t,1:J},v_{t,1:J})$ to the pool $\overrightarrow{(k,w,v)}$ (we also record the flow $k$)
% (where we also record the control flow $k$)
 \label{line:appendSamples}

\vspace{.25em}
\hrule

\vspace{.25em}
\setcounter{ALG@line}{14}

\State \label{line:weightAdjustment}%
let each $(k,w,v)$ in $\overrightarrow{(k,w,v)}$ induce $(\frac{w}{\hat{p}_{k}},v)$ in the output samples
$\overrightarrow{(w',v)}$ 
% samples $\overrightarrow{(w',x)}$ 
%  in the pool $\overrightarrow{(k,w,x)}$, replace each  $(k,w,x)$ with
%  % $(\frac{w}{\hat{p}_{k}},x)$ 
% $(\hat{p}_{k}\cdot \frac{w}{\hat{w}_{k}},x)$
% and obtain the output $\overrightarrow{(w',x)}$, where $\hat{w}_{k}=\sum_{(k,w',x')\in\overrightarrow{(k,w,x)}}w'$ is the sum of the weights associated to $k$

%  \vspace{.5em}
%  \Statex 
%  \parbox[t]{\linewidth}{
%  \textbf{(\ref{line:pullArm}': flow likelihood estim.)} Draw $J$ weighted samples $(w_{t,1:J},x'_{t,1:J})$ from the  weighted sem.\ $\sem{\StrLn(k)}^{\wst}$ in~(\ref{eq:weightedSemTypeWhole}), and 

% take the weights $w_{t,1:J}$ as observed likelihoods, and update $E(k,t),V(k,t), N(k,t)$ accordingly
%  }

%  \vspace{.5em}
%  \Statex
%  \parbox[t]{\linewidth}{
%  \textbf{(7'':  data sampling)} Draw $J$  samples $x_{t,1:K}$ from the \emph{normalized} sem.\ $\sem{\CP(\StrLn(k))}^{\st}$ in~(\ref{eq:unweightedSemTypeWhole})
%  }

%  \vspace{.5em}
%  \Statex Each round $t$ produces $K$  samples $x_{t,1:K}\in\R^{K}$. We return the collection $x_{1:T,1:K}$ of these samples. 

 \end{algorithmic}
 \end{algorithm}
%  \end{minipage}
% \end{adjustbox}
\end{figure*}

\begin{auxproof}
Details are in  \url{ichiro-note/infiniteArmedSamplingTrialsSep2019.tex}
\end{auxproof}

\begin{auxproof}
 We identify the (top-level) sampling problem of control flows as an \emph{infinite-armed sampling} problem. It is much like the \emph{multi-armed bandit} problem, a problem that is heavily studied in the reinforcement-learning literature. There are however two main differences.

 Firstly, the goal of the multi-armed bandit problem is to \emph{optimize}, while ours is to \emph{sample}, in both cases among those arms whose 
 true performance (or likelihood, respectively) is unknown. This difference is minor, since the so-called \emph{Gumbel max trick} allows us to turn a sampling problem into optimization. See~\cite{Yellott77,KuzminW05}.

 Secondly, we have \emph{infinitely many} arms, while multi-armed bandits typically have only finitely many arms. Indeed, an arm for us is a complete control flow; and presence of a loop necessarily yields infinitely many of them (Figure~\ref{fig:pCFGexample1}). We therefore use an algorithm from~\cite{WangAM08} that adapts upper-confidence-bound (UCB) methods to infinite arms. More specifically, we use the algorithm in Section~2.3 of~\cite{WangAM08} that has the anytime property. It alternates between 1) the usual UCB-trials among the \emph{known} arms; and 2) discovery of  \emph{unknown} arms.

 We sketch our adaptation of  the algorithm in~\cite{WangAM08} to sampling in Algorithm~\ref{algo:infinitelyArmedSamplingForControlFlows}. Combined with the (bottom-level) data sampling, it will give our hierarchical sampling algorithm (Algorithm~\ref{algo:ourHierarchicalSampler}). 

 Here are some explanations. Algorithm~\ref{algo:infinitelyArmedSamplingForControlFlows} combines a usual UCB-type algorithm (Lines~\ref{line:pullKnown1}--\ref{line:pullKnown2}) with discovery of new arms (Lines~\ref{line:discovernew1}). 
 Pulling an arm $k$ (Line~\ref{line:pullKnown2}) yields
 an \emph{observed likelihood} $X_{k,t}$, whose mean is the given likelihood $p(k)$.

 Throughout the algorithm we maintain the following data, for each round $t\in \Zpos$:
 1)
 the set $\mathcal{K}_{\text{known}}(t)\subseteq \mathcal{K}$ of already tried arms; 
 2)  the \emph{empirical mean} $E(k,t)\in \Rnn$, and the \emph{empirical variance} $V(k,t)\in \Rnn$,  of the previously observed likelihoods of each arm $k\in\mathcal{K}_{\text{known}}(t)$; and
 3) the \emph{visit count} $N(k,t)\in \Zpos$ of how many times each arm $k\in\mathcal{K}_{\text{known}}(t)$ has been tried. These data are suitably initialized at $t=0$. 

 In Line~\ref{line:pullKnown1}, the \emph{UCB-V score} $\UCBV(k,t)$ of a known arm $k\in\mathcal{K}_{\text{known}}$ be defined as follows. 
 \begin{equation}\label{eq:UCBVscore}
 \begin{aligned}
 &    \textstyle \UCBV(k,t) = E(k,t) + \sqrt{\frac{2\,V(k,t) \,\decayFn(t)}{N(k,t)}} + \frac{3\,\decayFn(t)}{N(k,t)},
 % \\
 % &\qquad\text{where }
 %  \decayFn
 \end{aligned} 
 \end{equation}
 where the \emph{decay function} $\decayFn$ is defined by
 \begin{math}
 \decayFn(t)= C_{1}\log(1+C_{2}\log t)
 \end{math}, 
 using parameters $C_{1}, C_{2}$. 

 Note that there are two kinds of ``exploration'': trying a known arm ($k$ can be chosen even if its empirical likelihood is zero, thanks to  the latter two terms in~(\ref{eq:UCBVscore})); and trying a new arm. In both cases, the exploration factors decay over time, along $\decayFn(t)$ and $\sqrt{t}$ (see Line~\ref{line:decideWhetherToExpand}), respectively. 
\end{auxproof}

\begin{myremark}\label{rem:ZhouEtAlFlowSampling}
 In~\cite{ZhouYTR20}, sampling  ``sub-programs'' (they correspond to our control flows) is thought of as a  problem of \emph{resource allocation}. Their solution is a UCB-based algorithm that prioritizes those sub-programs whose likelihood samples have a larger variance. While detailed comparison is future work, one can conceptually argue  that our IAS flow sampling is a viable alternative. See \ref{appendix:ZhouEtAlFlowSampling}.
% for further discussions.
\end{myremark}

\subsection{Our Hierarchical Sampling Algorithm}
\label{subsec:mainAlgorithm}
Our hierarchical sampling algorithm is 
 Algorithm~\ref{algo:ourHierarchicalSampler}. It refines Algorithm~\ref{algo:infinitelyArmedSampling}. 
Here are some highlights. 

\subsubsection{Use of SMC for Data Sampling}\label{subsubsec:whySMC}
We use sequential Monte Carlo (SMC) for the purpose of data sampling (the bottom level of Figure~\ref{fig:hierarchicalSamplerArch}). In particular, we do not use Markov chain Monte Carlo (MCMC), another class of well-accepted sampling algorithms. 

The reason is that, for our current purpose, we have to estimate the \emph{normalizing constant} (also called the \emph{Bayesian marginal likelihood}) of the distribution we are sampling from. This is trivial with SMC, as described below. In contrast, MCMC per se does not provide means to estimate normalizing constants; one has to rely on an external method, such as Chib's method. 

A detailed introduction of SMC is out of the scope of the paper and is deferred e.g.\ to~\cite{DoucetJ11tutorial}. A property of SMC that is important for us is that each sample in SMC carries not only its value $v$ but its weight $w$, that is, that each sample is of the form $(w, v)$. This is very much like in the weighted semantics of pCFGs in Section~\ref{subsec:pCFGSemantics}. See Example~\ref{ex:pCFGSemExample}; by SMC sampling, we can obtain a sample $(0,\false)$ whose weight is $0$ and thus does not contribute to the estimation of the posterior.

\subsubsection{Estimating Control Flow Likelihoods by SMC} 
In Algorithm~\ref{algo:ourHierarchicalSampler}, an arm $k=\flowl$ is a complete control flow. The algorithm is designed so that different flows $\flowl$ are pulled in proportion to the their likelihood $p_{\vec{l}}=p(\flowl)$ in  $\Gamma$'s execution. This  $p(\flowl)$ is expressed as follows; a proof is in \ref{appendix:justificationOfAuxSampling}. 
%\vspace{-1em}
\begin{equation}\label{eq:justifyAuxSampling}
 \begin{aligned}
  p(l_{1:N}\mid\Gamma)
 = &\textstyle
% \\
% \quad
  \int_{\sigma_{1:N}}\,p(\mathrm{d}\sigma_{1},l_{1}\mid\Gamma) 
   \,
  \bigl(\,
  \prod_{k=2}^{N} p(l_{k}\mid \sigma_{k-1}, l_{k-1},\Gamma)
  \,\bigr)
  \\
  &\quad\bigl(\,\textstyle
  \prod_{k=2}^{N} p(\mathrm{d}\sigma_{k}\mid \sigma_{k-1},l_{k-1:k},\Gamma)
  \,\bigr)
 % \,\mathrm{d}\sigma_{1:N} 
 \end{aligned}
\end{equation}
It can be shown by induction on $N$ that the right-hand side is
\begin{math}
 \int_{w,v}\,w\cdot\bigl(\sem{\StrLn(l_{1:N})}^{\wst}(\mathrm{d}w\times\mathrm{d}v)\bigr)
% \,\mathrm{d}w\,\mathrm{d}v
\end{math}. This value is estimated in Algorithm~\ref{algo:ourHierarchicalSampler} by sampling $(w_{t,1:J}, v_{t,1:J})$ from the weighted semantics $\sem{\StrLn(l_{1:N})}^{\wst}$ (from~(\ref{eq:weightedSemTypeWhole})) and taking the average of $w_{t, 1:J}$. Formally, 
\begin{align*}
  p(l_{1:N}\mid\Gamma)
 &=
   \textstyle\int_{w,v}\,w\cdot\bigl(\sem{\StrLn(l_{1:N})}^{\wst}(\mathrm{d}w\times\mathrm{d}v)\bigr)
 \qquad\text{by def.\ of $\sem{\StrLn(l_{1:N})}^{\wst}$}
 \\
 &\sim
  \frac{1}{J}\sum_{j=1}^{J} w_{j}
  \quad\text{where $J$ samples $(w_{1:J}, v_{1:J})$ are sampled from $\sem{\StrLn(l_{1:N})}^{\wst}$.}
\end{align*}
Note that SMC (unlike MCMC) allows direct sampling of \emph{weighted} values, and is thus suited for sampling from the weighted semantics $\sem{\StrLn(l_{1:N})}^{\wst}$, as we discussed in Section~\ref{subsubsec:whySMC}. 
This is what we do in Lines~10--11, using SMC. 

\subsubsection{Data Sampling} 
After estimating the  flow likelihood $p(l_{1:N}\mid\Gamma)$, we sample values of $\sem{\efinal}_{\sigma_{N}}$ from the inner integral
 in~(\ref{eq:equationForControlDataSeparation}).
 The inner integral is proportional to
\begin{math}
 \int\,w \cdot\bigl(\sem{\StrLn(l_{1:N})}^{\wst}(\mathrm{d}w\times\mathrm{d}v)\bigr)
%\,\mathrm{d}w%\mathrm{d}x
\end{math}, where  $\mathrm{d}v$ is a choice of a measurable subset in~(\ref{eq:equationForControlDataSeparation}) that we expect  the value $\sem{\efinal}_{\sigma_{N}}$ to belong to. Therefore,  samples of $\sem{\efinal}_{\sigma_{N}}$  can be given by sampling $(w_{t,1:J}, v_{t,1:J})$ from $\sem{\StrLn(l_{1:N})}^{\wst}$---which we already did in Line~10---and weighting the sample values $v_{t,1:J}$ by  $w_{t,1:J}$. 
% We assume that resampling in SMC forces $w_{t,1}=\dotsc = w_{t,J}$; then $x_{t,1:J}$ are themselves valid data samples, justifying Line~12. 
This justifies Line~\ref{line:appendSamples}.

\subsubsection{Weight Adjustment}
In Line~\ref{line:weightAdjustment}, from the weight $w$ of each $(k,w,v)$, we discount the empirical likelihood  $\hat{p}_{k}$ of $k$,
% ---its empirical approximate to be precise---
since it is already accounted for by  the frequency of $k$ in $\overrightarrow{(k,w,v)}$. 

\begin{auxproof}
 For faster convergence, an alternative to Line~\ref{line:weightAdjustment} is 
 % in the pool $\overrightarrow{(k,w,v)}$, 
 to replace each  $(k,w,v)$ with
 % $(\frac{w}{\hat{p}_{k}},v)$ 
 $(\hat{p}_{k}\cdot \frac{w}{\hat{w}_{k}},v)$
 to obtain the output $\overrightarrow{(w',v)}$, where $\hat{w}_{k}=\sum_{(k,w',x')\in\overrightarrow{(k,w,v)}}w'$ is the sum of the weights associated to $k$. This means that we use the infinite-armed sampling as a proposal in importance sampling. This is a good proposal distribution, as shown by the convergence discussed in Section~\ref{subsec:multiArmedSamplingForControlFlows}.
\end{auxproof}

\subsubsection{Logical Blacklisting of Control Flows}
In Algorithm~\ref{algo:ourHierarchicalSampler}, running Line~\ref{line:drawSamples} for a flow $k$ with zero likelihood is useless. We adopt \emph{logical blacklisting} to soundly remove some of such flows: if condition propagation of a flow $k$'s straight-line program $\StrLn(k)$ exposes \verb|obs(false)|, the flow $k$ is deemed logically infeasible and never picked henceforth. In our implementation, we apply logical blacklisting also to incomplete control flows---see Section~\ref{sec:implemenationAndExperiments} later, especially Table~\ref{table:collectiveResults}.

\begin{auxproof}
 \begin{wrapfigure}[7]{r}{0pt}
 \!\!\!\!\!\!\!\!\!\!\!\!
 \begin{tabular}{c}
 \scalebox{.8}{      \begin{tikzpicture}[level distance=3em,sibling distance=.5em]
 %\Tree [.S [.NP LaTeX ] [.VP [.V is ] [.NP fun ] ] ]
 % \Tree[.23 [. [.\node[draw]{2}; ] [.\node[draw]{3}; ] ] [. [.\node[draw]{1}; ] %\edge[draw=none];
 %       [.\node[draw]{4}; ] ] ]
 %
 \Tree[.$\bullet$  [.\node[draw]{2}; ] [.\node[draw]{1}; ] [.\node[draw]{3}; ] %\edge[draw=none];
       [.\node[draw]{4}; ]  ]
       \end{tikzpicture}
 }
 \\
 $\Longrightarrow$
 \raisebox{-.5\height}{      
 \scalebox{.8}{\begin{tikzpicture}[level distance=2em,sibling distance=.3em]
 %\Tree [.S [.NP LaTeX ] [.VP [.V is ] [.NP fun ] ] ]
 % \Tree[.23 [. [.\node[draw]{2}; ] [.\node[draw]{3}; ] ] [. [.\node[draw]{1}; ] %\edge[draw=none];
 %       [.\node[draw]{4}; ] ] ]
 %
 \Tree[.$\bullet$ [.3 [.\node[draw]{2}; ] [.\node[draw]{1}; ] ] [.7 [.\node[draw]{3}; ] %\edge[draw=none];
       [.\node[draw]{4}; ] ] ]
       \end{tikzpicture}
 }}
 \end{tabular}
 % \caption{Sampling arms via a tree}
 % \label{fig:UCBOrganizedAsATree}
 \end{wrapfigure}
 \textbf{(Tree-shaped UCB)} 
 The UCB-style sampling (Lines~\ref{line:pullKnown1}--\ref{line:pullKnown2} of Algorithm~\ref{algo:infinitelyArmedSamplingForControlFlows}) is in fact organized as a tree in our implementation. We do so for efficiency. See the figures on the right: the top figure has four arms, with their labels $2,1,\dotsc$ designating their UCB-V scores. In our implementation, these arms are organized as the tree below, where each internal node is labeled with the sum of the UCB-V scores of all the leaves below it. The tree structure  used here naturally arises from unraveling a pCFG. This organization of UCB in a tree makes our control-flow sampling look like \emph{Monte Carlo tree search} (MCTS); the difference is that we do not employ \emph{playout}, a critical procedure in MCTS that estimates the reward of an internal node.

\end{auxproof}

\begin{auxproof}
 \begin{myexample}\label{ex:hierarchicalSamplingEx2}
 Consider the pCFG in Figure~\ref{fig:pCFGexample2}. There are four complete control flows, each of which corresponding to the choices between $\{l^{(2)}, l^{(3)}\}$ and
 $\{l^{(5)}, l^{(6)}\}$. Let us therefore denote these complete control flows by $\CCF(\Gamma)=\{\flowl^{2,5},\flowl^{2,6},\flowl^{3,5}, \flowl^{3,6}\}$. 

 Let us first focus on the bottom, data-sampling level of our algorithm (Line~\ref{line:pCFGDataSampling}). Constant propagation yields the following straight-line programs. 
 See also Example~\ref{ex:strLn2}. 
 \begin{align*}
 \CP\bigl(\StrLn(\flowl^{2,6})\bigr)
 &=
 \bigl(\,
  \xrightarrow{\siginit}
 \linit
 \xrightarrow{1}
  l^{(2)}
 \xrightarrow{ c1 := \true}
  l^{(4)}
 \xrightarrow{1}
 l^{(6)}
 \xrightarrow{ c2 := \false}
 l^{(7)}
  \xrightarrow{\observeClause{\true}}
 %\xrightarrow{{\color{red}\observeClause{\true }}}
 \lfinal
 \xrightarrow{c1}
 \,
 \bigr)
 \\
 \CP\bigl(\StrLn(\flowl^{3,5})\bigr)
 &=
 \bigl(\,
  \xrightarrow{\siginit}
 \linit
 \xrightarrow{1}
  l^{(3)}
 \xrightarrow{ c1 := \false}
  l^{(4)}
 \xrightarrow{1}
 l^{(5)}
 \xrightarrow{ c2 := \true}
 l^{(7)}
  \xrightarrow{\observeClause{\true}}
 %\xrightarrow{{\color{red}\observeClause{\true }}}
 \lfinal
 \xrightarrow{c1}
 \,
 \bigr)
 \\
 \CP\bigl(\StrLn(\flowl^{2,5})\bigr)
 &=
 \CP\bigl(\StrLn(\flowl^{3,6})\bigr)
 \;=\;
 \bigl(\,
 \xrightarrow{\siginit}
 \linit\xrightarrow{\observeClause{\false}}
 \lfinal
 \xrightarrow{c1}
 \,
 \bigr)
 \end{align*}
 Therefore the $K$ samples $v_{\mathrm{final}}^{m,1:K}$ in Line~\ref{line:pCFGDataSampling} are 
 \begin{equation}\label{eq:exDataSamplingCoin}
 \begin{aligned}
  v_{\mathrm{final}}^{m,1:K}
 &=
 \begin{cases}
  (\true,\true, \dotsc,\true)
 &\text{if $\vec{l^{m}}=\flowl^{2,6}$}
 \\
  (\false,\false, \dotsc,\false)
 &\text{if $\vec{l^{m}}=\flowl^{3,5}$}
 \\
  \text{(an arbitrary Boolean vector of length $K$)}
 &\text{if $\vec{l^{m}}=\flowl^{2,5}$ or $\flowl^{3,6}$}
 \end{cases}
 \end{aligned}
 \end{equation}
 In the last case, samples are arbitrary because the model is conditioned by $\false$. This choice is justified by the Bayesian view on probabilistic programs.

 Now let us turn to the top, control flow-sampling level of our algorithm (Lines~\ref{line:startCtrlFlowSampling}--\ref{line:endCtrlFlowSampling}). As one easily sees, the proposal control flow $\vec{l^{*}}$ is chosen from the following distribution.
 \begin{displaymath}
 \flowl^{2,5} \text{ with } 0.36\cdot 0.36\enspace,\quad
 \flowl^{2,6} \text{ with } 0.36\cdot 0.64\enspace,\quad
 \flowl^{3,5} \text{ with } 0.64\cdot 0.36\enspace,\quad
 \flowl^{3,6} \text{ with } 0.64\cdot 0.64\enspace.
 \end{displaymath}
 In each case, the proposal weight $w^{*}$ is $1$---note that the probabilities (such as $0.36$ and $0.64$) for probabilistic locations do not affect the weight (Line~\ref{line:proposalForProbBranching}). 

 The likelihood $\pi^{*}$ for each control flow is as follows. See Example~\ref{ex:strLn2}. 
 \begin{displaymath}
 0 \text{ for $\flowl^{2,5}$ and $\flowl^{3,6}$;}
 \quad
 1 \text{ for $\flowl^{2,6}$ and $\flowl^{3,5}$.}
 \end{displaymath}
 All these lead to the following Markov chain for MCMC sampling of control flows. The control flows $\flowl^{2,5}, \flowl^{3,6}$ can be ignored because their likelihood is $0$. 
 \begin{displaymath}
 \begin{tikzpicture}[%>=stealth',
    shorten >=1pt,auto,node distance=2cm]
   \node[state] (26)      {$\flowl^{2,6}$};
   \node[state] (35)  [right of=26]    {$\flowl^{3,5}$};
   % \node[state] (35)  [below of=25]    {$\flowl^{3,5}$};
   % \node[state] (36)  [right of=35]    {$\flowl^{3,6}$};
   \path[->]          (26)  edge [loop left] node {$0.36\cdot 0.64+0.36^{2}+0.64^{2}$} (26);
    \path[->]          (26)  edge [bend left] node {$0.64\cdot 0.36$} (35);
   \path[->]          (35)  edge [loop right] node {$0.36^{2}+0.64\cdot 0.36+0.64^{2}$} (35);
    \path[->]          (35)  edge [bend left] node {$0.36\cdot 0.64$} (26);
   % \path[->]          (26)  edge [loop right] node {a} (26);
   % \path[->]          (35)  edge [loop left] node {a} (35);
   % \path[->]          (36)  edge [loop right] node {a} (35);
  
   % \node[state]         (q1) [right of=S]  {$q_1$};
   % \node[state]         (q2) [right of=q1] {$q_2$};

   % \path[->]          (S)  edge [loop above] node {a} (S);
   % \path[->, dashed]  (S)  edge              node {a} (q1);
   % \path[->, dotted]  (q1) edge [bend left]  node {a} (S);
   % \path[->>, dotted] (q1) edge             node {b} (q2);
   % \path              (q2) edge [loop above] node {b} (q2)
   %            edge [bend left]  node {b} (q1);
 \end{tikzpicture}
 \end{displaymath}
 The uniform distribution is stationary in the Markov chain. Therefore, overall, our hierarchical algorithm alternates between sampling $K$ $\true$'s and $K$ $\false$'s (see~(\ref{eq:exDataSamplingCoin})) with the same probability.
 % Overall, our top-level algorithm yields the following Markov chain for sampling control flows. Note that the weights are not normalized. 
 % \begin{displaymath}
 % \begin{tikzpicture}[%>=stealth',
 %    shorten >=1pt,auto,node distance=2cm]
 %   \node[state] (25)      {$\flowl^{2,5}$};
 %   \node[state] (26)  [right of=25]    {$\flowl^{2,6}$};
 %   \node[state] (35)  [below of=25]    {$\flowl^{3,5}$};
 %   \node[state] (36)  [right of=35]    {$\flowl^{3,6}$};
 %   \path[->]          (25)  edge [loop left] node {$0.36^2$+} (25);
 %   \path[->]          (26)  edge [loop right] node {a} (26);
 %   \path[->]          (35)  edge [loop left] node {a} (35);
 %   \path[->]          (36)  edge [loop right] node {a} (35);
  
 %   % \node[state]         (q1) [right of=S]  {$q_1$};
 %   % \node[state]         (q2) [right of=q1] {$q_2$};

 %   % \path[->]          (S)  edge [loop above] node {a} (S);
 %   % \path[->, dashed]  (S)  edge              node {a} (q1);
 %   % \path[->, dotted]  (q1) edge [bend left]  node {a} (S);
 %   % \path[->>, dotted] (q1) edge             node {b} (q2);
 %   % \path              (q2) edge [loop above] node {b} (q2)
 %   %            edge [bend left]  node {b} (q1);
 % \end{tikzpicture}
 % \end{displaymath}
 \end{myexample}
\end{auxproof}

% \begin{figure}[tbp]
%      \begin{minipage}[t]{.48\textwidth}
%       \footnotesize 
% \begin{lstlisting}[%numbers=right,
% basicstyle={\scriptsize\ttfamily},escapechar=|,
% caption={\scriptsize\texttt{ADS(dt)}},
% label={listing:ADS}]
% a $\sim$ exponential(1/0.3); x1 = -200; y2 $\sim$ normal(-250,125^2);
% v1 = 16.7; v2init $\sim$ unif(0,22.2); v2 = v2init;
% while (x1 <= 0 && a > 0) { // car2 is yet to notice
%   x1 = x1+dt*v1; y2 = y2+dt*v2; a = a-dt; }
% while (x1 <= 0 && v2 >= 0) { // car2 noticed and brakes
%   x1 = x1+dt*v1; v2 = v2-4*dt; y2 = y2+dt*v2; }
% observe(x1 > 0 && -v2*dt <= y2 <= v2*dt && 8.3 < v2);
% return v2init;
% \end{lstlisting}
%  \end{minipage}
% \centering
% \includegraphics[bb=63 156 227 321,clip,width=.25\textwidth]{ads.pdf}
% \caption{The program $\texttt{ADS}$ (automated driving system, Program~\ref{listing:ADS}). \texttt{car2} should yield and it notices \texttt{car1} at time  $\texttt{a}$. We sample settings (the initial position and velocity of \texttt{car2}) under which a nearmiss happens with \texttt{car2} travelling with $\ge$ 30 km/h.}
% \end{figure}

\section{Implementation and
 Experiments}\label{sec:implemenationAndExperiments}
%  experimental results 2020-06-02 12:42 JST \url{https://docs.google.com/spreadsheets/d/1rFSipdWqt0U63VL-ipoROGghkfFGqR1YXwtFtP4f_Sc/edit?ts=5ed5c75f#gid=1531603126}

%  Experimental results 2020-06-03 11:52 JST
% \url{https://docs.google.com/spreadsheets/d/14tvSwCXZMmxQ7k9abJVKaz6EVv0f5hTd-UxPyCeLWXI/edit?ts=5ed70304#gid=1540321353}
% Experimental results from 2020-06-04 early morning: \url{https://ichiro-ansible.s3-ap-northeast-1.amazonaws.com/output_yml_202006/output.yml} (these can supplement the other results). The same file is in \url{ec2ExperimentsAnsible/experiment_specific/output_202006040800JST.yml}

Our
  implementation in Clojure is called $\Schism$ (for SCalable HIerarchical
 SaMpling). It  builds on top of Anglican
 1.0.0~\citep{TolpinMYW16}. It receives a $\PIMP$ program,
 translates it to a pCFG, and runs
 Algorithm~\ref{algo:ourHierarchicalSampler}. Its parameters concern the SMC sampling in Line~10, Algorithm~\ref{algo:ourHierarchicalSampler}, namely 1) number of particles in SMC (we set it to 100); 2) timeout  for each SMC run (set to 2 seconds).

% For the parameters in the decay function $\decayFn$ in~(\ref{eq:UCBVscore}), 
% we used $C_{1}=1/1000$ and $C_{2}=10$. 

 We conducted experiments to assess the performance of
 $\Schism$.
%, and also to assess the potential of Algorithm~\ref{algo:ourHierarchicalSampler}. 
We compare with Anglican~\citep{TolpinMYW16}, a state-of-the-art
 probabilistic programming system. In Anglican experiments,  we used RMH, SMC and IPMCMC as sampling algorithms (with 100 particles for the latter two); the choice follows  the developers' recommendation.\footnote{\url{probprog.github.io/anglican/inference}. We did not use the variational inference (VI) algorithm BBVB that is also offered in Anglican. This is because 1) BBVB does not support  ``flat'' distributions such as the uniform distribution (we use them in our examples), and 2) our examples have many control flows with absolutely zero likelihood, which we observed making  VI's iterative optimization struggle. } We also implemented a translator from $\PIMP$ programs to Anglican queries. The experiments were  on  m4.xlarge instances of
 Amazon Web Service (4 vCPUs, 16 GB RAM), 
 Ubuntu 18.04. The target programs are  in Program~\ref{listing:unifCd}--\ref{listing:ADS}.
The  results are summarized in Table~\ref{table:collectiveResults}; more results, including some convergence plots, are in 
\ref{appendix:suppExpResults}.

\begin{table*}[tbp]
\footnotesize
%%\vspace{-1.5em}
% -------------------------------------------------------------------
    \begin{minipage}[t]{.3\textwidth}
      \footnotesize 
\begin{lstlisting}[%numbers=right,
basicstyle={\scriptsize\ttfamily},escapechar=|,
caption={\scriptsize\texttt{unifCd(t0)}},
label={listing:unifCd}]
p $\sim$ unif(0,1);
q = 1; t = 0;
while (p <= q) {
  q = q / 2;
  t = t + 1; }
observe(t >= t0);
return p;
\end{lstlisting}
\end{minipage}
\quad
% -------------------------------------------------------------------
%%\vspace{-1.5em}
    \begin{minipage}[t]{.3\textwidth}
      \footnotesize 
\begin{lstlisting}[%numbers=right,
basicstyle={\scriptsize\ttfamily},escapechar=|,
caption={\scriptsize\texttt{unifCd2(t0)}},
label={listing:unifCd2}]
p $\sim$ unif(0,1);
q = 1; x = 0;
while (p <= q) {
  q = q / 2;
  y $\sim$ normal(1,1);
  x = x+y; t = t+1;}
observe(t >= t0);
return x;
\end{lstlisting}
\end{minipage}
\quad
% -------------------------------------------------------------------
%%\vspace{-1.5em}
    \begin{minipage}[t]{.3\textwidth}
      \footnotesize 
\begin{lstlisting}[%numbers=right,
basicstyle={\scriptsize\ttfamily},escapechar=|,
caption={\scriptsize\texttt{poisCd(p,x0)}},
label={listing:poisCd}]
m $\sim$ poisson(p);
x = 0; n = m;
while (0 < n) {
  x = x+1; n = n-1; }
observe(x >= x0);
return m;
\end{lstlisting}
\end{minipage}
\quad
% % -------------------------------------------------------------------
% %%\vspace{-1.5em}
%     \begin{minipage}[t]{.19\textwidth}
%       \footnotesize 
% \begin{lstlisting}[%numbers=right,
% basicstyle={\scriptsize\ttfamily},escapechar=|,
% caption={\scriptsize\texttt{poisCd2(p,x0)}},
% label={listing:poisCd2}]
% m $\sim$ poisson(p);
% x = 0; n = m;
% while (0 < n) {
%   y $\sim$ unif(1,1.25);
%   x = x+y; n = n-1; }
% observe(x >= x0);
% return m;
% \end{lstlisting}
% \end{minipage}
% \;
% -------------------------------------------------------------------
%%\vspace{-1.5em}
\centering
    \begin{minipage}[t]{.7\textwidth}
      \footnotesize 
\begin{lstlisting}[%numbers=right,
basicstyle={\scriptsize\ttfamily},escapechar=|,
caption={\scriptsize\texttt{poisCdS(p)}},
label={listing:poisCdS}]
$\vec{\mathtt{o}}$ = [13.676, 14.015, 12.292, 13.970, 12.755];
m $\sim$ poisson(p); $\vec{\mathtt{x}}$ = $\vec{\mathtt{0}}$; n = m; 
while (1 < n) {
  $\vec{\mathtt{y}}$ $\sim$ unif(1,1.25); $\vec{\mathtt{x}}$ = $\vec{\mathtt{x}}$+$\vec{\mathtt{y}}$; n = n-1; }
observe($\vec{\mathtt{x}}$-3 $\le$ $\vec{\mathtt{o}}$ $\le$ $\vec{\mathtt{x}}$+3); observe(normal($\vec{\mathtt{x}}$,1)($\vec{\mathtt{o}}$));
return m;
\end{lstlisting}
\end{minipage}
\\
% -------------------------------------------------------------------
%%\vspace{-1.5em}
    \begin{minipage}[t]{.3\textwidth}
      \footnotesize 
\begin{lstlisting}[%numbers=right,
basicstyle={\scriptsize\ttfamily},escapechar=|,
caption={\scriptsize\texttt{geomIt(r,x0)}},
label={listing:geomIt}]
n = 0; c $\sim$ unif(0,1);
while (c <= r) {
  n = n+1; x = x+1; 
  c $\sim$ unif(0,1); }
observe(x >= x0);
return n;
\end{lstlisting}
\end{minipage}
\quad
% -------------------------------------------------------------------
%%\vspace{-1.5em}
    \begin{minipage}[t]{.3\textwidth}
      \footnotesize 
\begin{lstlisting}[%numbers=right,
basicstyle={\scriptsize\ttfamily},escapechar=|,
caption={\scriptsize\texttt{geomIt2(r,x0)}},
label={listing:geomIt2}]
n = 0; c $\sim$ unif(0,1);
while (c <= r) {
  y $\sim$ beta(n,1);
  n = n+1; x = x+y; 
  c $\sim$ unif(0,1); }
observe(x >= x0);
return n;
\end{lstlisting}
\end{minipage}
\quad
% -------------------------------------------------------------------
%%\vspace{-1.5em}
%     \begin{minipage}[t]{.19\textwidth}
%       \footnotesize 
% \begin{lstlisting}[%numbers=right,
% basicstyle={\scriptsize\ttfamily},escapechar=|,
% caption={\scriptsize\texttt{geomIt3(r,x0)}},
% label={listing:geomIt3}]
% n = 0; c $\sim$ unif(0,1);
% q $\sim$ unif(0,1);
% while (c <= r) {
%   y $\sim$ bernoulli(q);
%   n = n+1; x = x+y; 
%   c $\sim$ unif(0,1); }
% observe(x >= x0);
% return q;
% \end{lstlisting}
% \end{minipage}
% \;
% -------------------------------------------------------------------
%%\vspace{-1.5em}
    \begin{minipage}[t]{.3\textwidth}
      \footnotesize 
\begin{lstlisting}[%numbers=right,
basicstyle={\scriptsize\ttfamily},escapechar=|,
caption={\scriptsize\texttt{mixed(p)}},
label={listing:mixed}]
x $\sim$ normal(0,1);
if x > p {
  y $\sim$ normal(10,2);
} else {
  y $\sim$ gamma(3,3); }
return y;
\end{lstlisting}
\end{minipage}
\\\centering

% -------------------------------------------------------------------
%%\vspace{-1.5em}
    \begin{minipage}[t]{.7\textwidth}
      \footnotesize 
\begin{lstlisting}[%numbers=right,
basicstyle={\scriptsize\ttfamily},escapechar=|,
caption={\scriptsize\texttt{nestLp(p)}},
label={listing:nestLp}]
o = 167.54; m $\sim$ exponential(p); l = 0; n = 0;
while (l < m) {
  k = 0;
  while (k < n) {
    y $\sim$ unif(0.9, 1.1); x = x+y; k = k+1; }
  l = l+1; n = n+l; }
observe(x-3 <= o <= x+3); observe(normal(x,1)(o));
return m;
\end{lstlisting}
\end{minipage}
\\ 
% \;
% -------------------------------------------------------------------
%%\vspace{-1.5em}
     \begin{minipage}[t]{.8\textwidth}
      \footnotesize 
\begin{lstlisting}[%numbers=right,
basicstyle={\scriptsize\ttfamily},escapechar=|,
caption={\scriptsize\texttt{ADS(dt)}},
label={listing:ADS}]
a $\sim$ exponential(1/0.3); x1 = -200; y2 $\sim$ normal(-250,125^2);
v1 = 16.7; v2init $\sim$ unif(0,22.2); v2 = v2init;
while (x1 <= 0 && a > 0) { // car2 is yet to notice
  x1 = x1+dt*v1; y2 = y2+dt*v2; a = a-dt; }
while (x1 <= 0 && v2 >= 0) { // car2 noticed and brakes
  x1 = x1+dt*v1; v2 = v2-4*dt; y2 = y2+dt*v2; }
observe(x1 > 0 && -v2*dt <= y2 <= v2*dt && 8.3 < v2);
return v2init;
\end{lstlisting}
 \end{minipage}

% -------------------------------------------------------------------
%\vspace{-0em}
%     \begin{minipage}[t]{.25\textwidth}
% \captionof{figure}{The program $\texttt{ADS}$ (automated driving system, Program~\ref{listing:ADS}). \texttt{car2} should yield and it notices \texttt{car1} at time  $\texttt{a}$. We sample settings (the initial position and velocity of \texttt{car2}) under which a near miss happens with \texttt{car2} travelling with $\ge$ 30 km/h.}
% %\caption{The program $\texttt{ADS}$ (automated driving system, Program~\ref{listing:ADS}). \texttt{car2} should yield and it notices \texttt{car1} at time  $\texttt{a}$. We sample settings (the initial position and velocity of \texttt{car2}) under which a nearmiss happens with \texttt{car2} travelling with $\ge$ 30 km/h.}
% \label{fig:ads}
% \end{minipage}
% % -------------------------------------------------------------------
% \vspace{-1.5em}
%     \begin{minipage}[t]{.19\textwidth}
%       \footnotesize 
%   % the image generated by
%   % ./smallConvGraph.py samples_unifCd_18_Schism_600sec.txt --kl_gt_name uniform-conditioned --kl_gt_param 18 --savename uniform_conditioned_18_schism_600sec_conv_body.png
% \raisebox{-\height}{ \includegraphics[width=\textwidth]{uniform_conditioned_18_schism_600sec_conv_body.png}}
% \caption{convergence}\label{fig:conv}
% \end{minipage}
% \;
% 
% =======================================================================================
\end{table*}

\begin{figure}[tbp]
% -------------------------------------------------------------------
%\vspace{-0em}
      \footnotesize \centering
\raisebox{-\height}{ \includegraphics[bb=63 156 227 321,clip,width=3cm]{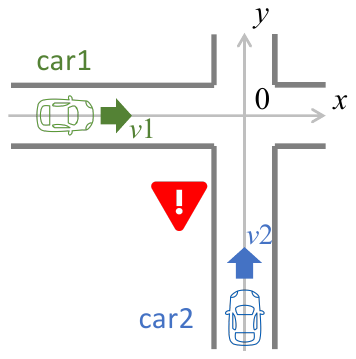}}
\caption{The program $\texttt{ADS}$ (automated driving system, Program~\ref{listing:ADS}). \texttt{car2} should yield and it notices \texttt{car1} at time  $\texttt{a}$. We sample settings (the initial position and velocity of \texttt{car2}) under which a near miss happens with \texttt{car2} travelling with $\ge$ 30 km/h.}
\label{fig:ads}
\end{figure}

\begin{table*}[tbp]
 \centering \caption{experimental results. Experiments ran for designated timeout seconds, or until 500K samples were obtained (``$\ge$500K''). For some programs,  the ground truth is known, and the KL-divergence from it is shown. When the ground truth is unknown,  the mean and standard deviation is shown. The numbers are the average of ten runs. Anglican-IPMCMC is not shown since for many programs it returned obviously wrong samples. See \ref{appendix:suppExpResults} for more details. 
% The columns ``sample'' show the numbers of samples  after two minutes of sampling. The ``gap'' is the value $|(\text{sample mean after 2 min.}) - (\text{sample mean after 1 min.})|$; a large ``gap'' suggests that the samples have not yet converged. The settings where the obtained samples are obviously wrong are designated by the red color.
}
\label{table:collectiveResults}
 \scriptsize
\begin{adjustbox}{angle=0, max width=\textwidth}
  \begin{tabular}{%@{}l@{\,}|c|c|c|c|c|c|c|
  p{2cm}||p{0.7cm}p{1.5cm}|p{0.7cm}p{1.45cm}|p{0.7cm}p{1.3cm}||p{0.7cm}p{1.3cm}|p{0.7cm}p{1.3cm}
  }
 method (timeout)
 &
 \multicolumn{2}{c|}{$\Schism$ (10 sec.)}
 &
 \multicolumn{2}{c|}{$\Schism$ (60 sec.)}
 &
 \multicolumn{2}{c||}{$\Schism$ (600 sec.)}
 &
 \multicolumn{2}{c|}{Anglican-RMH (60 sec.)}
 &
 \multicolumn{2}{c}{Anglican-SMC (60 sec.)}
 \\
 $\downarrow$ target program
  &
  samples
  &
  KL-div.
  &
  samples
  &
  KL-div.
  &
  samples
  &
  KL-div.
  &
  samples
  &
  KL-div.
  &
  samples
  &
  KL-div.
 \\\hline
% - detailed table:
%   https://docs.google.com/spreadsheets/d/1-Wgs9E1KNPKw4g8DtqwW3uZMHH9EgocI7b0LiG38jxQ/edit?ts=5ed8e9dc#gid=326400773
%   https://docs.google.com/spreadsheets/d/1vLkI7NsXdsB8iw1u08zfZn1M1gaCvLKC8QJSA4uuUvU/edit?ts=5ed90d98#gid=1791218101
%   https://docs.google.com/spreadsheets/d/11SjL5skgPqEAO2JEYD_55upgsB_tLDeWZC34Vpr9I9U/edit?ts=5ed9dd77#gid=1471224379
% - generated running
%   $ ./expr_summary_used_for_202006042200JST_2.py output_202006042200JST.yml --tex_source_output_location experiment_results_202006042200JST_2.tex
 \input{experiment_results_part}
 \end{tabular}
\end{adjustbox}
%=======================================================================================
 \scriptsize
 \centering \caption{ablation studies. 
\emph{full blkl.} blacklists both complete and incomplete  flows; \emph{leaf blkl.}  only looks at complete flows. \emph{cond.\ prop.\ only} does not blacklist; \emph{no cond.\ prop.} does not even apply condition propagation.
 Timeout is set to 600 sec. 
}
\label{table:ablation}
 \scriptsize
\begin{adjustbox}{angle=0, max width=\textwidth}
  \begin{tabular}{%@{}l@{\,}|c|c|c|c|c|c|c|
  p{1.73cm}||p{0.7cm}p{1.4cm}|p{0.7cm}p{1.3cm}|p{0.7cm}p{1.3cm}|p{0.7cm}p{1.3cm}||p{0.7cm}p{1.3cm}
  }
 method
 &
 \multicolumn{2}{c|}{$\Schism$, full blkl.}
 &
 \multicolumn{2}{c|}{$\Schism$, leaf blkl.}
 &
 \multicolumn{2}{c|}{$\Schism$, cond.\ prop.\ only}
 &
 \multicolumn{2}{c||}{$\Schism$, no cond.\ prop.}
 &
 \multicolumn{2}{c}{Anglican-SMC}
 % &
 % \multicolumn{2}{c}{Anglican-SMC (60 sec.)}
 \\
 $\downarrow$ target program
  &
  samples
  &
  KL-div.
  &
  samples
  &
  KL-div.
  &
  samples
  &
  KL-div.
  &
  samples
  &
  KL-div.
  % &
  % samples
  % &
  % KL-div.
 \\\hline
% - detailed table:
%   https://docs.google.com/spreadsheets/d/1-Wgs9E1KNPKw4g8DtqwW3uZMHH9EgocI7b0LiG38jxQ/edit?ts=5ed8e9dc#gid=326400773
%   https://docs.google.com/spreadsheets/d/1vLkI7NsXdsB8iw1u08zfZn1M1gaCvLKC8QJSA4uuUvU/edit?ts=5ed90d98#gid=1791218101
%   https://docs.google.com/spreadsheets/d/11SjL5skgPqEAO2JEYD_55upgsB_tLDeWZC34Vpr9I9U/edit?ts=5ed9dd77#gid=1471224379
% - generated running
%   $ ./expr_summary_used_for_202006042200JST_2.py output_202006042200JST.yml --tex_source_output_location experiment_results_202006042200JST_2.tex
 \input{experiment_results_ablation}
 \end{tabular}
\end{adjustbox}
\end{table*}

The programs $\texttt{mixed}$ and $\texttt{coin}$ have simple control structures (a couple of if branchings). For such programs, our control-data separation tends to have more overhead than advantage---compare the number of samples after 60 sec. 
%For  $\texttt{mixed(5)}$ we observe detrimented precision, too. The `if' branch must be rarely taken for $\texttt{p}=5$; yet random sampling (Line~\ref{line:random}, Algorithm~\ref{algo:infinitelyArmedSampling}) forces taking the branch. 
% This problem is less pronounced with more samples, as $\varepsilon_t$ becomes smaller.

The other programs feature a while loop. With them we observe benefits of our combination of  control-data separation and logical condition propagation. This is especially the case with harder instances, i.e.\ those with more restrictive conditioning (instances are sorted from easy to hard in Table~\ref{table:collectiveResults}). $\Schism$ can return samples of reasonable quality while Anglican struggles: see the results with  $\texttt{poisCd}$ and $\texttt{geomIt}$. See also $\texttt{unifCd}$, where KL-divergence differs a lot.  Condition propagation
% (Section~\ref{subsec:conditionPropag}) 
is crucial here: by logical reasoning, $\Schism$ blacklists a number of shallow control flows, quickly digging into deeper flows. Those deeper flows have tiny likelihoods and thus hard to find for Anglican.
At the same time,   $\texttt{geomIt}(0.5,5), (0.5,20)$ show that Anglican can be much faster and more precise when conditioning is not harsh, even in presence of a while loop. 

For many instances of the  programs with while loops,  IPMCMC ran very quickly but returned obviously wrong samples; see \ref{appendix:suppExpResults}. This seems to be because of the general challenge with  MCMC walks traversing different control flows~\citep[Section~4.2]{DBLP:journals/corr/abs-1809-10756}. 
% . This challenge is discussed in~\cite{HurNRS15,Kiselyov16};
% see also~\cite[Section~4.2]{DBLP:journals/corr/abs-1809-10756}. 

The programs  $\texttt{poisCdS}$ and $\texttt{nestLp}$ feature soft conditioning (by \texttt{observe(normal(x,1)(o))}), which Schism handles without problems. The programs $\texttt{nestLp}$ and  $\texttt{ADS}$ have more complicated control structures---nested  loops in  $\texttt{nestLp}$ and successive loops in $\texttt{ADS}$. For them, discovering feasible control flows is harder, resulting in $0$ samples for Schism after 60 sec. After 600 sec., however, Schism succeeds to obtain samples.

In Table~\ref{table:ablation}, we compare the performance of Schism under different policies (\emph{full blkl.} is default). We see that condition propagation and blacklisting both contribute significantly to sampling performance. In particular, for programs with complicated control structures such as $\texttt{nestLp}$ and $\texttt{ADS}$,  logical blacklisting of incomplete flows is crucial.

\begin{auxproof}
 We also note that convergence is generally slow with $\Schism$: for many programs,  KL-div.\ dropped sharply from 10 to 60 seconds, and to 600. This is no surprise: we need to sample many control flows before convergence of Algorithm~\ref{algo:infinitelyArmedSampling}, but each flow sampling involves SMC (Line~10, Algorithm~\ref{algo:ourHierarchicalSampler}) and takes time. In Figure~\ref{fig:conv} are the KL-div.\ (red), and the mean and the standard deviation (blue), plotted over samples for $\texttt{unifCd}(18)$, Schism, 600 sec. More of such plots are found in \ref{appendix:suppExpResults}. 
\end{auxproof}

Overall, we observe that $\Schism$'s combination of control-data separation and condition propagation successfully addresses the general challenge of  compatibility between sampling and control structures. Its performance is especially superior in programs with 1) while loops and 2)  restrictive conditioning (``rare events''). Programs with these features are widespread in many application domains.

One such application domain  of great practical relevance is \emph{testing of automotive systems}; see~\cite{DBLP:conf/cav/DreossiFGKRVS19} for a software science perspective of the domain. In that domain, indeed, many  models involve while loops for describing vehicle dynamics, and hazards such as near misses are often rare events---a combination that Schism is suited to. The last point is demonstrated by our example $\texttt{ADS}$ (Program~\ref{listing:ADS}, Figure~\ref{fig:ads}). Application of the current framework to a wider variety of practical problems in the domain---including \emph{scenario sampling} that involves sampling from discrete sets of options~\citep{QueirozBC19}---will require further scrutiny of the framework and its extension. This is an important direction of future work.

\begin{auxproof}
 \section{Related Work}\label{sec:relatedWork}

 Besides those which we mentioned in the earlier sections, the following are related to the current work. 

 \paragraph{Semantics of Probabilistic Programs}
 Mathematical semantics of probabilistic programs is a foundation not only of inference about their posteriors but also of program transformation. Therefore it is an active research topic. 

 For imperative probabilistic programs, the work~\cite{OlmedoGJKKM18} presents an extensive account on their state-transformer and predicate-transformer semantics. The semantics builds on the previous work on imperative programs with randomization (but without conditioning)~\cite{MorganMS96,Kozen81}. 
 % Equivalence between the two is shown. Distinction between divergence and observation failure. (This paper is REALLY easy)

 For functional probabilistic programs, devising sensible semantics is  more challenging, because the category $\mathbf{Meas}$ of measurable spaces is not Cartesian closed. Here, the recent approach by \emph{quasi Borel spaces} seems promising~\cite{HeunenKSY17,ScibiorKVSYCOMH18}.

 Forward state-transformer semantics:~\cite{Staton17,StatonYWHK16,HeunenKSY17,ScibiorKVSYCOMH18}. \cite{Staton17} and Kenta's paper with Bart are perhaps a good place to look at denotational semantics with continuous distributions. 

 \cite{Staton17}: concerned about finiteness of measures, and commutativity (Fubini). Unbounded measures are relevant in PPL since the value of a variable can keep growing. The language of~\cite{Staton17} does not include recursion, though (l.8, p.863).

 \cite{HeunenKSY17}: cool. A cartesian closed category of measurable spaces and functions, via an idea very much like \emph{realizability}. In the category, every randomness comes from that in $\mathbb{R}$. 

 \cite{ScibiorKVSYCOMH18}: no recursion but only primitive recursion (coming from inductive datatypes). They talk about \emph{trace MCMC}: how is it different from ours?

 \paragraph{Inference Methods}
 Many existing probabilistic programming frameworks employ sampling for inference (such as Anglican~\cite{TolpinMYW16},
 Venture~\cite{MansinghkaSP14}, and
 Stan~\cite{CarpenterGHLGBBGLR17}). Static, symbolic and exact inference algorithms (such as in Hakaru~\cite{ShanR17}, PSI~\cite{GehrMV16} and EfProb~\cite{ChoJ17}) tend to apply to restricted classes of programs, but their results come with correctness guarantee.

 The recent work~\cite{ScibiorKG18} presents a functional probabilistic programming framework based on Haskell. The framework aims at modular construction of various inference methods: using the monadic infrastructure of Haskell, it allows one to compositionally describe various advanced sampling methods such as particle marginal Metropolis--Hasting.

 \cite[Section~4.5]{ScibiorGG15}: Particle independent Metropolis Hastings (PIMH) is a PMCMC algorithm, where MH is run with a proposal distribution given by SMC.

 \cite{NoriHRS14}: R2. The user need to provide a suitable invariant to each while loop. Moreover, they ``prove'' correctness of their algorithm (l.-5, c.2, p.1) but their algorithm turned out to be wrong later~\cite[l.4, p.2]{HurNRS15}. 

 \cite{HurNRS15}: corrects the errors in R2~\cite{NoriHRS14}, but its experiments are only toys. Therefore it's not clear whether the current version of R2 is both correct and efficient, or not.

 \paragraph{Other Related Work}
 The work \cite{ClaretRNGB13} highlights the role of data flow analysis in inference for probabilistic programs. Technically, the work is focused on imperative programs with only Boolean variables and the Bernoulli distributions; other datatypes and distributions are accommodated by encoding and discretization. After all, their algorithm---that closely follows the denotational semantics of probabilistic programs via least fixed points---is for static, exact inference. This is different from the current work on sampling.

 \emph{Compositionality} in Bayesian inference seems to be a promising direction in probabilistic programming, exploiting the syntactic structure of probabilistic programs. For example, in the recent work~\cite{CusumanoTownerBGVM18PLDI}, a framework for \emph{incremental inference} has been introduced. We speculate that our current results about ``control flow-aware sampling'' has a big potential in compositional inference, a topic we plan to study in the future.

 One theme of the current work is the separation of discrete control flows from (often continuous) data. The theme appears also in~\cite{WuSHDR18}, although their goal is different from ours. In~\cite{WuSHDR18}, they introduce the notion of \emph{measure-theoretic Bayesian network}, in order to deal with distributions that combine discrete and continuous elements.

 \cite{DoucetJ11tutorial}: good place to learn SMC and particle filtering. Cited also e.g.\ in~\cite{ScibiorKVSYCOMH18}.

 \section{Conclusions and Future Work}

 Future work
 \begin{itemize}
 \item We shall further investigate the potential of our hierarchical sampling architecture (Figure~\ref{fig:hierarchicalSamplerArch}). This includes the use of language-based techniques other than condition propagation (such as martingale-based verification, see e.g.~\cite{ChakarovS13CAV,TakisakaOUH18}). We will also pursue goals other than the current one (namely the approximation the posterior distribution itself), such as estimating the mean, tail probabilities, and an element of the maximum likelihood.
 \item We will investigate other choices of sampling methods for the bottom level in Figure~\ref{fig:hierarchicalSamplerArch}. In particular, an efficient method that enables an even tighter integration with the top level (control-flow sampling) is desired. 
 \end{itemize}

 Sampling higher-order functional probabilistic programs via GoI. Ask Naohiko.~\cite{HoshinoMH14CSLLICS,MuroyaHH16}. 
 \begin{itemize}
 \item Because GoI is a sensible way of defining \emph{control flow} for functional programs... (Are there any other ways?)
 \item Monte Carlo tree search for adaptive MCMC sampling on the top level. A prototype algorithm is already here: see \url{monteCarloTreeSearchNotes20181116.pdf} in the same directory
 \item Besides this goal of approximating the posterior distribution itself, other goals---such as computing the mean, tail probabilities and an element of the maximum likelihood---shall be pursued, using the idea of the hierarchical architecture (Figure~\ref{fig:hierarchicalSamplerArch}). 
 \end{itemize}

 As we already discussed, we think of our proposed algorithm not as a definitive answer but as a demonstration of the potential of a general platform. We shall further investigate other techniques to be plugged in, from programming language theory and from statistics. See Section~\ref{subsec:ourSamplerAsAPlatform} for some concrete directions.

 Adaptive MCMC. One way is statistical, to adapt the branching proposal $g$ after observations of $\hat{p}(\vec{l^{1}}),\hat{p}(\vec{l^{2}}),\dotsc, \hat{p}(\vec{l^{m}})$. Another way is via static analysis, analyzing e.g.\ feasibility of control flows.

 \begin{itemize}
 \item Merging two sampling
       \begin{itemize}
	\item Possible if the data sampling also returns weight (which is not possible e.g.\ with MCMC methods)
       \end{itemize}
  \item Allow user annotation that specifies $g$
       \begin{itemize}
	\item For example, ``Execute this while loop body at least five times''
       \end{itemize}
 \end{itemize}

 Add soft conditioning?~\cite{StatonYWHK16}

 Particle Gibbs.  
 \begin{itemize}
 \item Usually:
 (one trace) $\Longleftrightarrow$ (many traces)
 \item For us:
 (one flow sample) $\Longleftrightarrow$ (many flow \& data samples)
 \end{itemize}
 See also conditional.tex by Shin-ya. 
 Seems promising but not necessarily for the current paper.

\end{auxproof}

\begin{auxproof}
 \section*{Acknowledgements}
 Thanks are due to 
 Bart Jacobs, 
 Ohad Kammar,
 Adam Scibior,
 and
 Akihisa Yamada
 for useful discussions; to Darren Wilkinson for his extensive set of blog entries that served us as a valuable information source on  sampling; and to the anonymous reviewers for previous versions of the paper for their wonderfully extensive and insightful comments and suggestions. 
 The authors are supported by ERATO
 HASUO Metamathematics for Systems Design Project (No.\ JPMJER1603)
 %(No.~\grantnum{http://dx.doi.org/10.13039/501100009024}{JPMJER1603}), 
 JST.
\end{auxproof}

\section*{Acknowledgements}
The authors are supported by ERATO HASUO  Metamathematics for Systems Design Project (No.\ JPMJER1603). I.H., K.S., and S.K. are supported by CREST CyPhAI Project (No.\ JPMJCR2012).

% \textbf{Do not} include acknowledgements in the initial version of
% the paper submitted for blind review.

% If a paper is accepted, the final camera-ready version can (and
% probably should) include acknowledgements. In this case, please
% place such acknowledgements in an unnumbered section at the
% end of the paper. Typically, this will include thanks to reviewers
% who gave useful comments, to colleagues who contributed to the ideas,
% and to funding agencies and corporate sponsors that provided financial

% % In the unusual situation where you want a paper to appear in the
% % references without citing it in the main text, use \nocite

\bibliography{myrefs}
\bibliographystyle{elsarticle-num}

\clearpage
\onecolumn
\appendix

% \section{Do \emph{not} have an appendix here}

% \textbf{\emph{Do not put content after the references.}}
% %
% Put anything that you might normally include after the references in a separate
% supplementary file.

% We recommend that you build supplementary material in a separate document.
% If you must create one PDF and cut it up, please be careful to use a tool that
% doesn't alter the margins, and that doesn't aggressively rewrite the PDF file.
% pdftk usually works fine. 

% \textbf{Please do not use Apple's preview to cut off supplementary material.} In
% previous years it has altered margins, and created headaches at the camera-ready
% stage. 
% %%%%%%%%%%%%%%%%%%%%%%%%%%%%%%%%%%%%%%%%%%%%%%%%%%%%%%%%%%%%%%%%%%%%%%%%%%%%%%%
% %%%%%%%%%%%%%%%%%%%%%%%%%%%%%%%%%%%%%%%%%%%%%%%%%%%%%%%%%%%%%%%%%%%%%%%%%%%%%%%

\section{Omitted Details}

\subsection{Derivation of Our Sampling Algorithm}\label{appendix:derivation}
%Before presenting our algorithm, 
% Here we sketch some equational reasoning that justify our control-data separation (Figure~\ref{fig:hierarchicalSamplerArch}). 

Let $\Gamma$ be a pCFG; our goal is to draw samples from the distribution 
\begin{math}
\sem{\Gamma}^{\st}
\end{math} in~(\ref{eq:unweightedSemTypeWhole}). 
 The relevant random variables are as follows: $N$  (the length of a complete control flow $\flowl$); $l_{1},\dotsc, l_{N}$ (the locations in the complete control flow $\flowl=l_{1}\dotsc l_{N}$); and $\sigma_{1},\dotsc,\sigma_{N}$ (the memory states at those locations). 
% Then the desired distribution is
% \begin{math}
%  p(\sem{\efinal}_{\sigma_{N}})
% \end{math}.
% (disregarding restriction to the return expression $\efinal$ for simplicity). 

By marginalization and the definition of conditional probability, we can transform
our target distribution $p(\sem{\efinal}_{\sigma_{N}}\in\mathrm{d}v\mid \Gamma)$ as follows.
\begin{align}
 % \begin{array}{ll}
  & p(\sem{\efinal}_{\sigma_{N}}\in\mathrm{d}v\mid \Gamma)
  \nonumber
\\
& =
 \textstyle
 \int_{\sigma_{N}}\, 
\delta_{\sem{\efinal}_{\sigma_{N}}}(\mathrm{d}v)
\cdot
p(\mathrm{d}\sigma_{N}
  \mid \Gamma)
 \nonumber
\\
& =
 \textstyle
 \int_{\sigma_{1:N}, l_{1:N}}\, 
\delta_{\sem{\efinal}_{\sigma_{N}}}(\mathrm{d}v)
\cdot
p(\mathrm{d}\sigma_{1:N}\times\mathrm{d}l_{1:N}
  \mid \Gamma)
 \qquad\text{marginalization}
 \nonumber
\\
& =
 \textstyle
 \int_{\sigma_{1:N}, l_{1:N}}\, 
\delta_{\sem{\efinal}_{\sigma_{N}}}(\mathrm{d}v)
\cdot
p(\mathrm{d}\sigma_{1:N}
  \mid l_{1:N}, \Gamma)
\cdot
p(\mathrm{d}l_{1:N}
  \mid \Gamma)
\qquad\text{conditional probabilities}
 \nonumber
\\
&=
 \textstyle\int_{l_{1:N}}\bigl(\,
 \textstyle
 \int_{\sigma_{1:N}}\,
\delta_{\sem{\efinal}_{\sigma_{N}}}(\mathrm{d}v)
\cdot
p(\mathrm{d}\sigma_{1:N}
  \mid l_{1:N}, \Gamma)
 \,\bigr)\cdot
p(\mathrm{d}l_{1:N}
  \mid \Gamma)
\nonumber
% \\
% &\qquad\qquad\qquad\qquad\qquad\qquad\qquad
%  \nonumber
% \\
% &= 
%---------------
% p(
% \sem{\efinal}_{\sigma_{N}}\in\mathrm{d}v
%  ,\, \mathrm{d}\sigma_{1:N}
%  ,\,
%   \mathrm{d}l_{1:N}
%  \mid \Gamma
%  )
%------------------
 % \cdot
 % p(\mathrm{d}\sigma_{1:N}\mid \Gamma)
 % \cdot
 % p(\mathrm{d}l_{1:N}\mid \Gamma)
 % \,
% \mathrm{d}\sigma_{1:N}\,
%  \mathrm{d}l_{1:N}
% \,
%  \mathrm{d}N
 % \qquad\text{marginalization}
 % \nonumber
% \\
% & =
%  \textstyle
%  \int_{\sigma_{1:N}, l_{1:N}}\, p(
% \sem{\efinal}_{\sigma_{N}}\in\mathrm{d}v
%   \mid
%  \sigma_{1:N}
%  ,\,
%   l_{1:N}
%  ,\,
%   \Gamma
%  )
% \cdot
% p(
%  \mathrm{d}\sigma_{1:N}
%   \mid
%   l_{1:N},
%   \Gamma
%  )
% \cdot
% p(
%   \mathrm{d}l_{1:N}
%   \mid
%   \Gamma
%  )\,
% % \mathrm{d}\sigma_{1:N}\,
%  % \mathrm{d}l_{1:N}
% % \,
% %  \mathrm{d}N
% % \qquad\text{conditional probabilities}
%  \nonumber
% \\
% &\qquad\qquad\qquad\qquad\qquad\qquad\qquad\text{conditional probabilities}
%  \nonumber
% \\
% & =
%  \textstyle
%  \int_{\sigma_{1:N}, l_{1:N}}\, p(
% \sem{\efinal}_{\sigma_{N}}\in\mathrm{d}v
%   \mid
%  \sigma_{N}
%  )
%  \cdot
% p(
%  \mathrm{d}\sigma_{1:N}
%   \mid
%   l_{1:N},
%   \Gamma
%  )
% \cdot
% p(
%  \mathrm{d}  l_{1:N}
%   \mid
%   \Gamma
%  )
% % \,
% % \mathrm{d}\sigma_{1:N}\,
% %  \mathrm{d}l_{1:N}
% % \,
% %  \mathrm{d}N
%  % \qquad\text{$\sem{\efinal}_{\sigma_{N}}$ depends only on $\sigma_{N}$}
%  \nonumber
% \\
% & \qquad\qquad\qquad\qquad\qquad\qquad\qquad\text{$\sem{\efinal}_{\sigma_{N}}$ depends only on $\sigma_{N}$}
%  \nonumber
\\
& =
 \textstyle\int_{l_{1:N}}\bigl(\,
 \textstyle
 \int_{\sigma_{1:N}}\, p(
\sem{\efinal}_{\sigma_{N}}\in\mathrm{d}v
  \mid
 \sigma_{N}
 )
 \cdot
p(
 \mathrm{d}\sigma_{1:N}
  \mid
  l_{1:N},
  \Gamma
 )
% \,
% \mathrm{d}\sigma_{1:N}
\,\bigr)\cdot
p(
  \mathrm{d}l_{1:N}
  \mid
  \Gamma
 )
% \,
%  \mathrm{d}l_{1:N}
% \,
%  \mathrm{d}N
 \nonumber
\end{align}
The last expression
 justifies the following hierarchical sampling scheme. At the top level, 
we sample a control flow $l_{1:N}$ from the distribution $p(\mathrm{d}l_{1:N}\mid\Gamma)$; at the bottom level, for each  control flow sample $l_{1:N}$, we sample data sequences $\sigma_{1:N}$, in the way
       that is \emph{conditioned} by the fixed control flow $l_{1:N}$. The latter corresponds to sampling data from the straight-line program $\StrLn(l_{1:N})$.

\subsection{Multi-Armed Sampling}\label{appendix:multiArmedSampling}
In our hierarchical sampling framework in Figure~\ref{fig:hierarchicalSamplerArch}, we formulate the top-level control flow sampling as the \emph{infinite-armed sampling} problem. As a step towards this problem, here we introduce and study the \emph{multi-armed sampling problem};  the infinite-armed sampling problem is 
its variation with infinitely many arms. 

\subsubsection{The Multi-Armed Sampling Problem} 
\label{subsubsec:finiteArmedSampling}
The setting is informally described as follows. We
have arms $\{1,2,\dotsc, K\}$, and each arm $k$ comes with its \emph{likelihood} $\truelik{k}$. Our goal is to sample (or ``pull'') arms from the set $\{1,2,\dotsc, K\}$ according to the likelihoods $\truelik{1}, \dotsc, \truelik{K}$. 

The challenge, however, is that the likelihoods $\truelik{1}, \dotsc, \truelik{K}$ are not know a priori. We assume that, if we pull an arm $k$ at time $t$, we sample a random variable which is denoted by $X_{k,t}$. 
 % observe a value represented by a random variable  $X_{k,t}$. 
We further assume that 
 $X_{k,t}$ is a random variable such that
\begin{itemize}
 \item its mean is the (unknown) true likelihood $\truelik{k}$ of the arm $k$, and
 \item it is time-homogeneous, i.e.\  $X_{k,1}, X_{k,2}, \dotsc$ are i.i.d.
\end{itemize}

Note that the problem is a ``sampling-variant'' of the classic \emph{multi-armed bandit (MAB)} problem in reinforcement learning (RL). In MAB, the goal is to \emph{optimize}---specifically, to maximize cumulative likelihood values---instead of to sample. MAB is an exemplar of the \emph{exploration-exploitation trade-off} in RL: by engaging the greedy (or ``exploitation-only'') strategy of keep pulling the empirically best-performing arm, one runs the risk of missing the actual best-performing arm, in case the latter happens to have performed empirically worse.

We formally state the problem. For the ease of theoretical analysis, we assume that the true likelihoods $\truelik{k}$, as well as the observed ones $X_{k,t}$, all take their values in the unit interval $[0,1]$. 

\begin{mydefinition}[the (finite) multi-armed sampling problem]
\label{def:finiteMultiArmedSampling}

 \begin{itemize}
  \item \textbf{Given:} a finite set $\mathcal{K}=\{1,\dotsc, K\}$ of arms. Each arm $k\in\mathcal{K}$, when pulled, returns an \emph{observed likelihood} that is given by a random variable $X_{k,t}$. We assume that $\{X_{k,t}\}_{t\in \Zpos}$ is i.i.d., and that the mean of $X_{k,t}$ is $\truelik{k}$. 
  \item \textbf{Goal:} pull one arm at each time  $t\in \Zpos$, producing a sequence of arms $k^{(1)}, k^{(2)},\dotsc$ (where $k^{(t)}\in \mathcal{K}$ for each $t\in \Zpos$), so that the vector
  \begin{align}
   &\left(\, \textstyle
      \frac{T_{1}(T)}{T}, \,\dotsc,\,\frac{T_{K}(T)}{T} \,\right),\quad
 % \nonumber
 %   \\
 %   & 
   \label{eq:visitCount}
    \text{where  $T_{k}(T)=\bigl|\{t\in [1,T]\mid k^{(t)}=k\}\bigr| $
    is the \emph{visit count}, }
  % (i.e. ``how many times $k_{i}$ is pulled'')}
  \end{align}
  converges to the vector
  \begin{math}
   \left(\,\frac{\truelik{1}}{\sum_{k}\truelik{k}},\,\dotsc,\,
    \frac{\truelik{K}}{\sum_{k}\truelik{k}}\,\right)
  \end{math}
  as $T\to \infty$. Here, the two vectors of length $K$ are understood as categorical distributions over the set $\{1, 2, \dotsc, K\}$ of arms. Convergence here precisely means the one in Theorem~\ref{thm:finitelyArmedSamplingEpsGreedyConv}. 
 \end{itemize}

\end{mydefinition}

\subsubsection{An $\varepsilon$-Greedy Algorithm}
\label{subsubsec:epsGreedyAlgoFiniteArmedSampling}
 %%%%%%%%%%%%%%%%%%%%%%%%%%%%%%%%%%%%%%%%%%%%%%%%%%%%%%%%%%%%%%%%%%%
\begin{figure*}[tbp]
% \begin{minipage}[t]{.48\textwidth}
 \null
 \begin{algorithm}[H]
 \caption{An $\varepsilon$-greedy algorithm for (finite) multi-armed sampling 
 }
 \label{algo:finitelyArmedSamplingEpsGreedy}
 \begin{algorithmic}[1]
 \Require The setting of Definition~\ref{def:finiteMultiArmedSampling} (with arms $1,\dotsc, K$), $T\in\Zpos$, and $\varepsilon_{t}= (\frac{K\log t}{t})^{\frac{1}{3}}$ for each $t\in \Zpos$
 \Ensure a sequence $k^{(1)},k^{(2)},\dotsc, k^{(T)}$ of arms from $\mathcal{K}=\{1,\dotsc, K\}$
 \State $t\gets 0$ \Comment{Initialization}
 \While{$t < T$}
  \State $t\gets t+1$
  \State $\mathsf{explore?}\gets\text{(sample from $[\mathsf{true}\mapsto \varepsilon_{t}, \mathsf{false}\mapsto 1-\varepsilon_{t}]$)}$
  \If{$\mathsf{explore?}=\mathsf{true}$}
    \State $k^{(t)}\gets \text{(sampled from the uniform distribution over $\mathcal{K}$)}$
           \label{line:explorationPick}
  \Else
    \State $k^{(t)}\gets 
           %\argmax_{k\in\mathcal{K}}  \hat{p}_{k,T_{k}(t-1)} 
           \text{(sampled according to the empirical likelihoods $\bigl[
                                  % \hat{p}_{1,T_{1}(t-1)},\dotsc,\hat{p}_{K,T_{K}(t-1)}
                                  \hat{p}_{1,t-1},\dotsc,\hat{p}_{K,t-1}
                                \bigr]$)}$
           \label{line:exploitationPick}
  \EndIf
    \State $p^{(t)}\gets \text{(sampled from $X_{k,t}$)}$
 \EndWhile
 \end{algorithmic}
 \end{algorithm}
 %%%%%%%%%%%%%%%%%%%%%%%%%%%%%%%%%%%%%%%%%%%%%%%%%%%%%%%%%%%%%%%%%%%
\end{figure*}
We propose the algorithm shown in Algorithm~\ref{algo:finitelyArmedSamplingEpsGreedy}, 
where $T_{k}(t)$ is the visit count from~(\ref{eq:visitCount}), and
% $\hat{p}_{k,T_{k}(t)}$ 		
$\hat{p}_{k,t}$ 
is the empirical mean of likelihoods observed by pulling the arm $k$, that is, 
\begin{equation}
%  \hat{p}_{k,T_{k}(t)} = \frac{1}{T_{k}(t)}\sum_{s\in[1,t]  \text{ such that } k^{(s)}=k} p^{(s)} \enspace, 
  \hat{p}_{k,t} = \frac{1}{T_{k}(t)}\sum_{s\in[1,t]  \text{ such that } k^{(s)}=k} p^{(s)} \enspace.
\end{equation}
The algorithm is exactly the same as the $\varepsilon$-greedy algorithm for MAB (for optimization), except that in Line~\ref{line:exploitationPick}, we sample according to the empirical likelihoods instead of pulling the empirically best-performing arm. The ``exploration'' rate $\varepsilon_{t}=(\frac{K\log t}{t})^{\frac{1}{3}}$ is the same as the one commonly used for MAB, too; see e.g.~\cite{Slivkins19,BubeckC12}.

\subsubsection{Convergence of the $\varepsilon$-Greedy Algorithm} \label{subsubsec:convEpsGreedyFiniteArmedSampling}
% We first note some basic facts. The following is Hoeffding's lemma. Recall that $p_{k}$ is the mean of $X_{k,t}$.
%  \begin{equation}\label{eq:hoeffding}
%  \mathbb{E}\left(e^{\lambda (X_{k,t}-p_{k})}\right) \le \psi(\lambda)\enspace,\quad
%  \mathbb{E}\left(e^{\lambda (p_{k}-X_{k,t})}\right) \le \psi(\lambda)\enspace.
%  \end{equation}
It turns out that our $\varepsilon$-Greedy algorithm (Algorithm~\ref{algo:finitelyArmedSamplingEpsGreedy}) indeed achieves the goal described in Definition~\ref{def:finiteMultiArmedSampling}, as we show in Theorem~\ref{thm:finitelyArmedSamplingEpsGreedyConv}. The convergence speed is slower, however, compared to the MAB (optimization) case. For MAB, the $\varepsilon$-Greedy algorithm achieves a regret bound $T^{2/3}\cdot O(K\log T)^{1/3}$, that is, regret $ O(K\log T/T)^{1/3}$ per trial. See~\cite{Slivkins19}.

\begin{mytheorem}\label{thm:finitelyArmedSamplingEpsGreedyConv}
 The output $k^{(1)},k^{(2)},\dotsc, k^{(T)}$ of Algorithm~\ref{algo:finitelyArmedSamplingEpsGreedy} achieves the goal of Definition~\ref{def:finiteMultiArmedSampling}. Specifically, we have
\begin{equation}\label{eq:propfinitelyArmedSamplingEpsGreedyConvStmt}
 \left|\,\frac{\mathbb{E}(T_{k}(T))}{T} - \frac{p_{k}}{\sum_{k}p_{k}}
  \,\right| =O\Bigl(K^{\frac{7}{3}}\Bigl(\frac{\log T}{T}\Bigr)^{\frac{1}{4}}\Bigr)
 \quad\text{for each $k\in \mathcal{K}$.}
\end{equation}
\end{mytheorem}
\begin{myproof}
 We follow the proof structure outlined in~\cite{Slivkins19}---it emphasizes the role of so-called \emph{clean events}---adapting it to the current sampling setting. 

 In what follows, we proceed leaving two constants $\alpha,\beta$ as parameters---$\alpha$ will appear in~(\ref{eq:defOfL}), and $\beta$ occurs in the definition of the \emph{exploration rate}
\begin{displaymath}
 \varepsilon_{t}= \left(\frac{K\log t}{t}\right)^{\beta}\enspace.
\end{displaymath}
We assume $0< \alpha,\beta< 1$. In the course of the proof, we collect constraints on $\alpha$ and $\beta$, and in the end we choose the values of $\alpha,\beta$ so that the resulting error bound is optimal. We will see that $\alpha=\frac{1}{4}$ and $\beta=\frac{1}{3}$ are optimal, yielding the exploration rate $\varepsilon_{t}$ in Algorithm~\ref{algo:finitelyArmedSamplingEpsGreedy}.

Let us consider the following events.
\begin{equation}\label{eq:E1t}
 \begin{aligned}
 &E_{1}(t) \quad\text{(where $t\in [1,T]$)}:
 \\
 &\quad
 \left[
 \parbox[c]{.7\textwidth}{
\begin{math}
   \forall k\in \mathcal{K}.\, \forall s\in [1,t].\quad
   \bigl|\,
   %\hat{p}_{k,T_{k}(s)}
\hat{p}_{k,s}
   -p_{k}\,\bigr| \le \sqrt{\frac{
 %\gamma
 \log T}{T_{k}(s)}}, 
\end{math} \quad and
\\
\begin{math}
 \forall k\in \mathcal{K}.\quad
  T_{k}(t)\ge 
% \left(\frac{(t-1)\log (t-1)}{K}\right)^{2/3}
 \frac{t^{1-\beta}\bigl(\log t\bigr)^{\beta}}{K^{1-\beta}}
 - \sqrt{t\bigl(\log t - \log(\log t)\bigr)}
\end{math}
}\right]
 % \\
 % &E_{2}:
 % && \text{(not yet...)}
\end{aligned}
\end{equation}
We claim that 
\begin{equation}\label{eq:E1Claiim}
 \mathbb{P}(E_{1}(t)) = 1 - 
%O\Bigl(K\Bigl(\frac{\log t}{t}\Bigr)^{1/3}\Bigr)
O\Bigl(K\Bigl(\frac{\log t}{t}\Bigr)^{2}\Bigr)
 \enspace.
\end{equation}
To see this, first observe that 
\begin{equation}\label{eq:2019-07-07-0218}
 \mathbb{P}\left(
  \bigl|\,
%  \hat{p}_{k,T_{k}(s)}
\hat{p}_{k,s}
  -p_{k}\,\bigr| \le \sqrt{\frac{
 %\gamma
  \log T}{T_{k}(s)}}
 \right) \ge 1 - \frac{2}{T^{2
  %\gamma
}}
\end{equation}
holds by Hoeffding's inequality for each $k$ and $s$. 
\begin{auxproof}
\begin{myproposition}[Hoeffding's inequality]\label{prop:Hoeffding}
 Let $X_{1},\dotsc, X_{n}$ be independence random variables whose values are in the interval $[0,1]$. Let $\hat{X}=\frac{1}{n}(X_{1}+\cdots + X_{n})$ denote their empirical mean. Then, for each nonnegative $t$, the following holds.
\begin{displaymath}
 \mathbb{P}\bigl(\,\bigl|\,\hat{X} - \mathbb{E}(\hat{X})\,\bigr|\ge t\,\bigr)
 \;\le\; 2e^{-2nt^{2}}
\end{displaymath}
\end{myproposition}
To derive~(\ref{eq:E1Claiim}), we take $T_{k}(s)$ and $\sqrt{\frac{2\log T}{T_{k}(s)}}$ as $n$ and $t$ in Proposition~\ref{prop:Hoeffding}, respectively. The resulting tail probability is, indeed,
\begin{align*}
 2e^{-2nt^{2}}
 &=
 2\exp\bigl(-2\cdot T_{k}(s)\cdot \frac{2\log T}{T_{k}(s)}  \bigr) = 
 2\exp(-4 \log T) = \frac{2}{T^{4}}\enspace.
\end{align*}
\end{auxproof}
For the second condition in $E_{1}(t)$ in~(\ref{eq:E1t}), we 
argue as follows. According to Algorithm~\ref{algo:finitelyArmedSamplingEpsGreedy}, the visit count $T_{k}(t)$ is the sum of the number of the exploration picks (Line~\ref{line:explorationPick}) and that of the usual picks (Line~\ref{line:exploitationPick}). By counting only the exploration picks, and also noting that $\varepsilon_{t} = (\frac{K\log t}{t})^{\beta}$  is decreasing with respect to $t$, we observe that 
\begin{equation}\label{eq:2019-07-07-0156}
 T_{k}(t) \ge \mathsf{RB}(\varepsilon_{t} K^{-1}, t)
 = 
  \mathsf{RB}\left(\frac{(\log t)^{\beta}}{K^{1-\beta}t^{\beta}},\, t\right),
\end{equation}
where $\mathsf{RB}(\alpha,s)$ is the ``repeated Bernoulli'' random variable for the number of heads when a coin with bias $\alpha$ is tossed $s$ times. This observation is used in the following reasoning.
\begin{equation}\label{eq:2019-07-07-0217}
 \begin{aligned}
& \mathbb{P}\left[
  T_{k}(t)\ge 
\frac{t^{1-\beta}(\log t)^{\beta}}{K^{1-\beta}}
% \left(\frac{(t-1)\log (t-1)}{K}\right)^{2/3}
 - \sqrt{t\bigl(\log t - \log(\log t)\bigr)}
 \right]
\\
& \ge \mathbb{P}\left[
 \mathsf{RB}\left(\frac{(\log t)^{\beta}}{K^{1-\beta}t^{\beta}},\, t\right)\ge 
\frac{t^{1-\beta}(\log t)^{\beta}}{K^{1-\beta}}
% \left(\frac{(t-1)\log (t-1)}{K}\right)^{2/3}
 - \sqrt{t\bigl(\log t - \log(\log t)\bigr)}
 \right]
%\qquad\text{by~(\ref{eq:2019-07-07-0156})}
\\
&\ge
1-\exp\left(-2\left(\sqrt{\frac{\log t - \log(\log t)}{t}}\right)^{2}t\right)
\qquad\text{by Hoeffding's inequality}
\\
&=
1-\left(\frac{\log t}{t}
\right)^{2}
\enspace.
\end{aligned}
\end{equation}
Combining~(\ref{eq:2019-07-07-0218}) and~(\ref{eq:2019-07-07-0217}) we obtain
\begin{equation}\label{eq:2019-07-27-1845}
 \mathbb{P}(E_{1}(t))\ge 1 - KT\frac{2}{T^{2
  %\gamma
}}- K\left(\frac{\log t}{t}\right)^{2}
\end{equation}
which proves the claim~(\ref{eq:E1Claiim}). 
Note that, in deriving~(\ref{eq:2019-07-27-1845}), we used the following basic 
 principle:
\begin{displaymath}
  \mathbb{P}(\bigwedge_{i=1}^{n}A_{i}) 
\;=\;
   1 - \mathbb{P}(\bigvee_{i=1}^{n}\overline{A_{i}}) 
 \;\ge \;
 1 -
 \sum_{i=1}^{n}\mathbb{P}(A_{i})\enspace.
\end{displaymath}

Towards the statement~(\ref{eq:propfinitelyArmedSamplingEpsGreedyConvStmt}) of the proposition, 
we introduce a constant
\begin{equation}\label{eq:defOfL}
  L=\lceil T^{1-\alpha}(\log T)^{\alpha}\rceil\enspace,
\end{equation}
where $\alpha$ is a parameter that will later be chosen by optimization (as we discussed at the beginning of the proof). 
The intuition of $L$ is as follows: the first $L$ samples $k^{(1)},\dotsc, k^{(L)}$ (out of $T$ samples in total) are \emph{transient} ones, and they are inessential in the convergence guarantee in Theorem~\ref{thm:finitelyArmedSamplingEpsGreedyConv}. 

We focus on suitable clean events (namely $E_{1}(L)\land E_{1}(T)$), and estimate the deviation from the true  likelihood  
$\frac{p_{k}}{\sum_{k\in\mathcal{K}} p_{k}}$.
\begin{align*}
 &\mathbb{E}\left(\frac{T_{k}(T)}{T} 
- 
\frac{p_{k}}{\sum_{k\in\mathcal{K}} p_{k}}
\;\bigg|\; E_{1}(L)\land E_{1}(T)\right)
\\
 &=
 \mathbb{E}\left(
 \frac{T_{k}(L)}{T}+ 
 \frac{1}{T}
\sum_{t=L+1}^{T}
\frac{\varepsilon_{t}}{K}
+
 \frac{1}{T}
\sum_{t=L+1}^{T}(1-\varepsilon_{t})\frac{
 %\hat{p}_{k,T_{k}(t)}
\hat{p}_{k,t}
}{\sum_{k\in\mathcal{K}} 
%\hat{p}_{k,T_{k}(t)}
\hat{p}_{k,t}
}
- 
\frac{p_{k}}{\sum_{k\in\mathcal{K}} p_{k}}
\;\bigg|\; E_{1}(L)\land E_{1}(T)
\right)
% - T\frac{p_{k}}{\sum_{k\in\mathcal{K}} p_{k}}
\\
&
\qquad
\parbox{.8\textwidth}{where the first three terms stand for  samples from $t\le L$, exploration picks after $L$, and exploitation picks after $L$, respectively}
\\
&\le 
 \mathbb{E}\left(
 \frac{L}{T}+ 
 \frac{\varepsilon_{L}(T-L)}{KT}
+
 \frac{1}{T}
\sum_{t=L+1}^{T}
    \frac{p_{k}+\sqrt{\frac{
 %\gamma
 \log T}{T_{k}(t)}}
        }{
      \sum_{k\in\mathcal{K}} \left(p_{k} - \sqrt{\frac{
 %\gamma
 \log T}{T_{k}(t)}}\right)}
- 
\frac{p_{k}}{\sum_{k\in\mathcal{K}} p_{k}}
\;\Bigg|\; E_{1}(L)\land E_{1}(T)
\right)
% - T\frac{p_{k}}{\sum_{k\in\mathcal{K}} p_{k}}
\\&\qquad\text{by $T_{k}(L)\le L$, $
%\varepsilon_T\le
 \varepsilon_{t}\le \varepsilon_{L}$ for $t\in[L+1,T]$, and $E_{1}(T)$}
\\
&\le 
 \mathbb{E}\left(
 \frac{L}{T}+ 
 \frac{\varepsilon_{L}T}{KT}
+
\frac{1}{T}
\sum_{t=L+1}^{T}
    \frac{p_{k}+\sqrt{\frac{
     %\gamma
 \log T}{T_{k}(L)}}
        }{
      \sum_{k\in\mathcal{K}} \left(p_{k} - \sqrt{\frac{
  %\gamma
 \log T}{T_{k}(L)}}\right)}
- 
\frac{p_{k}}{\sum_{k\in\mathcal{K}} p_{k}}
\;\Bigg|\; E_{1}(L)\land E_{1}(T)
\right)
% - T\frac{p_{k}}{\sum_{k\in\mathcal{K}} p_{k}}
\\&\qquad\text{by $T_{k}(L)\le T_{k}(t)$ for $t\in [L+1,T]$}
\\
&\le 
 \mathbb{E}\left(
 \frac{L}{T}+ 
 \frac{\varepsilon_{L}}{K}
+
%\sum_{t=L+1}^{T}
\left(    \frac{p_{k}+\sqrt{\frac{
  %\gamma
 \log T}{T_{k}(L)}}
        }{
      \sum_{k\in\mathcal{K}} \left(p_{k} - \sqrt{\frac{
   %\gamma
 \log T}{T_{k}(L)}}\right)}
-
\frac{p_{k}}{\sum_{k\in\mathcal{K}} p_{k}}
\right)
\;\Bigg|\; E_{1}(L)\land E_{1}(T)
\right)
\numberthis\label{eq:threeTerms}
% \\
% &\color{lightgray}
% \le 
%  \mathbb{E}\left(
%  L+ \sum_{t=L+1}^{T}
%     \frac{p_{k}+\sqrt{\frac{
%     %\gamma
%    \log T}{T_{k}(L)}}
%         }{
%       \sum_{k\in\mathcal{K}} \left(p_{k} - \sqrt{\frac{
%       %\gamma
%  \log T}{T_{k}(L)}}\right)}
% \;\Bigg|\; E_{1}(L)\land E_{1}(T)
% \right)
% \quad\text{by $T_{k}(L)\le T_{k}(t)$ for $t\in [L+1,T]$}
% \\
% &\color{lightgray}\le 
%  \mathbb{E}\left(
%  L+ \sum_{t=L+1}^{T-1}
%     \frac{p_{k}+\sqrt{\frac{2\log T}{
%  \frac{L^{2/3}(\log L)^{1/3}}{K^{2/3}}
%  - \sqrt{\frac{\log L - \log(\log L)}{6L}}
% }}
%         }{
%       \sum_{k\in\mathcal{K}} \left(p_{k} - \sqrt{\frac{2\log T}{
%  \frac{L^{2/3}(\log L)^{1/3}}{K^{2/3}}
%  - \sqrt{\frac{\log L - \log(\log L)}{6L}}
% }}\right)}
% \;\Bigg|\; E_{1}(L)\land E_{1}(T)
% \right)
% \quad\text{by $E_{1}(L)$.}
% \numberthis\label{eq:2019-07-27-1856}
% \\
\end{align*}

%In~(\ref{eq:2019-07-27-1856}), 
We shall bound,  from above, the three terms in the last expression.
\begin{align*}
 \frac{L}{T} &= O\left(\left(\frac{\log T}{T}\right)^{\alpha}\right)
 \numberthis\label{eq:firstTerm}
% \\
%  \quad  \tag{\color{red}first term!!}
\\
 \frac{\varepsilon_{L}}{K}
 &=
 \frac{1}{K} \left(\frac{K\log L}{L}\right)^{\beta} 
\\
 &=
 O\left(
\frac{1}{K^{1-\beta}}
 \frac{\bigl((1-\alpha)\log T + \alpha \log (\log T)\bigr)^{\beta}}{T^{\beta(1-\alpha)}(\log T)^{\alpha\beta}}
\right)
\\
 &= 
 O\left(
 \frac{1}{K^{1-\beta}}
\left(\frac{\log T}{T}\right)^{\beta(1-\alpha)}
\right)
 \numberthis\label{eq:secondTerm}
% \\
%  \tag{\color{red}second term!!}
%  \\
% T
% \left(    \frac{p_{k}+\sqrt{\frac{\gamma\log T}{T_{k}(L)}}
%         }{
%       \sum_{k\in\mathcal{K}} \left(p_{k} - \sqrt{\frac{\gamma\log T}{T_{k}(L)}}\right)}
% -
% \frac{p_{k}}{\sum_{k\in\mathcal{K}} p_{k}}
% \right)
% \;\Bigg|\; E_{1}(L)\land E_{1}(T)
% \right)
\end{align*}
On the third term in~(\ref{eq:threeTerms}),
% \begin{math}
%  T
%  \left(    \frac{p_{k}+\sqrt{\frac{\gamma\log T}{T_{k}(L)}}
%         }{
%       \sum_{k\in\mathcal{K}} \left(p_{k} - \sqrt{\frac{\gamma\log T}{T_{k}(L)}}\right)}
%  -
%  \frac{p_{k}}{\sum_{k\in\mathcal{K}} p_{k}}
%  \right)
% \end{math},
we proceed as follows. 
\begin{align*}
 T_{k}(L)&\ge 
 \frac{L^{1-\beta}\bigl(\log L\bigr)^{\beta}}{K^{1-\beta}}
 - \sqrt{L\bigl(\log L - \log(\log L)\bigr)}
\quad\text{by $E_{1}(L)$,}
\\
 \frac{L^{1-\beta}(\log L)^{\beta}}{K^{1-\beta}}
 &= \Theta\Bigl(T^{(1-\alpha)(1-\beta)}(\log T)^{\alpha(1-\beta)}\bigl(\log(T^{1-\alpha}(\log T)^{\alpha})\bigr)^{\beta}
   \frac{1}{K^{1-\beta}}
 \Bigr)
 \\
 &= \Theta\Bigl(T^{(1-\alpha)(1-\beta)}(\log T)^{\alpha(1-\beta)}\bigl(
 %\log(T^{2/3}(\log T)^{1/3})
 (1-\alpha)\log T + \alpha\log(\log T)
\bigr)^{\beta}
  \frac{1}{K^{1-\beta}}
\Bigr)
%  \\
%  &= \Theta\Bigl(T^{4/9}(\log T)^{2/9}\bigl(
%  %\log(T^{2/3}(\log T)^{1/3})
%  \frac{2}{3}\log T
% \bigr)^{1/3}\Bigr)
 \\
 &= \Theta\bigl(T^{(1-\alpha)(1-\beta)}(\log T)^{\alpha(1-\beta)+\beta}
    \frac{1}{K^{1-\beta}}
   \bigr)
 \\
 &= \Theta\bigl(T^{(1-\alpha)(1-\beta)}(\log T)^{1-(1-\alpha)(1-\beta)}
    \frac{1}{K^{1-\beta}}
   \bigr)
 \\
 &
 \qquad\text{by $\alpha(1-\beta)+\beta =1-(1-\alpha)(1-\beta)$,}
\numberthis\label{eq:fracLlogLK}
\end{align*}
\begin{align*}
&\sqrt{L\bigl(\log L - \log(\log L)\bigr)}
\\
&=
\Theta\left(\left(
T^{1-\alpha}(\log T)^{\alpha}
\bigl((1-\alpha)\log T + \alpha \log(\log T) - \log ((1-\alpha)\log T + \alpha \log(\log T))\bigr)
\right)^{\frac{1}{2}}\right)
\\
&=
\Theta\left(
T^{\frac{1-\alpha}{2}}(\log T)^{\frac{1+\alpha}{2}}
\right)\enspace.
\numberthis\label{eq:sqrtLogLLogLogL}
\end{align*}
Let us now impose a condition
\begin{equation}\label{eq:alphaBetaCond}
 (1-\alpha)(1-\beta) > \frac{1-\alpha}{2}\enspace,
\end{equation}
which will be taken into account when we choose the values of $\alpha,\beta$ in the end. This assumption implies that~(\ref{eq:fracLlogLK}) dominates~(\ref{eq:sqrtLogLLogLogL}), and hence that
\begin{align}\label{eq:boundingTkL}
 T_{k}(L) \;\ge\; \Theta\left(T^{(1-\alpha)(1-\beta)}(\log T)^{1-(1-\alpha)(1-\beta)}\frac{1}{K^{1-\beta}}\right)\enspace.
\end{align}

We continue the estimation of the third term in~(\ref{eq:threeTerms}). Let $\delta$ denote $\sqrt{\frac{
%\gamma
\log T}{T_{k}(L)}}$ occurring therein. It follows from~(\ref{eq:boundingTkL}) that
\begin{math}
 \delta\le \Theta \Bigl(
%\gamma^{\frac{1}{2}}
K^{\frac{1-\beta}{2}}
 \left(\frac{\log T}{T}\right)^{\frac{(1-\alpha)(1-\beta)}{2}}
\Bigr)
\end{math};
in particular, $\delta$ tends to $0$ as $T\to \infty$ since $\alpha,\beta\in(0,1)$.
\begin{align*}
 \left(    \frac{p_{k}+\delta
        }{
      \sum_{k\in\mathcal{K}} \left(p_{k} - \delta\right)}
 -
 \frac{p_{k}}{\sum_{k\in\mathcal{K}} p_{k}}
 \right)
&=  
 \left(    \frac{p_{k}+\delta
        }{
      (\sum_{k\in\mathcal{K}} p_{k}) - K\delta}
 -
 \frac{p_{k}}{\sum_{k\in\mathcal{K}} p_{k}}
 \right)
\\
&
= O\left(
  (p_{k}+\delta)
  \left(
  \frac{1}{\sum_{k\in\mathcal{K}} p_{k}}
  +
  \frac{2K\delta}{\left(\sum_{k\in\mathcal{K}} p_{k}\right)^{2}}
\right)
 - 
\frac{p_{k}}{\sum_{k\in\mathcal{K}} p_{k}}
\right)
\numberthis\label{eq:2019-07-27-1959}
\\
&
= O\left(
  \frac{\delta}{\sum_{k\in\mathcal{K}} p_{k}}
  +
  \frac{2K\delta(p_{k}+\delta)}{\left(\sum_{k\in\mathcal{K}} p_{k}\right)^{2}}
\right)
\qquad
\\
&
= O\left(
 K\delta
\right)
\\
&
=
O\Bigl(
%\gamma^{\frac{1}{2}} 
K^{\frac{3-\beta}{2}}
\left(\frac{\log T}{T}\right)^{\frac{(1-\alpha)(1-\beta)}{2}}
\Bigr)
%  O\left(
%  K\frac{(\log T)^{4/9}}{T^{4/9}}
% \right)
\enspace.
\numberthis\label{eq:thirdTerm}
% \\
% &\quad \tag{\color{red}third term!!}
\end{align*}
In~(\ref{eq:2019-07-27-1959}) we used the following general fact. Let $a$ a constant, and  $\varphi$ be a function of $T$ that takes nonnegative values and tends to $0$ as $T\to \infty$. Then $\frac{1}{a-\varphi}=O(\frac{1}{a}+\frac{2\varphi}{a^{2}})$. 
\begin{auxproof}
\begin{align*}
  \frac{(a-\varphi)(a+2\varphi)}{a^{2}}-1
 & =
  \frac{a^{2}+a\varphi -2\varphi^{2}}{a^{2}}-1
 \\& =
  \frac{\varphi (a - 2\varphi)}{a^{2}}
\end{align*}
and the last value is nonnegative for sufficiently large $T$. 
\end{auxproof}

Combining~(\ref{eq:threeTerms}) and~(\ref{eq:firstTerm},\ref{eq:secondTerm},\ref{eq:thirdTerm}), we obtain, for each $k\in\mathcal{K}= \{1,2,\dotsc, K\}$,
\begin{align*}
& \mathbb{E}\left(\frac{T_{k}(T)}{T} 
- 
\frac{p_{k}}{\sum_{k\in\mathcal{K}} p_{k}}
\;\bigg|\; E_{1}(L)\land E_{1}(T)\right)
\\
& =
 O\left(
 \left(\frac{\log T}{T}\right)^{\alpha}
 +
 \frac{1}{K^{1-\beta}} \left(\frac{\log T}{T}\right)^{\beta(1-\alpha)}
 + 
% \gamma^{\frac{1}{2}} 
K^{\frac{3-\beta}{2}}
\left(\frac{\log T}{T}\right)^{\frac{(1-\alpha)(1-\beta)}{2}}
\right)\enspace.
\end{align*}
Since 
$\sum_{k\in \mathcal{K}}\frac{T_{k}(T)}{T} =
\sum_{k\in \mathcal{K}}\frac{p_{k}}{\sum_{k\in\mathcal{K}} p_{k}}
=1
$, we also obtain a lower bound of the difference, namely
\begin{align*}
& \mathbb{E}\left(
\frac{p_{k}}{\sum_{k\in\mathcal{K}} p_{k}}
- 
\frac{T_{k}(T)}{T} 
\;\bigg|\; E_{1}(L)\land E_{1}(T)\right)
\\
&=
\sum_{k'\in \mathcal{K}\setminus\{k\}}
 \mathbb{E}\left(\frac{T_{k'}(T)}{T} 
- 
\frac{p_{k'}}{\sum_{k\in\mathcal{K}} p_{k}}
\;\bigg|\; E_{1}(L)\land E_{1}(T)\right)
\\
& =
 O\left(
 K\left(\frac{\log T}{T}\right)^{\alpha}
 +
 K^{\beta} \left(\frac{\log T}{T}\right)^{\beta(1-\alpha)}
 + 
% \gamma^{\frac{1}{2}} 
K^{\frac{5-\beta}{2}}
\left(\frac{\log T}{T}\right)^{\frac{(1-\alpha)(1-\beta)}{2}}
\right)\enspace.
\end{align*}
Combining the above two, we obtain
\begin{equation}\label{eq:boundForAbsValue}
\begin{aligned}
&   \mathbb{E}\left(\;\bigg|\frac{T_{k}(T)}{T} 
 - 
 \frac{p_{k}}{\sum_{k\in\mathcal{K}} p_{k}}
 \bigg|
 \;\bigg|\; E_{1}(L)\land E_{1}(T)\right)
 \\
&=
 O\left(
 K\left(\frac{\log T}{T}\right)^{\alpha}
 +
 K^{\beta} \left(\frac{\log T}{T}\right)^{\beta(1-\alpha)}
 + 
 % \gamma^{\frac{1}{2}} 
 K^{\frac{5-\beta}{2}}
 \left(\frac{\log T}{T}\right)^{\frac{(1-\alpha)(1-\beta)}{2}}
 \right)\enspace.
\end{aligned}
\end{equation}

We are ready to show the statement of the proposition.
\begin{align*}
& 
\left|
\frac{\mathbb{E}(T_{k}(T))}{T} - \frac{p_{k}}{\sum_{k}p_{k}}
\right|
\\
&=
\mathbb{E}
\left(
\left|\,
\frac{T_{k}(T)}{T} - \frac{p_{k}}{\sum_{k}p_{k}}
\,\right|
\,\Bigg|\,
E_{1}(L)\land E_{1}(T)
\right)
\cdot 
\mathbb{P}\bigl(\,E_{1}(L)\land E_{1}(T)\,\bigr)
\\
&\quad
+
\mathbb{E}
\left(
\left|\,
\frac{T_{k}(T)}{T} - \frac{p_{k}}{\sum_{k}p_{k}}
\,\right|
\,\Bigg|\,
\overline{E_{1}(L)}\lor \overline{E_{1}(T)}
\right)
\cdot 
\mathbb{P}\bigl(\,\overline{E_{1}(L)}\lor \overline{E_{1}(T)}\,\bigr)
%--------------------------------------------
\\
&\le
\mathbb{E}
\left(
\left|\,
\frac{T_{k}(T)}{T} - \frac{p_{k}}{\sum_{k}p_{k}}
\,\right|
\,\Bigg|\,
E_{1}(L)\land E_{1}(T)
\right)
+
\mathbb{P}\bigl(\,\overline{E_{1}(L)}\lor \overline{E_{1}(T)}\,\bigr)
\\
&=
O\left(
 K\left(\frac{\log T}{T}\right)^{\alpha}
 +
 K^{\beta} \left(\frac{\log T}{T}\right)^{\beta(1-\alpha)}
 + 
% \gamma^{\frac{1}{2}} 
K^{\frac{5-\beta}{2}}
\left(\frac{\log T}{T}\right)^{\frac{(1-\alpha)(1-\beta)}{2}}
\right)
\\
&\qquad\qquad
+O\left(
K\Bigl(\frac{\log L}{L}\Bigr)^{2}
+
K\Bigl(\frac{\log T}{T}\Bigr)^{2}
\right)
\qquad
\text{by~(\ref{eq:E1Claiim},\ref{eq:boundForAbsValue})
%~(\ref{eq:threeTerms},\ref{eq:firstTerm},\ref{eq:secondTerm},\ref{eq:thirdTerm})
}
\\
&=
O\left(
 K\left(\frac{\log T}{T}\right)^{\alpha}
 +
 K^{\beta} \left(\frac{\log T}{T}\right)^{\beta(1-\alpha)}
 + 
% \gamma^{\frac{1}{2}} 
K^{\frac{5-\beta}{2}}
\left(\frac{\log T}{T}\right)^{\frac{(1-\alpha)(1-\beta)}{2}}\right.
\\
&\qquad\qquad
\left.
+
K \left(\frac{\log T}{T}\right)^{2(1-\alpha)}
+
K\Bigl(\frac{\log T}{T}\Bigr)^{2}
\right)
\numberthis\label{eq:boundInTheEnd}
\end{align*}

We now choose the parameters $\alpha,\beta\in(0,1)$ so that~(\ref{eq:boundInTheEnd}) is optimal, that is, so that the minimum of the powers of the five terms (namely $\alpha$, $\beta(1-\alpha)$, $\frac{(1-\alpha)(1-\beta)}{2}$, $2(1-\alpha)$, $2$) is maximal. The choice turns out to be
$\alpha=\frac{1}{4}, \beta=\frac{1}{3}$,\footnote{
Obtained by a numeric solver in MATLAB. 
% The command run is
% fmincon(@(x) -(min([x(1), x(2)*(1-x(1)), (1-x(1))*(1-x(2))/2, 2*(1-x(1)), 2])), [0.5, 0.5], [diag([-1, -1]); diag([1, 1])], [0; 0; 1; 1])+
} which yields
\begin{math}
  \alpha=\beta(1-\alpha)=\frac{(1-\alpha)(1-\beta)}{2} = 1/4
\end{math}, and $2(1-\alpha)=3/2$. The choice of $\alpha,\beta$ satisfies the additional condition~(\ref{eq:alphaBetaCond}) that we introduced in the course of the proof, too. 

The bound~(\ref{eq:boundInTheEnd}) under the above choice of $\alpha,\beta$ yields the desired bound. This concludes the proof. \myqed
% \begin{math}
%  \alpha=\beta(1-\alpha)=\frac{(1-\alpha)(1-\beta)}{2}
% \end{math}
% yields $\alpha=1/4, \beta=1/3$. 
\end{myproof}

\subsection{Comparison with the Flow Sampling Strategy in~\cite{ZhouYTR20}}\label{appendix:ZhouEtAlFlowSampling}
 In~\cite{ZhouYTR20}, traversing different ``sub-programs'' (that correspond to control flows in this paper) is thought of as a  problem of \emph{resource allocation}. Their solution to the problem is a UCB-based algorithm  adapted from~\cite{RainforthZLTWYvdM18}.  Their algorithm do not aim to sample flows in proportion to their likelihoods; instead, it allocates more resources to those flows whose flow likelihood samples have a larger variance. Doing so follows ideas in \emph{stratified sampling} and  accelerates convergence of the \emph{mean value} of  flow likelihoods samples. See~\cite[Appendix~F]{ZhouYTR20}.

One can argue as follows:  in  probabilistic program inference, our interest is in return values, instead of in flow likelihoods. This suggests a modification of the algorithm in~\cite{ZhouYTR20}, so that it samples more often those control flows that have a larger variance of \emph{return value samples}. 

Even in the last modification, the stratified sampling-style variance-guided resource allocation strategy aims at fast convergence of the \emph{mean value}, instead of the return value distribution  itself. This does not suit such cases in which our interest is beyond the mean value. For example, in automotive system safety, international standards such as ISO 26262 require rare hazards to be identified and addressed. See the example \texttt{ADS} in Program~\ref{listing:ADS} \& Figure~\ref{fig:ads}.

With the arguments in the above in mind, proportional flow sampling emerges as a viable alternative. Not knowing 1) the data distributions for different control flows or 2) the user's statistical interest (mean, variance, other statistics, or the return value distribution itself), choosing control flows in proportion to their likelihoods seems to be the best one can do. This justifies our study of IAS sampling of control flows; further comparison is future work.

\subsection{Justification of Flow Likelihood Estimation}\label{appendix:justificationOfAuxSampling}
The equality~(\ref{eq:justifyAuxSampling}) in Section~\ref{subsec:mainAlgorithm} is shown as follows. Here we use the definition of conditional probability; we also use the principle of conditional independence (i.e.\ the Markovian property of pCFGs) to derive equalities such as $p(\sigma_{3}\mid \sigma_{1:2},l_{1:3},\Gamma)=p(\sigma_{3}\mid \sigma_{2},l_{2:3},\Gamma)$. We also note that
 the set of complete control flows $l_{1:N}$ is countable and thus it makes sense to speak of the probability $  p(l_{1:N}\mid\Gamma)$. 
\begin{equation}\label{eq:justifyAuxSamplingAppendix}
 \begin{array}{ll}
  p(l_{1:N}\mid\Gamma)
 &=
  \int_{\sigma_{1:N}}\,p(\mathrm{d}\sigma_{1:N},l_{1:N}\mid\Gamma)
 % \,\mathrm{d}\sigma_{1:N}
 \\
 &=
  \int_{\sigma_{1:N}}\,p(\mathrm{d}\sigma_{1},l_{1}\mid\Gamma) \,p(\mathrm{d}\sigma_{2:N},l_{2:N}\mid \sigma_{1},l_{1},\Gamma)
% \,\mathrm{d}\sigma_{1:N}
 \\
 &=
  \int_{\sigma_{1:N}}\,p(\mathrm{d}\sigma_{1},l_{1}\mid\Gamma) 
  \,p(l_{2}\mid \sigma_{1},l_{1},\Gamma)
  \,p(\mathrm{d}\sigma_{2}\mid \sigma_{1},l_{1:2},\Gamma)
  % \qquad
  % \\\hfill
  \,
   p(\mathrm{d}\sigma_{3:N},l_{3:N}\mid \sigma_{1:2},l_{1:2},\Gamma)
% \,\mathrm{d}\sigma_{1:N}
 \\
 &=\cdots
 \\
 &=
  \int_{\sigma_{1:N}}\,p(\mathrm{d}\sigma_{1},l_{1}\mid\Gamma) 
   \cdot
  \left(
  \prod_{k=2}^{N} p(l_{k}\mid \sigma_{k-1},l_{k-1},\Gamma)
  \right)
 % \\\hfill
  \cdot
  \left(
  \prod_{k=2}^{N} p(\mathrm{d}\sigma_{k}\mid \sigma_{k-1},l_{k-1:k},\Gamma)
  \right)
% \,\mathrm{d}\sigma_{1:N}
 \end{array}
\end{equation}

\section{Supplementary Experimental Results}\label{appendix:suppExpResults}
The following experiment data is supplementary to Section~\ref{sec:implemenationAndExperiments}. The new target programs (with parameters) are in Program~\ref{listing:ProbProgramCoinEmulationParam}--\ref{listing:obsInLoopParam}.

\begin{figure*}[tbp]
\footnotesize\centering
    \begin{minipage}[t]{.3\textwidth}
      \footnotesize 
% -------------------------------------------------------------------
%\vspace{-1.5em}
\begin{lstlisting}[%numbers=right,
basicstyle={\scriptsize\ttfamily},escapechar=|,caption={\scriptsize\texttt{coin(bias)}},label={listing:ProbProgramCoinEmulationParam}]
bool c1, c2 := true;
ifp (bias)
  then c1 := true;
  else c1 := false;
ifp (bias)
  then c2 := true;
  else c2 := false;
observe(!(c1 = c2));
return(c1);
\end{lstlisting}
\end{minipage}
\quad
    \begin{minipage}[t]{.3\textwidth}
      \footnotesize 
% \vspace{-1.5em}
\begin{lstlisting}[%numbers=right,
basicstyle={\scriptsize\ttfamily},escapechar=|,caption={\scriptsize\texttt{obsLoop(x0,n0)}},label={listing:obsInLoopParam}]
double x := 0;
double y := 0;
int n := 0;
while (x < x0) {
  n := n + 1;
  y $\sim$ normal(1,1);
  observe(0 <= y <= 2);
  x := x + y;}
observe(n >= n0);
return(n);
\end{lstlisting}
\end{minipage}
\end{figure*}

\begin{table*}[tbp]
 \centering \caption{experimental results with more details, supplementing Table~\ref{table:collectiveResults}. 
% The columns ``sample'' show the numbers of samples  after two minutes of sampling. The ``gap'' is the value $|(\text{sample mean after 2 min.}) - (\text{sample mean after 1 min.})|$; a large ``gap'' suggests that the samples have not yet converged. The settings where the obtained samples are obviously wrong are designated by the red color.
}
\label{table:collectiveResultsSuppl}
\begin{adjustbox}{angle=90}
 \scriptsize
\begin{adjustbox}{scale=.67}
 \begin{minipage}{100em}
 Comparing our proposal ($\Schism$) with Anglican (with different sampling algorithms). ``A.'' stands for Anglican. Experiments ran for designated timeout seconds, or until 500K samples were obtained (marked with ``$\ge$500K''). For the first group of programs, the ground truth is known, so the KL-divergence from it to the samples is shown. For the latter programs, the ground truth is not known, so the mean and standard deviation is shown. The numbers are the average of ten runs.

 \vspace{2em}
 \begin{tabular}{%@{}l@{\,}|c|c|c|c|c|c|c|
  r||p{0.6cm}p{1.5cm}|p{0.6cm}p{1.5cm}|p{0.6cm}p{1.5cm}||p{0.6cm}p{1.5cm}|p{0.6cm}p{1.5cm}|p{0.6cm}p{1.5cm}
   ||p{0.6cm}p{1.5cm}|p{0.6cm}p{1.5cm}|p{0.6cm}p{1.5cm}
  }
  method (timeout)
 &
 \multicolumn{2}{c|}{$\Schism$ (10 sec.)}
 &
 \multicolumn{2}{c|}{$\Schism$ (60 sec.)}
 &
 \multicolumn{2}{c||}{$\Schism$ (600 sec.)}
 &
 \multicolumn{2}{c|}{A.-RMH (60 sec.)}
 &
 \multicolumn{2}{c|}{A.-SMC (60 sec.)}
 &
 \multicolumn{2}{c||}{A.-IPMCMC (60 sec.)}
 &
 \multicolumn{2}{c|}{A.-RMH (600 sec.)}
 &
 \multicolumn{2}{c|}{A.-SMC (600 sec.)}
 &
 \multicolumn{2}{c}{A.-IPMCMC (600 sec.)}
 % &
 % \multicolumn{2}{c|}{\texttt{poisCd(6,30)}}
 % &
 % \multicolumn{2}{c}{\texttt{geomIt2(0.5,20)}}
 \\
 target program
  &
  samples
  &
  KL-div.
  &
  samples
  &
  KL-div.
  &
  samples
  &
  KL-div.
  &
  samples
  &
  KL-div.
  &
  samples
  &
  KL-div.
  &
  samples
  &
  KL-div.
  &
  samples
  &
  KL-div.
  &
  samples
  &
  KL-div.
  &
  samples
  &
  KL-div.
 \\\hline
 % - detailed table:
 %   https://docs.google.com/spreadsheets/d/1-Wgs9E1KNPKw4g8DtqwW3uZMHH9EgocI7b0LiG38jxQ/edit?ts=5ed8e9dc#gid=326400773
 %   https://docs.google.com/spreadsheets/d/1vLkI7NsXdsB8iw1u08zfZn1M1gaCvLKC8QJSA4uuUvU/edit?ts=5ed90d98#gid=1791218101
 %   https://docs.google.com/spreadsheets/d/11SjL5skgPqEAO2JEYD_55upgsB_tLDeWZC34Vpr9I9U/edit?ts=5ed9dd77#gid=1471224379
 % - generated by running
 %   $ ./expr_summary_used_for_202006042200JST.py output_202006042200JST.yml --tex_source_output_location experiment_results_202006042200JST.tex
 % ** TODO ** KL divergence for uniform_conditioned_2
 \input{experiment_results_whole}
 \end{tabular}
\end{minipage}\end{adjustbox}
\end{adjustbox}
\end{table*}

\begin{figure*}[tbp]

\begin{minipage}{.45\textwidth}
% https://ichiro-ansible.s3.amazonaws.com/tmp/hierarchical-sampling/experiment_results/37e54d2021_0202_145840_UTC/convergence.png
 \raisebox{-\height}{ \includegraphics[width=\textwidth]{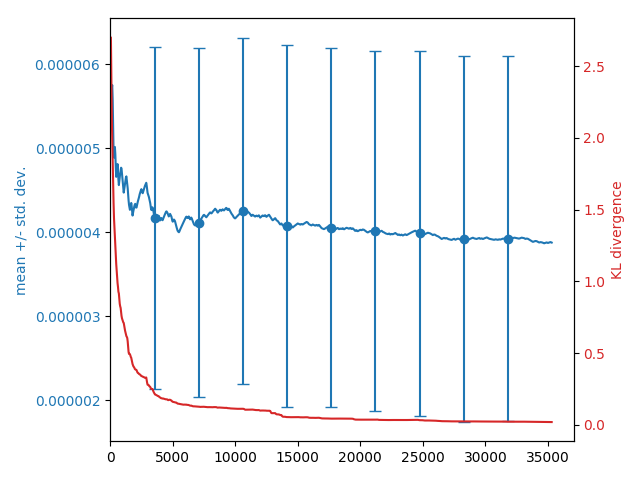}}
\caption{KL divergence (red), and mean and standard deviation (blue), $\texttt{unifCd}(18)$, Schism, 600 seconds}
\end{minipage}
\quad
% \begin{minipage}{.45\textwidth}
%  \raisebox{-\height}{ \includegraphics[width=\textwidth]{mixed_0_schism_600sec_conv.png}}
% \caption{KL divergence (red), and mean and standard deviation (blue), $\texttt{mixed}(0)$, Schism, 600 seconds}
% \end{minipage}
% \quad
\begin{minipage}{.45\textwidth}
% https://ichiro-ansible.s3.amazonaws.com/tmp/hierarchical-sampling/experiment_results/9a4ca12021_0131_103718_UTC/convergence.png
 \raisebox{-\height}{ \includegraphics[width=\textwidth]{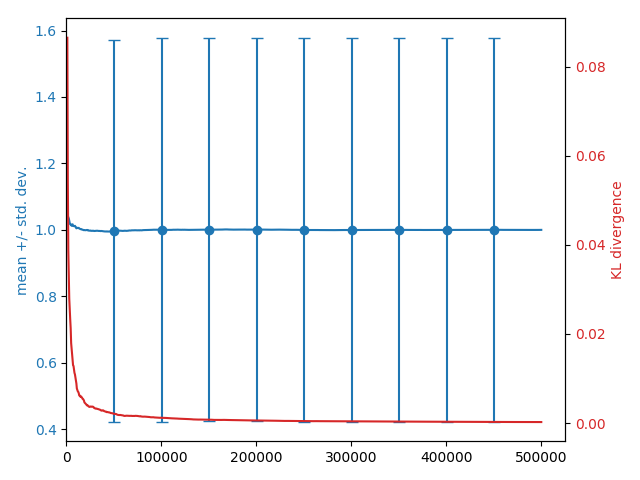}}
\caption{KL divergence (red), and mean and standard deviation (blue), $\texttt{mixed}(5)$, Schism, 600 seconds}
\end{minipage}

\begin{minipage}{.45\textwidth}
% https://ichiro-ansible.s3.amazonaws.com/tmp/hierarchical-sampling/experiment_results/dfb55b2021_0202_160411_UTC/convergence.png
 \raisebox{-\height}{ \includegraphics[width=\textwidth]{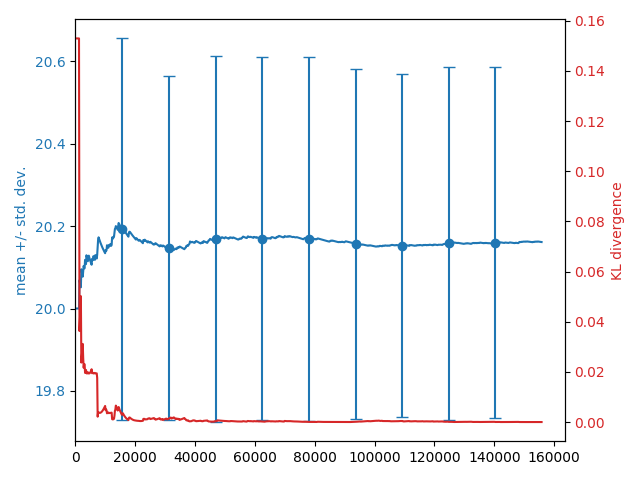}}
\caption{KL divergence (red), and mean and standard deviation (blue), $\texttt{poisCd}(3,20)$, Schism, 600 seconds}
\end{minipage}
\quad
\begin{minipage}{.45\textwidth}
% https://ichiro-ansible.s3.amazonaws.com/tmp/hierarchical-sampling/experiment_results/e991b52021_0131_110625_UTC/convergence.png
 \raisebox{-\height}{ \includegraphics[width=\textwidth]{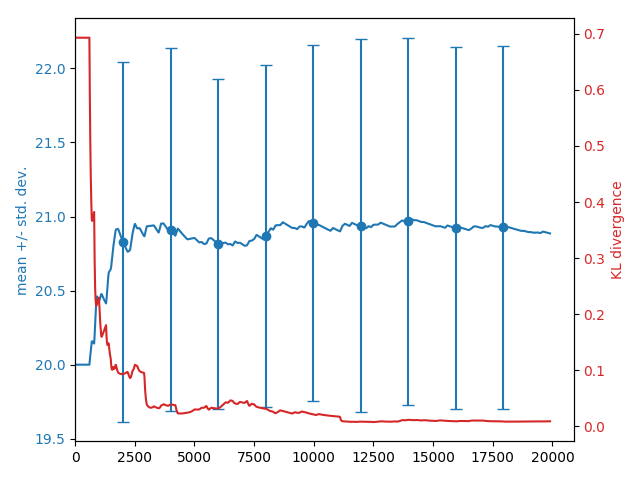}}
\caption{KL divergence (red), and mean and standard deviation (blue), $\texttt{geomIt}(0.5,20)$, Schism, 600 seconds}
\end{minipage}

\end{figure*}

% \section{}
% \label{}

% %% The Appendices part is started with the command \appendix;
% %% appendix sections are then done as normal sections
% %% \appendix

% %% \section{}
% %% \label{}

% %% If you have bibdatabase file and want bibtex to generate the
% %% bibitems, please use
% %%
% %%  \bibliographystyle{elsarticle-num} 
% %%  \bibliography{<your bibdatabase>}

% %% else use the following coding to input the bibitems directly in the
% %% TeX file.




% \end{thebibliography}

\end{document}
\endinput
%%
%% End of file `elsarticle-template-num.tex'.

% LocalWords:  nii sokendai ey Eberhart kyoto Katsumata addressline LLC
% LocalWords:  Chiyoda Shonan Hayama Kanagawa ShinNihon Yurakucho ifp
% LocalWords:  Yoshida Honmachi Prog escapechar bool obsLoop obsInLoop
% LocalWords:  ProbProgramCoinEmulation TolpinMYW MansinghkaSP Pyro le
% LocalWords:  CarpenterGHLGBBGLR DNNs pmlr Hakaru ShanR GehrMV EfProb
% LocalWords:  ChoJ variational resampling MCMC MAB NoriHRS branchings
% LocalWords:  IAS pCFGs lcllll pCFG AgrawalCN lu ru MorganMS ZhouYTR
% LocalWords:  unnormalized unif CDF cond condPropDemo Winskel priori
% LocalWords:  linebackgroundcolor nilpotency Slivkins UCB MCTS playout
% LocalWords:  SCalable HIerarchical SaMpling RMH IPMCMC BBVB xlarge dt
% LocalWords:  vCPUs unifCd poisCd poisson poisCdS geomIt nestLp init
% LocalWords:  blkl QueirozBC Kenta's PPL compositionally PIMH PMCMC MH
% LocalWords:  monteCarloTreeSearchNotes pdf tex Ohad Kammar Scibior RL
% LocalWords:  Akihisa Yamada JPMJER JST CyPhAI JPMJCR Hoeffding's nt
% LocalWords:  Markovian ProbProgramCoinEmulationParam obsInLoopParam

%% file: experiment_results_part.tex
\texttt{unifCd(10)} & 3.07K & 0.281 & 9.37K & 0.0793 & 37.6K & 0.0174 & $\ge$500K & 0.869 & 492K & 0.0938 \\
\texttt{unifCd(20)} & 1.99K & 0.416 & 6.82K & 0.103 & 34.5K & 0.02 & 376K & 6.44 & 550 & 5.21 \\
\hline
\texttt{poisCd(6,20)} & 2.11K & 0.0618 & 7.43K & 0.00764 & 72.6K & 0.000879 & 323K & 0.000141 & 1.25K & 0.108 \\
\texttt{poisCd(6,30)} & 1.14K & 0.0726 & 8.84K & 0.00577 & 98.4K & 0.000294 & 0 & --- & 0 & --- \\
\hline
\texttt{geomIt(0.5,20)} & 1.42K & 0.261 & 4.97K & 0.0425 & 20K & 0.0114 & 139K & 0.00119 & 0 & --- \\
\texttt{geomIt(0.1,5)} & 3.05K & 2.08 & 8.54K & 2.0 & 27.6K & 1.98 & 411K & 1.95 & 2.66K & 1.95 \\
\texttt{geomIt(0.1,20)} & 1.31K & 2.14 & 4.73K & 2.04 & 20.1K & 1.98 & 0 & --- & 0 & --- \\
\hline
\texttt{mixed(0)} & 35.5K & 0.0813 & 198K & 0.0734 & $\ge$500K & 0.0724 & 457K & 0.0738 & $\ge$500K & 0.0724 \\
\hline
\texttt{coin(0.1)} & 55.5K & 1.67e-06 & 270K & 6.82e-08 & $\ge$500K & 2.03e-08 & $\ge$500K & 4.65e-05 & $\ge$500K & 7.66e-06 \\
\texttt{coin(0.001)} & 55.4K & 1.66e-06 & 270K & 6.86e-08 & $\ge$500K & 2.05e-08 & $\ge$500K & 0.00204 & 280K & 0.000143 \\
\hline & samples & mean $\pm$ std & samples & mean $\pm$ std & samples & mean $\pm$ std & samples & mean $\pm$ std & samples & mean $\pm$ std \\
\hline
\texttt{unifCd2(10)} & 2.34K & 10.6 $\pm$ 3.35 & 7.07K & 10.8 $\pm$ 3.45 & 26.8K & 10.9 $\pm$ 3.51 & 214K & 11.2 $\pm$ 3.67 & 266K & 11.0 $\pm$ 3.59 \\
\texttt{unifCd2(20)} & 1.29K & 20.3 $\pm$ 4.56 & 4.92K & 20.7 $\pm$ 4.65 & 23.1K & 20.9 $\pm$ 4.71 & 83K & 21.6 $\pm$ 4.69 & 300 & 21.8 $\pm$ 2.26 \\
\hline
\texttt{geomIt2(0.5,20)} & 830 & 22.5 $\pm$ 0.258 & 3.32K & 23.5 $\pm$ 1.02 & 14.7K & 23.9 $\pm$ 1.33 & 38.8K & 24.0 $\pm$ 1.44 & 0 & --- \\
\texttt{geomIt2(0.1,5)} & 2.24K & 6.4 $\pm$ 0.535 & 6.27K & 6.44 $\pm$ 0.575 & 21.3K & 6.46 $\pm$ 0.594 & 134K & 6.53 $\pm$ 0.636 & 0 & --- \\
\texttt{geomIt2(0.1,20)} & 860 & 22.2 $\pm$ 0.335 & 3.36K & 22.5 $\pm$ 0.572 & 14.5K & 22.5 $\pm$ 0.617 & 0 & --- & 0 & --- \\
\hline
\texttt{obsLoop(3,10)} & 1.9K & 10.0 $\pm$ 0.136 & 4.58K & 10.0 $\pm$ 0.169 & 13.1K & 10.1 $\pm$ 0.209 & 141K & 10.1 $\pm$ 0.299 & 0 & --- \\
\texttt{obsLoop(3,12)} & 1.72K & 12.0 $\pm$ 0.0772 & 4.27K & 12.0 $\pm$ 0.153 & 12.4K & 12.0 $\pm$ 0.172 & 0 & --- & 0 & --- \\
\hline
\texttt{poisCdS(12)} & 100 & 15.0 $\pm$ 0.0 & 1.23K & 20.3 $\pm$ 0.0209 & 7.28K & 25.7 $\pm$ 1.25 & 16.5K & 25.2 $\pm$ 1.38 & 9.3K & 25.5 $\pm$ 1.31 \\
\texttt{poisCdS(3)} & 100 & 14.0 $\pm$ 0.0 & 1.21K & 19.4 $\pm$ 0.168 & 13.5K & 23.3 $\pm$ 1.06 & 0 & --- & 0 & --- \\
\hline
\texttt{nestLp(1)} & 0 & --- & 0 & --- & 1.47K & 9.41 $\pm$ 0.279 & 0 & --- & 0 & --- \\
\texttt{nestLp(3)} & 0 & --- & 0 & --- & 1.22K & 9.28 $\pm$ 0.238 & 0 & --- & 0 & --- \\
\hline
\texttt{ADS(0.5)} & 0 & --- & 0 & --- & 14.3K & 20.9 $\pm$ 1.13 & 0 & --- & 0 & --- 

%% file: experiment_results_ablation.tex
\texttt{unifCd(10)} & 37.6K & 0.0174 & 29.6K & 0.0244 & 23.4K & 0.0295 & 13.1K & 2.18 & $\ge$500K & 0.0944 \\
\texttt{unifCd(20)} & 34.5K & 0.02 & 22.5K & 0.0317 & 15.1K & 0.0443 & 0 & --- & 5.88K & 2.9 \\
\hline
\texttt{poisCd(6,30)} & 98.4K & 0.000294 & 15.6K & 0.00312 & 11.3K & 0.00647 & 0 & --- & 0 & --- \\
\hline & samples & mean $\pm$ std & samples & mean $\pm$ std & samples & mean $\pm$ std & samples & mean $\pm$ std & samples & mean $\pm$ std \\
\hline
\texttt{nestLp(1)} & 1.47K & 9.41 $\pm$ 0.279 & 0 & --- & 0 & --- & 0 & --- & 0 & --- \\
\texttt{nestLp(3)} & 1.22K & 9.28 $\pm$ 0.238 & 0 & --- & 0 & --- & 0 & --- & 0 & --- \\
\hline
\texttt{ADS(0.5)} & 14.3K & 20.9 $\pm$ 1.13 & 0 & --- & 0 & --- & 0 & --- & 0 & --- 

%% file: experiment_results_whole.tex
\texttt{unifCd(10)} & 3.07K & 0.281 & 9.37K & 0.0793 & 37.6K & 0.0174 & $\ge$500K & 0.869 & 492K & 0.0938 & 0 & --- & $\ge$500K & 0.866 & $\ge$500K & 0.0944 & 0 & --- \\
\texttt{unifCd(15)} & 2.86K & 0.34 & 7.92K & 0.0983 & 35.7K & 0.0185 & $\ge$500K & 3.96 & 18.1K & 1.83 & 0 & --- & $\ge$500K & 3.73 & 179K & 0.32 & 0 & --- \\
\texttt{unifCd(18)} & 2.18K & 0.373 & 7.22K & 0.104 & 35.5K & 0.019 & 442K & 6.03 & 2.15K & 3.9 & 0 & --- & $\ge$500K & 5.76 & 21.1K & 1.7 & 0 & --- \\
\texttt{unifCd(20)} & 1.99K & 0.416 & 6.82K & 0.103 & 34.5K & 0.02 & 376K & 6.44 & 550 & 5.21 & 0 & --- & $\ge$500K & 6.5 & 5.88K & 2.9 & 0 & --- \\
\hline
\texttt{poisCd(6,20)} & 2.11K & 0.0618 & 7.43K & 0.00764 & 72.6K & 0.000879 & 323K & 0.000141 & 1.25K & 0.108 & $\ge$500K & inf & $\ge$500K & 0.000108 & 14.9K & 0.0223 & $\ge$500K & inf \\
\texttt{poisCd(6,30)} & 1.14K & 0.0726 & 8.84K & 0.00577 & 98.4K & 0.000294 & 0 & --- & 0 & --- & $\ge$500K & inf & 0 & --- & 0 & --- & $\ge$500K & inf \\
\texttt{poisCd(3,20)} & 1.93K & 0.0308 & 15.1K & 0.00178 & 152K & 6.49e-05 & 0 & --- & 0 & --- & $\ge$500K & inf & 0 & --- & 0 & --- & $\ge$500K & inf \\
\texttt{poisCd(3,30)} & 0 & --- & 0 & --- & 0 & --- & 0 & --- & 0 & --- & $\ge$500K & nan & 0 & --- & 0 & --- & $\ge$500K & nan \\
\hline
\texttt{geomIt(0.5,5)} & 3.11K & 0.0881 & 8.34K & 0.0201 & 27.7K & 0.00646 & 324K & 0.000158 & $\ge$500K & 0.000496 & $\ge$500K & 0.000726 & $\ge$500K & 0.0001 & $\ge$500K & 0.000551 & $\ge$500K & 0.000708 \\
\texttt{geomIt(0.5,20)} & 1.42K & 0.261 & 4.97K & 0.0425 & 20K & 0.0114 & 139K & 0.00119 & 0 & --- & $\ge$500K & inf & $\ge$500K & 0.0002 & 2.2K & 0.162 & $\ge$500K & inf \\
\texttt{geomIt(0.1,5)} & 3.05K & 2.08 & 8.54K & 2.0 & 27.6K & 1.98 & 411K & 1.95 & 2.66K & 1.95 & $\ge$500K & inf & $\ge$500K & 1.95 & 28.9K & 1.97 & $\ge$500K & inf \\
\texttt{geomIt(0.1,20)} & 1.31K & 2.14 & 4.73K & 2.04 & 20.1K & 1.98 & 0 & --- & 0 & --- & $\ge$500K & inf & 0 & --- & 0 & --- & $\ge$500K & inf \\
\hline
\texttt{mixed(0)} & 35.5K & 0.0813 & 198K & 0.0734 & $\ge$500K & 0.0724 & 457K & 0.0738 & $\ge$500K & 0.0724 & $\ge$500K & 0.0729 & $\ge$500K & 0.0729 & $\ge$500K & 0.0724 & $\ge$500K & 0.0733 \\
\hline
\texttt{coin(0.1)} & 55.5K & 1.67e-06 & 270K & 6.82e-08 & $\ge$500K & 2.03e-08 & $\ge$500K & 4.65e-05 & $\ge$500K & 7.66e-06 & $\ge$500K & 7.32e-06 & $\ge$500K & 1.38e-05 & $\ge$500K & 6.74e-06 & $\ge$500K & 3.9e-06 \\
\texttt{coin(0.001)} & 55.4K & 1.66e-06 & 270K & 6.86e-08 & $\ge$500K & 2.05e-08 & $\ge$500K & 0.00204 & 280K & 0.000143 & $\ge$500K & 0.000332 & $\ge$500K & 0.00386 & $\ge$500K & 3.91e-05 & $\ge$500K & 0.000804 \\
\texttt{coin(0.00001)} & 0 & --- & 0 & --- & 0 & --- & 0 & --- & 0 & --- & 0 & --- & 0 & --- & 0 & --- & 0 & --- \\
\hline & samples & mean $\pm$ std & samples & mean $\pm$ std & samples & mean $\pm$ std & samples & mean $\pm$ std & samples & mean $\pm$ std & samples & mean $\pm$ std & samples & mean $\pm$ std & samples & mean $\pm$ std & samples & mean $\pm$ std \\
\hline
\texttt{unifCd2(10)} & 2.34K & 10.6 $\pm$ 3.35 & 7.07K & 10.8 $\pm$ 3.45 & 26.8K & 10.9 $\pm$ 3.51 & 214K & 11.2 $\pm$ 3.67 & 266K & 11.0 $\pm$ 3.59 & $\ge$500K & 11.0 $\pm$ 3.69 & $\ge$500K & 10.9 $\pm$ 3.54 & $\ge$500K & 11.0 $\pm$ 3.63 & $\ge$500K & 10.9 $\pm$ 3.65 \\
\texttt{unifCd2(15)} & 1.73K & 15.4 $\pm$ 4.0 & 5.77K & 15.8 $\pm$ 4.11 & 24.8K & 15.9 $\pm$ 4.17 & 168K & 15.5 $\pm$ 3.96 & 8.67K & 16.0 $\pm$ 4.25 & $\ge$500K & 11.9 $\pm$ 7.12 & $\ge$500K & 15.8 $\pm$ 4.09 & 95.2K & 16.0 $\pm$ 4.28 & $\ge$500K & 12.5 $\pm$ 7.13 \\
\texttt{unifCd2(18)} & 1.53K & 18.4 $\pm$ 4.33 & 5.06K & 18.7 $\pm$ 4.44 & 23.8K & 18.9 $\pm$ 4.52 & 89.8K & 18.7 $\pm$ 4.35 & 1.37K & 19.2 $\pm$ 4.86 & $\ge$500K & 4.0 $\pm$ 5.72 & $\ge$500K & 19.2 $\pm$ 4.4 & 12.2K & 18.9 $\pm$ 4.73 & $\ge$500K & 4.47 $\pm$ 6.18 \\
\texttt{unifCd2(20)} & 1.29K & 20.3 $\pm$ 4.56 & 4.92K & 20.7 $\pm$ 4.65 & 23.1K & 20.9 $\pm$ 4.71 & 83K & 21.6 $\pm$ 4.69 & 300 & 21.8 $\pm$ 2.26 & $\ge$500K & 2.54 $\pm$ 3.49 & 463K & 22.2 $\pm$ 4.72 & 3.5K & 20.8 $\pm$ 4.83 & $\ge$500K & 2.35 $\pm$ 2.95 \\
\hline
\texttt{poisCd2(6,20)} & 1.45K & 18.2 $\pm$ 0.429 & 4.82K & 18.4 $\pm$ 0.673 & 35K & 18.5 $\pm$ 0.752 & 98.4K & 18.5 $\pm$ 0.801 & 4.28K & 18.4 $\pm$ 0.713 & $\ge$500K & 13.7 $\pm$ 6.16 & $\ge$500K & 18.5 $\pm$ 0.806 & 44.3K & 18.5 $\pm$ 0.781 & $\ge$500K & 13.9 $\pm$ 5.85 \\
\texttt{poisCd2(6,30)} & 1.03K & 26.5 $\pm$ nan & 3.98K & 27.1 $\pm$ 0.512 & 41.7K & 27.2 $\pm$ 0.64 & 0 & --- & 0 & --- & $\ge$500K & 6.0 $\pm$ 2.45 & 0 & --- & 0 & --- & $\ge$500K & 6.0 $\pm$ 2.45 \\
\texttt{poisCd2(3,20)} & 1.45K & 18.1 $\pm$ 0.323 & 6.97K & 18.2 $\pm$ 0.444 & 69.8K & 18.2 $\pm$ 0.496 & 0 & --- & 0 & --- & $\ge$500K & 3.0 $\pm$ 1.73 & 209K & 18.2 $\pm$ 0.485 & 0 & --- & $\ge$500K & 3.0 $\pm$ 1.73 \\
\texttt{poisCd2(3,30)} & 1.45K & 26.0 $\pm$ nan & 10.9K & 26.0 $\pm$ nan & 112K & 26.0 $\pm$ nan & 0 & --- & 0 & --- & $\ge$500K & 3.0 $\pm$ 1.73 & 0 & --- & 0 & --- & $\ge$500K & 3.0 $\pm$ 1.73 \\
\hline
\texttt{geomIt2(0.5,5)} & 2.13K & 7.35 $\pm$ 0.934 & 6.21K & 7.63 $\pm$ 1.18 & 21K & 7.79 $\pm$ 1.35 & 191K & 7.96 $\pm$ 1.54 & $\ge$500K & 7.91 $\pm$ 1.53 & $\ge$500K & 7.91 $\pm$ 1.55 & $\ge$500K & 7.96 $\pm$ 1.55 & $\ge$500K & 7.91 $\pm$ 1.54 & $\ge$500K & 7.92 $\pm$ 1.55 \\
\texttt{geomIt2(0.5,20)} & 830 & 22.5 $\pm$ 0.258 & 3.32K & 23.5 $\pm$ 1.02 & 14.7K & 23.9 $\pm$ 1.33 & 38.8K & 24.0 $\pm$ 1.44 & 0 & --- & $\ge$500K & 1.0 $\pm$ 1.42 & 400K & 24.1 $\pm$ 1.55 & 0 & --- & $\ge$500K & 1.0 $\pm$ 1.42 \\
\texttt{geomIt2(0.1,5)} & 2.24K & 6.4 $\pm$ 0.535 & 6.27K & 6.44 $\pm$ 0.575 & 21.3K & 6.46 $\pm$ 0.594 & 134K & 6.53 $\pm$ 0.636 & 0 & --- & $\ge$500K & 0.111 $\pm$ 0.351 & $\ge$500K & 6.52 $\pm$ 0.634 & 0 & --- & $\ge$500K & 0.135 $\pm$ 0.449 \\
\texttt{geomIt2(0.1,20)} & 860 & 22.2 $\pm$ 0.335 & 3.36K & 22.5 $\pm$ 0.572 & 14.5K & 22.5 $\pm$ 0.617 & 0 & --- & 0 & --- & $\ge$500K & 0.111 $\pm$ 0.352 & 0 & --- & 0 & --- & $\ge$500K & 0.111 $\pm$ 0.351 \\
\hline
\texttt{obsLoop(3,5)} & 2.76K & 5.13 $\pm$ 0.353 & 6.13K & 5.17 $\pm$ 0.409 & 15K & 5.2 $\pm$ 0.453 & 274K & 5.22 $\pm$ 0.493 & 0 & --- & 0 & --- & $\ge$500K & 5.22 $\pm$ 0.488 & 0 & --- & 0 & --- \\
\texttt{obsLoop(3,8)} & 2.06K & 8.05 $\pm$ 0.211 & 5.15K & 8.09 $\pm$ 0.271 & 13.5K & 8.1 $\pm$ 0.318 & 203K & 8.11 $\pm$ 0.345 & 0 & --- & 0 & --- & $\ge$500K & 8.11 $\pm$ 0.344 & 0 & --- & 0 & --- \\
\texttt{obsLoop(3,10)} & 1.9K & 10.0 $\pm$ 0.136 & 4.58K & 10.0 $\pm$ 0.169 & 13.1K & 10.1 $\pm$ 0.209 & 141K & 10.1 $\pm$ 0.299 & 0 & --- & 0 & --- & $\ge$500K & 10.1 $\pm$ 0.297 & 0 & --- & 0 & --- \\
\texttt{obsLoop(3,12)} & 1.72K & 12.0 $\pm$ 0.0772 & 4.27K & 12.0 $\pm$ 0.153 & 12.4K & 12.0 $\pm$ 0.172 & 0 & --- & 0 & --- & 0 & --- & 0 & --- & 0 & --- & 0 & --- \\
\texttt{obsLoop(3,15)} & 1.09K & 15.1 $\pm$ 0.0952 & 3.67K & 15.1 $\pm$ 0.126 & 11.8K & 15.0 $\pm$ 0.0757 & 0 & --- & 0 & --- & 0 & --- & 0 & --- & 0 & --- & 0 & --- \\
\texttt{obsLoop(3,20)} & 262 & 20.0 $\pm$ 0.0 & 712 & 20.4 $\pm$ 0.0 & 3.92K & 20.3 $\pm$ nan & 0 & --- & 0 & --- & 0 & --- & 0 & --- & 0 & --- & 0 & --- \\
\hline
\texttt{poisCdS(12)} & 100 & 15.0 $\pm$ 0.0 & 1.23K & 20.3 $\pm$ 0.0209 & 7.28K & 25.7 $\pm$ 1.25 & 16.5K & 25.2 $\pm$ 1.38 & 9.3K & 25.5 $\pm$ 1.31 & 111K & 25.0 $\pm$ 1.86 & 296K & 25.8 $\pm$ 1.42 & 96.4K & 25.7 $\pm$ 1.41 & $\ge$500K & 25.5 $\pm$ 1.57 \\
\texttt{poisCdS(3)} & 100 & 14.0 $\pm$ 0.0 & 1.21K & 19.4 $\pm$ 0.168 & 13.5K & 23.3 $\pm$ 1.06 & 0 & --- & 0 & --- & 472K & 7.93 $\pm$ 1.32 & 0 & --- & 0 & --- & $\ge$500K & 7.94 $\pm$ 1.31 \\
\hline
\texttt{nestLp(0.1)} & 0 & --- & 0 & --- & 1.54K & 9.5 $\pm$ 0.29 & 0 & --- & 0 & --- & 0 & --- & 0 & --- & 0 & --- & 0 & --- \\
\texttt{nestLp(1)} & 0 & --- & 0 & --- & 1.47K & 9.41 $\pm$ 0.279 & 0 & --- & 0 & --- & 0 & --- & 0 & --- & 0 & --- & 0 & --- \\
\texttt{nestLp(3)} & 0 & --- & 0 & --- & 1.22K & 9.28 $\pm$ 0.238 & 0 & --- & 0 & --- & 0 & --- & 0 & --- & 0 & --- & 0 & --- \\
\hline
\texttt{ADS(0.5)} & 0 & --- & 0 & --- & 14.3K & 20.9 $\pm$ 1.13 & 0 & --- & 0 & --- & $\ge$500K & 11.1 $\pm$ 6.41 & 0 & --- & 0 & --- & $\ge$500K & 11.1 $\pm$ 6.41 